\documentclass[a4paper,11pt,fleqn]{cas-sc}
\usepackage{tabularx,longtable,placeins,microtype}
\providecommand{\Description}[1]{}
\newcommand{\transferthesis}{The modern pretrain, fine-tune, and post-train pipeline recombines classical transfer mechanisms while retaining their problems of source mismatch, negative transfer, and retention.}
\newcolumntype{Y}{>{\raggedright\arraybackslash}X}

\ExplSyntaxOn
\cs_new_protected:Npn \eswa_citation:nn #1#2 {
  \tl_set:Nn \l_tmpa_tl {#2}
  \regex_replace_all:nnN {\s+} {} \l_tmpa_tl
  \cs_if_exist:cTF {ESWA#1\l_tmpa_tl}
    {\use:c {ESWA#1\l_tmpa_tl}}
    {\PackageError{eswa}{Missing~citation:~#2}{Rebuild~the~citation~file.}}
}
\NewDocumentCommand \citep {m} {\eswa_citation:nn {paren}{#1}}
\NewDocumentCommand \citet {m} {\eswa_citation:nn {text}{#1}}
\ExplSyntaxOff
\expandafter\def\csname ESWAparenbrohan2022rt1\endcsname{(\protect\hyperlink{ref-brohan2022rt1}{Brohan et al., 2023})}
\expandafter\def\csname ESWAparenoneill2023open,bu2025agibot\endcsname{(\protect\hyperlink{ref-bu2025agibot}{AgiBot World Contributors et al., 2025}; \protect\hyperlink{ref-oneill2023open}{Open X-Embodiment Collaboration et al., 2024})}
\expandafter\def\csname ESWAparentaylor2009transfer,lazaric2012transfer\endcsname{(\protect\hyperlink{ref-lazaric2012transfer}{Lazaric, 2012}; \protect\hyperlink{ref-taylor2009transfer}{Taylor \& Stone, 2009})}
\expandafter\def\csname ESWAparenkober2013reinforcement\endcsname{(\protect\hyperlink{ref-kober2013reinforcement}{Kober et al., 2013})}
\expandafter\def\csname ESWAparentobin2017domain,finn2017model,bousmalis2018using\endcsname{(\protect\hyperlink{ref-bousmalis2018using}{Bousmalis et al., 2018}; \protect\hyperlink{ref-finn2017model}{Finn, Abbeel, et al., 2017}; \protect\hyperlink{ref-tobin2017domain}{Tobin et al., 2017})}
\expandafter\def\csname ESWAparenoneill2023open,kim2024openvla,amin2025pistar\endcsname{(\protect\hyperlink{ref-kim2024openvla}{Kim, Pertsch, et al., 2025}; \protect\hyperlink{ref-oneill2023open}{Open X-Embodiment Collaboration et al., 2024}; \protect\hyperlink{ref-amin2025pistar}{Physical Intelligence, Amin, et al., 2025})}
\expandafter\def\csname ESWAparenoneill2023open,hejna2024remix\endcsname{(\protect\hyperlink{ref-hejna2024remix}{Hejna et al., 2025}; \protect\hyperlink{ref-oneill2023open}{Open X-Embodiment Collaboration et al., 2024})}
\expandafter\def\csname ESWAparenxing2025shortcut\endcsname{(\protect\hyperlink{ref-xing2025shortcut}{Xing et al., 2025})}
\expandafter\def\csname ESWAparenhansen2023pretraining,dasari2023unbiased\endcsname{(\protect\hyperlink{ref-dasari2023unbiased}{Dasari et al., 2023}; \protect\hyperlink{ref-hansen2023pretraining}{Hansen et al., 2023})}
\expandafter\def\csname ESWAparenzhou2025chatvla,driess2025knowledge\endcsname{(\protect\hyperlink{ref-driess2025knowledge}{Driess et al., 2025}; \protect\hyperlink{ref-zhou2025chatvla}{Zhou, Zhu, et al., 2025})}
\expandafter\def\csname ESWAparenmuratore2022robot\endcsname{(\protect\hyperlink{ref-muratore2022robot}{Muratore et al., 2022})}
\expandafter\def\csname ESWAparenjaquier2025transfer\endcsname{(\protect\hyperlink{ref-jaquier2025transfer}{Jaquier et al., 2025})}
\expandafter\def\csname ESWAparenfiroozi2025foundation\endcsname{(\protect\hyperlink{ref-firoozi2025foundation}{Firoozi et al., 2025})}
\expandafter\def\csname ESWAparenpan2010survey,taylor2009transfer\endcsname{(\protect\hyperlink{ref-pan2010survey}{Pan \& Yang, 2010}; \protect\hyperlink{ref-taylor2009transfer}{Taylor \& Stone, 2009})}
\expandafter\def\csname ESWAparenwang2019characterizing,zhang2023survey\endcsname{(\protect\hyperlink{ref-wang2019characterizing}{Z. Wang et al., 2019}; \protect\hyperlink{ref-zhang2023survey}{Zhang et al., 2023})}
\expandafter\def\csname ESWAparenbendavid2010theory\endcsname{(\protect\hyperlink{ref-bendavid2010theory}{Ben-David et al., 2010})}
\expandafter\def\csname ESWAparenross2011reduction\endcsname{(\protect\hyperlink{ref-ross2011reduction}{Ross et al., 2011})}
\expandafter\def\csname ESWAparenmansour2009domain\endcsname{(\protect\hyperlink{ref-mansour2009domain}{Mansour et al., 2009})}
\expandafter\def\csname ESWAparenkearns2002near\endcsname{(\protect\hyperlink{ref-kearns2002near}{Kearns \& Singh, 2002})}
\expandafter\def\csname ESWAparenarjovsky2019irm\endcsname{(\protect\hyperlink{ref-arjovsky2019irm}{Arjovsky et al., 2019})}
\expandafter\def\csname ESWAparenshah2023gnm,laskin2023incontext,brohan2023rt2\endcsname{(\protect\hyperlink{ref-laskin2023incontext}{Laskin et al., 2023}; \protect\hyperlink{ref-shah2023gnm}{Shah, Sridhar, Bhorkar, et al., 2023}; \protect\hyperlink{ref-brohan2023rt2}{Zitkovich et al., 2023})}
\expandafter\def\csname ESWAparentaylor2009transfer\endcsname{(\protect\hyperlink{ref-taylor2009transfer}{Taylor \& Stone, 2009})}
\expandafter\def\csname ESWAparenpumacay2024colosseum,gao2025taxonomy\endcsname{(\protect\hyperlink{ref-gao2025taxonomy}{Gao et al., 2026}; \protect\hyperlink{ref-pumacay2024colosseum}{Pumacay et al., 2024})}
\expandafter\def\csname ESWAparenliu2023libero,zhou2025chatvla\endcsname{(\protect\hyperlink{ref-liu2023libero}{B. Liu et al., 2023}; \protect\hyperlink{ref-zhou2025chatvla}{Zhou, Zhu, et al., 2025})}
\expandafter\def\csname ESWAparenthrun1995lifelong,sutton1999between,kearns2002near\endcsname{(\protect\hyperlink{ref-kearns2002near}{Kearns \& Singh, 2002}; \protect\hyperlink{ref-sutton1999between}{Sutton et al., 1999}; \protect\hyperlink{ref-thrun1995lifelong}{Thrun \& Mitchell, 1995})}
\expandafter\def\csname ESWAparenoneill2023open,kim2024openvla,amin2025pistar,assran2025vjepa\endcsname{(\protect\hyperlink{ref-assran2025vjepa}{Assran et al., 2025}; \protect\hyperlink{ref-kim2024openvla}{Kim, Pertsch, et al., 2025}; \protect\hyperlink{ref-oneill2023open}{Open X-Embodiment Collaboration et al., 2024}; \protect\hyperlink{ref-amin2025pistar}{Physical Intelligence, Amin, et al., 2025})}
\expandafter\def\csname ESWAtexttobin2017domain\endcsname{Tobin et al. (\protect\hyperlink{ref-tobin2017domain}{2017})}
\expandafter\def\csname ESWAtextsadeghi2017cad2rl\endcsname{Sadeghi and Levine (\protect\hyperlink{ref-sadeghi2017cad2rl}{2017})}
\expandafter\def\csname ESWAtextpeng2018sim\endcsname{Peng et al. (\protect\hyperlink{ref-peng2018sim}{2018})}
\expandafter\def\csname ESWAtexttan2018sim\endcsname{J. Tan et al. (\protect\hyperlink{ref-tan2018sim}{2018})}
\expandafter\def\csname ESWAparenandrychowicz2020learning\endcsname{(\protect\hyperlink{ref-andrychowicz2020learning}{Andrychowicz et al., 2020})}
\expandafter\def\csname ESWAparenakkaya2019solving\endcsname{(\protect\hyperlink{ref-akkaya2019solving}{OpenAI et al., 2019})}
\expandafter\def\csname ESWAparenchebotar2019closing\endcsname{(\protect\hyperlink{ref-chebotar2019closing}{Chebotar et al., 2019})}
\expandafter\def\csname ESWAparenmuratore2021data\endcsname{(\protect\hyperlink{ref-muratore2021data}{Muratore et al., 2021})}
\expandafter\def\csname ESWAtextmuratore2022robot\endcsname{Muratore et al. (\protect\hyperlink{ref-muratore2022robot}{2022})}
\expandafter\def\csname ESWAparenrajeswaran2017epopt\endcsname{(\protect\hyperlink{ref-rajeswaran2017epopt}{Rajeswaran et al., 2017})}
\expandafter\def\csname ESWAparenpinto2017robust\endcsname{(\protect\hyperlink{ref-pinto2017robust}{Pinto et al., 2017})}
\expandafter\def\csname ESWAtexthwangbo2019learning\endcsname{Hwangbo et al. (\protect\hyperlink{ref-hwangbo2019learning}{2019})}
\expandafter\def\csname ESWAparenjohannink2019residual\endcsname{(\protect\hyperlink{ref-johannink2019residual}{Johannink et al., 2019})}
\expandafter\def\csname ESWAparenhe2025asap\endcsname{(\protect\hyperlink{ref-he2025asap}{T. He, Gao, et al., 2025})}
\expandafter\def\csname ESWAparensong2024learning\endcsname{(\protect\hyperlink{ref-song2024learning}{Song et al., 2025})}
\expandafter\def\csname ESWAparentorne2024reconciling\endcsname{(\protect\hyperlink{ref-torne2024reconciling}{Torne et al., 2024})}
\expandafter\def\csname ESWAtextlee2020learning\endcsname{Lee et al. (\protect\hyperlink{ref-lee2020learning}{2020})}
\expandafter\def\csname ESWAparenkumar2021rma\endcsname{(\protect\hyperlink{ref-kumar2021rma}{Kumar et al., 2021})}
\expandafter\def\csname ESWAparenradosavovic2024real\endcsname{(\protect\hyperlink{ref-radosavovic2024real}{Radosavovic et al., 2024})}
\expandafter\def\csname ESWAparenmakoviychuk2021isaac\endcsname{(\protect\hyperlink{ref-makoviychuk2021isaac}{Makoviychuk et al., 2021})}
\expandafter\def\csname ESWAtextrudin2021learning\endcsname{Rudin et al. (\protect\hyperlink{ref-rudin2021learning}{2022})}
\expandafter\def\csname ESWAparenzhuang2023robot,handa2023dextreme,qi2022inhand\endcsname{(\protect\hyperlink{ref-handa2023dextreme}{Handa et al., 2023}; \protect\hyperlink{ref-qi2022inhand}{Qi, Kumar, et al., 2023}; \protect\hyperlink{ref-zhuang2023robot}{Zhuang et al., 2023})}
\expandafter\def\csname ESWAparenkaufmann2023champion\endcsname{(\protect\hyperlink{ref-kaufmann2023champion}{Kaufmann et al., 2023})}
\expandafter\def\csname ESWAparenhe2024learning,he2024omnih2o,he2025hover\endcsname{(\protect\hyperlink{ref-he2024learning}{T. He et al., 2024}; \protect\hyperlink{ref-he2024omnih2o}{T. He, Luo, et al., 2025}; \protect\hyperlink{ref-he2025hover}{T. He, Xiao, et al., 2025})}
\expandafter\def\csname ESWAparenma2024eureka\endcsname{(\protect\hyperlink{ref-ma2024eureka}{Ma, Liang, Wang, Huang, et al., 2024})}
\expandafter\def\csname ESWAparenma2024dreureka\endcsname{(\protect\hyperlink{ref-ma2024dreureka}{Ma, Liang, Wang, Wang, et al., 2024})}
\expandafter\def\csname ESWAparenbousmalis2018using\endcsname{(\protect\hyperlink{ref-bousmalis2018using}{Bousmalis et al., 2018})}
\expandafter\def\csname ESWAparenjames2019sim,kalashnikov2018qtopt\endcsname{(\protect\hyperlink{ref-james2019sim}{James et al., 2019}; \protect\hyperlink{ref-kalashnikov2018qtopt}{Kalashnikov et al., 2018})}
\expandafter\def\csname ESWAparenqureshi2025splatsim\endcsname{(\protect\hyperlink{ref-qureshi2025splatsim}{Qureshi et al., 2025})}
\expandafter\def\csname ESWAparenwang2022tacto\endcsname{(\protect\hyperlink{ref-wang2022tacto}{S. Wang et al., 2022})}
\expandafter\def\csname ESWAtextchurch2021tactile\endcsname{Church et al. (\protect\hyperlink{ref-church2021tactile}{2022})}
\expandafter\def\csname ESWAparenhiguera2024sparsh\endcsname{(\protect\hyperlink{ref-higuera2024sparsh}{Higuera et al., 2025})}
\expandafter\def\csname ESWAparenqi2023general\endcsname{(\protect\hyperlink{ref-qi2023general}{Qi, Yi, et al., 2023})}
\expandafter\def\csname ESWAparenyu2024mimictouch,liu2024maniwav\endcsname{(\protect\hyperlink{ref-liu2024maniwav}{Z. Liu et al., 2025}; \protect\hyperlink{ref-yu2024mimictouch}{K. Yu et al., 2025})}
\expandafter\def\csname ESWAtextmatas2018sim\endcsname{Matas et al. (\protect\hyperlink{ref-matas2018sim}{2018})}
\expandafter\def\csname ESWAtextseita2020fabric\endcsname{Seita et al. (\protect\hyperlink{ref-seita2020fabric}{2020})}
\expandafter\def\csname ESWAtextganapathi2021correspondences\endcsname{Ganapathi et al. (\protect\hyperlink{ref-ganapathi2021correspondences}{2021})}
\expandafter\def\csname ESWAtextha2021flingbot\endcsname{Ha and Song (\protect\hyperlink{ref-ha2021flingbot}{2022})}
\expandafter\def\csname ESWAparenhwangbo2019learning,he2025asap,torne2024reconciling\endcsname{(\protect\hyperlink{ref-he2025asap}{T. He, Gao, et al., 2025}; \protect\hyperlink{ref-hwangbo2019learning}{Hwangbo et al., 2019}; \protect\hyperlink{ref-torne2024reconciling}{Torne et al., 2024})}
\expandafter\def\csname ESWAparenkumar2021rma,radosavovic2024real\endcsname{(\protect\hyperlink{ref-kumar2021rma}{Kumar et al., 2021}; \protect\hyperlink{ref-radosavovic2024real}{Radosavovic et al., 2024})}
\expandafter\def\csname ESWAparenthrun1995lifelong\endcsname{(\protect\hyperlink{ref-thrun1995lifelong}{Thrun \& Mitchell, 1995})}
\expandafter\def\csname ESWAparencully2015robots\endcsname{(\protect\hyperlink{ref-cully2015robots}{Cully et al., 2015})}
\expandafter\def\csname ESWAparenrusu2016policy\endcsname{(\protect\hyperlink{ref-rusu2016policy}{Rusu, Colmenarejo, et al., 2016})}
\expandafter\def\csname ESWAparenrusu2016progressive,rusu2017sim\endcsname{(\protect\hyperlink{ref-rusu2016progressive}{Rusu, Rabinowitz, et al., 2016}; \protect\hyperlink{ref-rusu2017sim}{Rusu et al., 2017})}
\expandafter\def\csname ESWAparenbarreto2017successor\endcsname{(\protect\hyperlink{ref-barreto2017successor}{Barreto et al., 2017})}
\expandafter\def\csname ESWAparentouati2021learning,touati2023does\endcsname{(\protect\hyperlink{ref-touati2023does}{Touati et al., 2023}; \protect\hyperlink{ref-touati2021learning}{Touati \& Ollivier, 2021})}
\expandafter\def\csname ESWAparentirinzoni2025zero\endcsname{(\protect\hyperlink{ref-tirinzoni2025zero}{Tirinzoni et al., 2025})}
\expandafter\def\csname ESWAparenhe2026sample\endcsname{(\protect\hyperlink{ref-he2026sample}{Y. He et al., 2027})}
\expandafter\def\csname ESWAtexthe2025fewshot\endcsname{Y. He et al. (\protect\hyperlink{ref-he2025fewshot}{2025})}
\expandafter\def\csname ESWAtexthe2026decomposition\endcsname{Y. He, Romero-Cano, et al. (\protect\hyperlink{ref-he2026decomposition}{2026})}
\expandafter\def\csname ESWAparensutton1999between\endcsname{(\protect\hyperlink{ref-sutton1999between}{Sutton et al., 1999})}
\expandafter\def\csname ESWApareneysenbach2019diversity\endcsname{(\protect\hyperlink{ref-eysenbach2019diversity}{Eysenbach et al., 2019})}
\expandafter\def\csname ESWAparenpertsch2020accelerating\endcsname{(\protect\hyperlink{ref-pertsch2020accelerating}{Pertsch et al., 2020})}
\expandafter\def\csname ESWAparenzhuang2023robot,akkaya2019solving\endcsname{(\protect\hyperlink{ref-akkaya2019solving}{OpenAI et al., 2019}; \protect\hyperlink{ref-zhuang2023robot}{Zhuang et al., 2023})}
\expandafter\def\csname ESWAparenfinn2017model,nichol2018first\endcsname{(\protect\hyperlink{ref-finn2017model}{Finn, Abbeel, et al., 2017}; \protect\hyperlink{ref-nichol2018first}{Nichol et al., 2018})}
\expandafter\def\csname ESWAparenduan2016rl,rakelly2019efficient\endcsname{(\protect\hyperlink{ref-duan2016rl}{Duan et al., 2016}; \protect\hyperlink{ref-rakelly2019efficient}{Rakelly et al., 2019})}
\expandafter\def\csname ESWAparennagabandi2019learning\endcsname{(\protect\hyperlink{ref-nagabandi2019learning}{Nagabandi et al., 2019})}
\expandafter\def\csname ESWAparenyu2019metaworld\endcsname{(\protect\hyperlink{ref-yu2019metaworld}{T. Yu et al., 2019})}
\expandafter\def\csname ESWAparenlaskin2023incontext\endcsname{(\protect\hyperlink{ref-laskin2023incontext}{Laskin et al., 2023})}
\expandafter\def\csname ESWAparenkalashnikov2021mt\endcsname{(\protect\hyperlink{ref-kalashnikov2021mt}{Kalashnikov et al., 2021})}
\expandafter\def\csname ESWAparenjang2022bcz\endcsname{(\protect\hyperlink{ref-jang2022bcz}{E. Jang et al., 2022})}
\expandafter\def\csname ESWAparennair2020awac\endcsname{(\protect\hyperlink{ref-nair2020awac}{A. Nair et al., 2020})}
\expandafter\def\csname ESWAparennakamoto2023calql\endcsname{(\protect\hyperlink{ref-nakamoto2023calql}{Nakamoto et al., 2023})}
\expandafter\def\csname ESWAparenball2023efficient\endcsname{(\protect\hyperlink{ref-ball2023efficient}{Ball et al., 2023})}
\expandafter\def\csname ESWAparenluo2024serl,luo2025precise\endcsname{(\protect\hyperlink{ref-luo2024serl}{Luo et al., 2024}; \protect\hyperlink{ref-luo2025precise}{Luo et al., 2025})}
\expandafter\def\csname ESWAparenyuan2025policy,ren2024dppo\endcsname{(\protect\hyperlink{ref-ren2024dppo}{Ren et al., 2025}; \protect\hyperlink{ref-yuan2025policy}{Yuan et al., 2025})}
\expandafter\def\csname ESWAparenchen2018hardware\endcsname{(\protect\hyperlink{ref-chen2018hardware}{T. Chen et al., 2018})}
\expandafter\def\csname ESWAparenwang2018nervenet\endcsname{(\protect\hyperlink{ref-wang2018nervenet}{T. Wang et al., 2018})}
\expandafter\def\csname ESWAparenhuang2020one\endcsname{(\protect\hyperlink{ref-huang2020one}{Huang et al., 2020})}
\expandafter\def\csname ESWAtextkurin2021morphology\endcsname{Kurin et al. (\protect\hyperlink{ref-kurin2021morphology}{2021})}
\expandafter\def\csname ESWAparengupta2022metamorph\endcsname{(\protect\hyperlink{ref-gupta2022metamorph}{Gupta et al., 2022})}
\expandafter\def\csname ESWAparenai2025towards\endcsname{(\protect\hyperlink{ref-ai2025towards}{Ai et al., 2025})}
\expandafter\def\csname ESWAparenoneill2023open\endcsname{(\protect\hyperlink{ref-oneill2023open}{Open X-Embodiment Collaboration et al., 2024})}
\expandafter\def\csname ESWAparendoshi2024scaling\endcsname{(\protect\hyperlink{ref-doshi2024scaling}{Doshi et al., 2025})}
\expandafter\def\csname ESWAparenyang2024pushing\endcsname{(\protect\hyperlink{ref-yang2024pushing}{J. Yang et al., 2024})}
\expandafter\def\csname ESWAparenwang2024scaling\endcsname{(\protect\hyperlink{ref-wang2024scaling}{L. Wang et al., 2024})}
\expandafter\def\csname ESWAparendasari2019robonet\endcsname{(\protect\hyperlink{ref-dasari2019robonet}{Dasari et al., 2019})}
\expandafter\def\csname ESWAparenreed2022generalist,bousmalis2023robocat\endcsname{(\protect\hyperlink{ref-bousmalis2023robocat}{Bousmalis et al., 2023}; \protect\hyperlink{ref-reed2022generalist}{Reed et al., 2022})}
\expandafter\def\csname ESWAparenghosh2024octo\endcsname{(\protect\hyperlink{ref-ghosh2024octo}{Octo Model Team et al., 2024})}
\expandafter\def\csname ESWAparenchen2024mirage\endcsname{(\protect\hyperlink{ref-chen2024mirage}{L. Y. Chen et al., 2024})}
\expandafter\def\csname ESWAparenchen2024rovi\endcsname{(\protect\hyperlink{ref-chen2024rovi}{L. Y. Chen et al., 2025})}
\expandafter\def\csname ESWAparenye2024latent\endcsname{(\protect\hyperlink{ref-ye2024latent}{Ye et al., 2025})}
\expandafter\def\csname ESWAparenzheng2025universal\endcsname{(\protect\hyperlink{ref-zheng2025universal}{Zheng et al., 2025})}
\expandafter\def\csname ESWAparenbu2025univla\endcsname{(\protect\hyperlink{ref-bu2025univla}{Bu et al., 2025})}
\expandafter\def\csname ESWAparenshah2023gnm\endcsname{(\protect\hyperlink{ref-shah2023gnm}{Shah, Sridhar, Bhorkar, et al., 2023})}
\expandafter\def\csname ESWAparenshah2023vint\endcsname{(\protect\hyperlink{ref-shah2023vint}{Shah, Sridhar, Dashora, et al., 2023})}
\expandafter\def\csname ESWAparensridhar2024nomad\endcsname{(\protect\hyperlink{ref-sridhar2024nomad}{Sridhar et al., 2024})}
\expandafter\def\csname ESWAparensavva2019habitat\endcsname{(\protect\hyperlink{ref-savva2019habitat}{Savva et al., 2019})}
\expandafter\def\csname ESWAparenshah2022lmnav\endcsname{(\protect\hyperlink{ref-shah2022lmnav}{Shah, Osiński, et al., 2023})}
\expandafter\def\csname ESWAparenoneill2023open,wu2024robomind\endcsname{(\protect\hyperlink{ref-oneill2023open}{Open X-Embodiment Collaboration et al., 2024}; \protect\hyperlink{ref-wu2024robomind}{K. Wu et al., 2025})}
\expandafter\def\csname ESWAparenwalke2023bridgedata,khazatsky2024droid,bu2025agibot\endcsname{(\protect\hyperlink{ref-bu2025agibot}{AgiBot World Contributors et al., 2025}; \protect\hyperlink{ref-khazatsky2024droid}{Khazatsky et al., 2024}; \protect\hyperlink{ref-walke2023bridgedata}{Walke et al., 2023})}
\expandafter\def\csname ESWAparenfang2023rh20t\endcsname{(\protect\hyperlink{ref-fang2023rh20t}{Fang et al., 2024})}
\expandafter\def\csname ESWAparenbu2025agibot\endcsname{(\protect\hyperlink{ref-bu2025agibot}{AgiBot World Contributors et al., 2025})}
\expandafter\def\csname ESWAparenhejna2024remix\endcsname{(\protect\hyperlink{ref-hejna2024remix}{Hejna et al., 2025})}
\expandafter\def\csname ESWAtextxing2025shortcut\endcsname{Xing et al. (\protect\hyperlink{ref-xing2025shortcut}{2025})}
\expandafter\def\csname ESWAtextlin2025data\endcsname{Lin et al. (\protect\hyperlink{ref-lin2025data}{2025})}
\expandafter\def\csname ESWAtextsaxena2025what\endcsname{Saxena et al. (\protect\hyperlink{ref-saxena2025what}{2025})}
\expandafter\def\csname ESWAtextyang2026data\endcsname{J. Yang et al. (\protect\hyperlink{ref-yang2026data}{2026})}
\expandafter\def\csname ESWAparenchen2024mirage,chen2024rovi\endcsname{(\protect\hyperlink{ref-chen2024mirage}{L. Y. Chen et al., 2024}; \protect\hyperlink{ref-chen2024rovi}{L. Y. Chen et al., 2025})}
\expandafter\def\csname ESWAparenhejna2024remix,saxena2025what\endcsname{(\protect\hyperlink{ref-hejna2024remix}{Hejna et al., 2025}; \protect\hyperlink{ref-saxena2025what}{Saxena et al., 2025})}
\expandafter\def\csname ESWAparenargall2009survey\endcsname{(\protect\hyperlink{ref-argall2009survey}{Argall et al., 2009})}
\expandafter\def\csname ESWAparenduan2017oneshot,finn2017oneshot,james2018taskembedded\endcsname{(\protect\hyperlink{ref-duan2017oneshot}{Duan et al., 2017}; \protect\hyperlink{ref-finn2017oneshot}{Finn, Yu, et al., 2017}; \protect\hyperlink{ref-james2018taskembedded}{James et al., 2018})}
\expandafter\def\csname ESWAparenyu2018oneshot\endcsname{(\protect\hyperlink{ref-yu2018oneshot}{T. Yu et al., 2018})}
\expandafter\def\csname ESWAparentorabi2018behavioral\endcsname{(\protect\hyperlink{ref-torabi2018behavioral}{Torabi et al., 2018})}
\expandafter\def\csname ESWAparenzakka2021xirl\endcsname{(\protect\hyperlink{ref-zakka2021xirl}{Zakka et al., 2022})}
\expandafter\def\csname ESWAparenbahl2022human,shaw2022videodex,wang2023mimicplay\endcsname{(\protect\hyperlink{ref-bahl2022human}{Bahl et al., 2022}; \protect\hyperlink{ref-shaw2022videodex}{Shaw et al., 2023}; \protect\hyperlink{ref-wang2023mimicplay}{C. Wang et al., 2023})}
\expandafter\def\csname ESWAparenbahl2023affordances\endcsname{(\protect\hyperlink{ref-bahl2023affordances}{Bahl et al., 2023})}
\expandafter\def\csname ESWAparenjain2024vid2robot\endcsname{(\protect\hyperlink{ref-jain2024vid2robot}{Jain et al., 2024})}
\expandafter\def\csname ESWAparenkareer2024egomimic\endcsname{(\protect\hyperlink{ref-kareer2024egomimic}{Kareer et al., 2025})}
\expandafter\def\csname ESWAparenlepert2025phantom\endcsname{(\protect\hyperlink{ref-lepert2025phantom}{Lepert et al., 2025})}
\expandafter\def\csname ESWAparenzhao2023learning,fu2024mobile\endcsname{(\protect\hyperlink{ref-fu2024mobile}{Fu, Zhao, \& Finn, 2025}; \protect\hyperlink{ref-zhao2023learning}{Zhao et al., 2023})}
\expandafter\def\csname ESWAparenchi2024universal\endcsname{(\protect\hyperlink{ref-chi2024universal}{Chi et al., 2024})}
\expandafter\def\csname ESWAparenwang2024dexcap\endcsname{(\protect\hyperlink{ref-wang2024dexcap}{C. Wang et al., 2024})}
\expandafter\def\csname ESWAparenchi2023diffusion\endcsname{(\protect\hyperlink{ref-chi2023diffusion}{Chi et al., 2025})}
\expandafter\def\csname ESWAparenpeng2020learning\endcsname{(\protect\hyperlink{ref-peng2020learning}{Peng et al., 2020})}
\expandafter\def\csname ESWAparencheng2024expressive\endcsname{(\protect\hyperlink{ref-cheng2024expressive}{Cheng et al., 2024})}
\expandafter\def\csname ESWAparenfu2024humanplus\endcsname{(\protect\hyperlink{ref-fu2024humanplus}{Fu, Zhao, Wu, et al., 2025})}
\expandafter\def\csname ESWAparenli2024okami\endcsname{(\protect\hyperlink{ref-li2024okami}{Li et al., 2025a})}
\expandafter\def\csname ESWAparenqiu2025humanoid\endcsname{(\protect\hyperlink{ref-qiu2025humanoid}{Qiu et al., 2025})}
\expandafter\def\csname ESWAparenmandlekar2023mimicgen,jiang2024dexmimicgen\endcsname{(\protect\hyperlink{ref-jiang2024dexmimicgen}{Z. Jiang et al., 2025}; \protect\hyperlink{ref-mandlekar2023mimicgen}{Mandlekar et al., 2023})}
\expandafter\def\csname ESWAparenjang2025dreamgen\endcsname{(\protect\hyperlink{ref-jang2025dreamgen}{J. Jang et al., 2025})}
\expandafter\def\csname ESWAparenhuang2022language\endcsname{(\protect\hyperlink{ref-huang2022language}{Huang et al., 2022})}
\expandafter\def\csname ESWAparenahn2022saycan\endcsname{(\protect\hyperlink{ref-ahn2022saycan}{Ichter et al., 2023})}
\expandafter\def\csname ESWAparendriess2023palme\endcsname{(\protect\hyperlink{ref-driess2023palme}{Driess et al., 2023})}
\expandafter\def\csname ESWAparenhuang2022inner,liang2023code,huang2023voxposer\endcsname{(\protect\hyperlink{ref-huang2022inner}{Huang, Xia, et al., 2023}; \protect\hyperlink{ref-huang2023voxposer}{Huang, Wang, et al., 2023}; \protect\hyperlink{ref-liang2023code}{Liang et al., 2023})}
\expandafter\def\csname ESWAparenparisi2022unsurprising\endcsname{(\protect\hyperlink{ref-parisi2022unsurprising}{Parisi et al., 2022})}
\expandafter\def\csname ESWAparennair2022r3m\endcsname{(\protect\hyperlink{ref-nair2022r3m}{S. Nair et al., 2023})}
\expandafter\def\csname ESWAparenmajumdar2023where\endcsname{(\protect\hyperlink{ref-majumdar2023where}{Majumdar et al., 2023})}
\expandafter\def\csname ESWAparenradosavovic2022realworld,ma2023vip,karamcheti2023languagedriven,shridhar2021cliport,shen2023distilled\endcsname{(\protect\hyperlink{ref-karamcheti2023languagedriven}{Karamcheti et al., 2023}; \protect\hyperlink{ref-ma2023vip}{Ma et al., 2023}; \protect\hyperlink{ref-radosavovic2022realworld}{Radosavovic et al., 2023}; \protect\hyperlink{ref-shen2023distilled}{Shen et al., 2023}; \protect\hyperlink{ref-shridhar2021cliport}{Shridhar et al., 2022})}
\expandafter\def\csname ESWAparenhansen2023pretraining\endcsname{(\protect\hyperlink{ref-hansen2023pretraining}{Hansen et al., 2023})}
\expandafter\def\csname ESWAparendasari2023unbiased\endcsname{(\protect\hyperlink{ref-dasari2023unbiased}{Dasari et al., 2023})}
\expandafter\def\csname ESWAparenburns2024what\endcsname{(\protect\hyperlink{ref-burns2024what}{Burns et al., 2025})}
\expandafter\def\csname ESWAparenschneider2024surprising\endcsname{(\protect\hyperlink{ref-schneider2024surprising}{Schneider et al., 2024})}
\expandafter\def\csname ESWAparenshang2024theia,jiang2025robots\endcsname{(\protect\hyperlink{ref-jiang2025robots}{G. Jiang et al., 2025}; \protect\hyperlink{ref-shang2024theia}{Shang et al., 2025})}
\expandafter\def\csname ESWAparenshridhar2022peract,goyal2023rvt,ze2024dp3\endcsname{(\protect\hyperlink{ref-goyal2023rvt}{Goyal et al., 2023}; \protect\hyperlink{ref-shridhar2022peract}{Shridhar et al., 2023}; \protect\hyperlink{ref-ze2024dp3}{Ze et al., 2024})}
\expandafter\def\csname ESWAtexthe2026confidence\endcsname{Y. He (\protect\hyperlink{ref-he2026confidence}{2026})}
\expandafter\def\csname ESWAparenbjorck2025groot,figure2025helix\endcsname{(\protect\hyperlink{ref-figure2025helix}{Figure AI, 2025}; \protect\hyperlink{ref-bjorck2025groot}{NVIDIA, Bjorck, et al., 2025})}
\expandafter\def\csname ESWAparenghosh2024octo,liu2024rdt\endcsname{(\protect\hyperlink{ref-liu2024rdt}{S. Liu et al., 2025}; \protect\hyperlink{ref-ghosh2024octo}{Octo Model Team et al., 2024})}
\expandafter\def\csname ESWAparenbrohan2023rt2\endcsname{(\protect\hyperlink{ref-brohan2023rt2}{Zitkovich et al., 2023})}
\expandafter\def\csname ESWAparenkim2024openvla\endcsname{(\protect\hyperlink{ref-kim2024openvla}{Kim, Pertsch, et al., 2025})}
\expandafter\def\csname ESWAparenblack2024pi0\endcsname{(\protect\hyperlink{ref-black2024pi0}{Black et al., 2025})}
\expandafter\def\csname ESWAparenpertsch2025fast\endcsname{(\protect\hyperlink{ref-pertsch2025fast}{Pertsch et al., 2025})}
\expandafter\def\csname ESWAparenkim2025oft\endcsname{(\protect\hyperlink{ref-kim2025oft}{Kim, Finn, et al., 2025})}
\expandafter\def\csname ESWAparenqu2025spatialvla\endcsname{(\protect\hyperlink{ref-qu2025spatialvla}{Qu et al., 2025})}
\expandafter\def\csname ESWAparenli2026robovlm\endcsname{(\protect\hyperlink{ref-li2026robovlm}{Li et al., 2026})}
\expandafter\def\csname ESWAparenzawalski2024ecot\endcsname{(\protect\hyperlink{ref-zawalski2024ecot}{Zawalski et al., 2025})}
\expandafter\def\csname ESWAparenzhen20243dvla\endcsname{(\protect\hyperlink{ref-zhen20243dvla}{Zhen et al., 2024})}
\expandafter\def\csname ESWAparenshukor2025smolvla\endcsname{(\protect\hyperlink{ref-shukor2025smolvla}{Shukor et al., 2025})}
\expandafter\def\csname ESWAparenbjorck2025groot\endcsname{(\protect\hyperlink{ref-bjorck2025groot}{NVIDIA, Bjorck, et al., 2025})}
\expandafter\def\csname ESWAparengemini2025robotics,gemini2025robotics15\endcsname{(\protect\hyperlink{ref-gemini2025robotics}{Gemini Robotics Team, Abeyruwan, et al., 2025}; \protect\hyperlink{ref-gemini2025robotics15}{Gemini Robotics Team, Abdolmaleki, et al., 2025})}
\expandafter\def\csname ESWAparenfigure2025helix\endcsname{(\protect\hyperlink{ref-figure2025helix}{Figure AI, 2025})}
\expandafter\def\csname ESWAparenblack2025pi05\endcsname{(\protect\hyperlink{ref-black2025pi05}{Physical Intelligence, Black, et al., 2025})}
\expandafter\def\csname ESWAparenai2026pi07\endcsname{(\protect\hyperlink{ref-ai2026pi07}{Physical Intelligence et al., 2026})}
\expandafter\def\csname ESWAparenzhou2025chatvla\endcsname{(\protect\hyperlink{ref-zhou2025chatvla}{Zhou, Zhu, et al., 2025})}
\expandafter\def\csname ESWAparendriess2025knowledge\endcsname{(\protect\hyperlink{ref-driess2025knowledge}{Driess et al., 2025})}
\expandafter\def\csname ESWAparentan2025ript\endcsname{(\protect\hyperlink{ref-tan2025ript}{S. Tan et al., 2025})}
\expandafter\def\csname ESWAparenamin2025pistar\endcsname{(\protect\hyperlink{ref-amin2025pistar}{Physical Intelligence, Amin, et al., 2025})}
\expandafter\def\csname ESWAparendu2023learning\endcsname{(\protect\hyperlink{ref-du2023learning}{Du et al., 2023})}
\expandafter\def\csname ESWAparenwu2024unleashing\endcsname{(\protect\hyperlink{ref-wu2024unleashing}{H. Wu et al., 2024})}
\expandafter\def\csname ESWAparencheang2024gr2\endcsname{(\protect\hyperlink{ref-cheang2024gr2}{Cheang et al., 2024})}
\expandafter\def\csname ESWAparenyang2023unisim,agarwal2025cosmos\endcsname{(\protect\hyperlink{ref-agarwal2025cosmos}{NVIDIA, Agarwal, et al., 2025}; \protect\hyperlink{ref-yang2023unisim}{S. Yang et al., 2024})}
\expandafter\def\csname ESWAparenassran2025vjepa\endcsname{(\protect\hyperlink{ref-assran2025vjepa}{Assran et al., 2025})}
\expandafter\def\csname ESWAparenzhou2025dinowm\endcsname{(\protect\hyperlink{ref-zhou2025dinowm}{G. Zhou et al., 2025})}
\expandafter\def\csname ESWAparenhu2025video\endcsname{(\protect\hyperlink{ref-hu2025video}{Hu et al., 2025})}
\expandafter\def\csname ESWAparenbrohan2023rt2,ahn2022saycan\endcsname{(\protect\hyperlink{ref-ahn2022saycan}{Ichter et al., 2023}; \protect\hyperlink{ref-brohan2023rt2}{Zitkovich et al., 2023})}
\expandafter\def\csname ESWAparenkim2024openvla,kim2025oft,pertsch2025fast\endcsname{(\protect\hyperlink{ref-kim2024openvla}{Kim, Pertsch, et al., 2025}; \protect\hyperlink{ref-kim2025oft}{Kim, Finn, et al., 2025}; \protect\hyperlink{ref-pertsch2025fast}{Pertsch et al., 2025})}
\expandafter\def\csname ESWAparenloquercio2020deep,loquercio2021learning\endcsname{(\protect\hyperlink{ref-loquercio2020deep}{Loquercio et al., 2020}; \protect\hyperlink{ref-loquercio2021learning}{Loquercio et al., 2021})}
\expandafter\def\csname ESWAparensong2023reaching\endcsname{(\protect\hyperlink{ref-song2023reaching}{Song et al., 2023})}
\expandafter\def\csname ESWApareneschmann2026raptor\endcsname{(\protect\hyperlink{ref-eschmann2026raptor}{Eschmann et al., 2026})}
\expandafter\def\csname ESWAparenli2018watergan\endcsname{(\protect\hyperlink{ref-li2018watergan}{Li et al., 2018})}
\expandafter\def\csname ESWAparenchaffre2025sim\endcsname{(\protect\hyperlink{ref-chaffre2025sim}{Chaffre et al., 2025})}
\expandafter\def\csname ESWAparencarlucho2018adaptive\endcsname{(\protect\hyperlink{ref-carlucho2018adaptive}{Carlucho et al., 2018})}
\expandafter\def\csname ESWAparenthuruthel2019model,thuruthel2019soft\endcsname{(\protect\hyperlink{ref-thuruthel2019model}{Thuruthel, Falotico, et al., 2019}; \protect\hyperlink{ref-thuruthel2019soft}{Thuruthel, Shih, et al., 2019})}
\expandafter\def\csname ESWAparenschaff2023sim\endcsname{(\protect\hyperlink{ref-schaff2023sim}{Schaff et al., 2023})}
\expandafter\def\csname ESWAparenchen2017decentralized,long2018towards\endcsname{(\protect\hyperlink{ref-chen2017decentralized}{Y. F. Chen, Liu, et al., 2017}; \protect\hyperlink{ref-long2018towards}{Long et al., 2018})}
\expandafter\def\csname ESWAparenchen2017socially,fan2020distributed\endcsname{(\protect\hyperlink{ref-chen2017socially}{Y. F. Chen, Everett, et al., 2017}; \protect\hyperlink{ref-fan2020distributed}{Fan et al., 2020})}
\expandafter\def\csname ESWAparenyu2019metaworld,james2020rlbench,mees2022calvin,liu2023libero,tao2024maniskill3\endcsname{(\protect\hyperlink{ref-james2020rlbench}{James et al., 2020}; \protect\hyperlink{ref-liu2023libero}{B. Liu et al., 2023}; \protect\hyperlink{ref-mees2022calvin}{Mees et al., 2022}; \protect\hyperlink{ref-tao2024maniskill3}{Tao et al., 2025}; \protect\hyperlink{ref-yu2019metaworld}{T. Yu et al., 2019})}
\expandafter\def\csname ESWAparenpumacay2024colosseum\endcsname{(\protect\hyperlink{ref-pumacay2024colosseum}{Pumacay et al., 2024})}
\expandafter\def\csname ESWAparengao2025taxonomy\endcsname{(\protect\hyperlink{ref-gao2025taxonomy}{Gao et al., 2026})}
\expandafter\def\csname ESWAtexthe2026observation\endcsname{Y. He, Wang, et al. (\protect\hyperlink{ref-he2026observation}{2026b})}
\expandafter\def\csname ESWAtexthe2026benchmark\endcsname{Y. He, Wang, et al. (\protect\hyperlink{ref-he2026benchmark}{2026a})}
\expandafter\def\csname ESWAparenli2024simpler\endcsname{(\protect\hyperlink{ref-li2024simpler}{Li et al., 2025b})}
\expandafter\def\csname ESWAparenzhou2025autoeval\endcsname{(\protect\hyperlink{ref-zhou2025autoeval}{Zhou, Atreya, et al., 2025})}
\expandafter\def\csname ESWAparenatreya2025roboarena\endcsname{(\protect\hyperlink{ref-atreya2025roboarena}{Atreya et al., 2025})}
\expandafter\def\csname ESWAparenvincent2024how\endcsname{(\protect\hyperlink{ref-vincent2024how}{Vincent et al., 2024})}
\expandafter\def\csname ESWAparenkurin2021morphology\endcsname{(\protect\hyperlink{ref-kurin2021morphology}{Kurin et al., 2021})}
\expandafter\def\csname ESWAparenlin2025data,saxena2025what\endcsname{(\protect\hyperlink{ref-lin2025data}{Lin et al., 2025}; \protect\hyperlink{ref-saxena2025what}{Saxena et al., 2025})}
\expandafter\def\csname ESWAparentobin2017domain,akkaya2019solving\endcsname{(\protect\hyperlink{ref-akkaya2019solving}{OpenAI et al., 2019}; \protect\hyperlink{ref-tobin2017domain}{Tobin et al., 2017})}
\expandafter\def\csname ESWAparenyang2026data\endcsname{(\protect\hyperlink{ref-yang2026data}{J. Yang et al., 2026})}
\expandafter\def\csname ESWAparenhejna2024remix,xing2025shortcut\endcsname{(\protect\hyperlink{ref-hejna2024remix}{Hejna et al., 2025}; \protect\hyperlink{ref-xing2025shortcut}{Xing et al., 2025})}
\expandafter\def\csname ESWAparenshah2023gnm,shah2023vint\endcsname{(\protect\hyperlink{ref-shah2023gnm}{Shah, Sridhar, Bhorkar, et al., 2023}; \protect\hyperlink{ref-shah2023vint}{Shah, Sridhar, Dashora, et al., 2023})}
\expandafter\def\csname ESWAparenchen2024mirage,ye2024latent\endcsname{(\protect\hyperlink{ref-chen2024mirage}{L. Y. Chen et al., 2024}; \protect\hyperlink{ref-ye2024latent}{Ye et al., 2025})}
\expandafter\def\csname ESWAparenchi2024universal,zhao2023learning\endcsname{(\protect\hyperlink{ref-chi2024universal}{Chi et al., 2024}; \protect\hyperlink{ref-zhao2023learning}{Zhao et al., 2023})}
\expandafter\def\csname ESWAparenmajumdar2023where,hansen2023pretraining,dasari2023unbiased\endcsname{(\protect\hyperlink{ref-dasari2023unbiased}{Dasari et al., 2023}; \protect\hyperlink{ref-hansen2023pretraining}{Hansen et al., 2023}; \protect\hyperlink{ref-majumdar2023where}{Majumdar et al., 2023})}
\expandafter\def\csname ESWAparenhwangbo2019learning,kumar2021rma\endcsname{(\protect\hyperlink{ref-hwangbo2019learning}{Hwangbo et al., 2019}; \protect\hyperlink{ref-kumar2021rma}{Kumar et al., 2021})}
\expandafter\def\csname ESWAparenghosh2024octo,kim2025oft\endcsname{(\protect\hyperlink{ref-kim2025oft}{Kim, Finn, et al., 2025}; \protect\hyperlink{ref-ghosh2024octo}{Octo Model Team et al., 2024})}
\expandafter\def\csname ESWAparenluo2025precise,yuan2025policy\endcsname{(\protect\hyperlink{ref-luo2025precise}{Luo et al., 2025}; \protect\hyperlink{ref-yuan2025policy}{Yuan et al., 2025})}
\expandafter\def\csname ESWAparenchen2024mirage,torne2024reconciling\endcsname{(\protect\hyperlink{ref-chen2024mirage}{L. Y. Chen et al., 2024}; \protect\hyperlink{ref-torne2024reconciling}{Torne et al., 2024})}
\expandafter\def\csname ESWAparenye2024latent,bu2025univla\endcsname{(\protect\hyperlink{ref-bu2025univla}{Bu et al., 2025}; \protect\hyperlink{ref-ye2024latent}{Ye et al., 2025})}
\expandafter\def\csname ESWAparenqiu2025humanoid,bahl2023affordances,chi2024universal\endcsname{(\protect\hyperlink{ref-bahl2023affordances}{Bahl et al., 2023}; \protect\hyperlink{ref-chi2024universal}{Chi et al., 2024}; \protect\hyperlink{ref-qiu2025humanoid}{Qiu et al., 2025})}
\expandafter\def\csname ESWAparendriess2025knowledge,ahn2022saycan\endcsname{(\protect\hyperlink{ref-driess2025knowledge}{Driess et al., 2025}; \protect\hyperlink{ref-ahn2022saycan}{Ichter et al., 2023})}
\expandafter\def\csname ESWAparentorne2024reconciling,qureshi2025splatsim\endcsname{(\protect\hyperlink{ref-qureshi2025splatsim}{Qureshi et al., 2025}; \protect\hyperlink{ref-torne2024reconciling}{Torne et al., 2024})}
\expandafter\def\csname ESWAparending2024which\endcsname{(\protect\hyperlink{ref-ding2024which}{Ding et al., 2024})}
\expandafter\def\csname ESWAparenhejna2024remix,xing2025shortcut,yang2026data\endcsname{(\protect\hyperlink{ref-hejna2024remix}{Hejna et al., 2025}; \protect\hyperlink{ref-xing2025shortcut}{Xing et al., 2025}; \protect\hyperlink{ref-yang2026data}{J. Yang et al., 2026})}
\expandafter\def\csname ESWAparenlesort2020continual\endcsname{(\protect\hyperlink{ref-lesort2020continual}{Lesort et al., 2020})}
\expandafter\def\csname ESWAparenliu2023libero,zhou2025chatvla,driess2025knowledge\endcsname{(\protect\hyperlink{ref-driess2025knowledge}{Driess et al., 2025}; \protect\hyperlink{ref-liu2023libero}{B. Liu et al., 2023}; \protect\hyperlink{ref-zhou2025chatvla}{Zhou, Zhu, et al., 2025})}
\expandafter\def\csname ESWAparenfang2023rh20t,qi2023general\endcsname{(\protect\hyperlink{ref-fang2023rh20t}{Fang et al., 2024}; \protect\hyperlink{ref-qi2023general}{Qi, Yi, et al., 2023})}
\expandafter\def\csname ESWAparenyang2023unisim,assran2025vjepa\endcsname{(\protect\hyperlink{ref-assran2025vjepa}{Assran et al., 2025}; \protect\hyperlink{ref-yang2023unisim}{S. Yang et al., 2024})}
\expandafter\def\csname ESWAparenhsu2023simtolabtoreal\endcsname{(\protect\hyperlink{ref-hsu2023simtolabtoreal}{Hsu et al., 2023})}
\expandafter\def\csname ESWAparengemini2025robotics15,figure2025helix,ai2026pi07\endcsname{(\protect\hyperlink{ref-figure2025helix}{Figure AI, 2025}; \protect\hyperlink{ref-gemini2025robotics15}{Gemini Robotics Team, Abdolmaleki, et al., 2025}; \protect\hyperlink{ref-ai2026pi07}{Physical Intelligence et al., 2026})}
\expandafter\def\csname ESWAparenatreya2025roboarena,zhou2025autoeval,li2024simpler\endcsname{(\protect\hyperlink{ref-atreya2025roboarena}{Atreya et al., 2025}; \protect\hyperlink{ref-li2024simpler}{Li et al., 2025b}; \protect\hyperlink{ref-zhou2025autoeval}{Zhou, Atreya, et al., 2025})}
\expandafter\def\csname ESWAtextandrychowicz2020learning\endcsname{Andrychowicz et al. (\protect\hyperlink{ref-andrychowicz2020learning}{2020})}
\expandafter\def\csname ESWAtexthanda2023dextreme\endcsname{Handa et al. (\protect\hyperlink{ref-handa2023dextreme}{2023})}
\expandafter\def\csname ESWAtextkaufmann2023champion\endcsname{Kaufmann et al. (\protect\hyperlink{ref-kaufmann2023champion}{2023})}
\expandafter\def\csname ESWAtextakkaya2019solving\endcsname{OpenAI et al. (\protect\hyperlink{ref-akkaya2019solving}{2019})}
\expandafter\def\csname ESWAtextbousmalis2018using\endcsname{Bousmalis et al. (\protect\hyperlink{ref-bousmalis2018using}{2018})}
\expandafter\def\csname ESWAtextchaffre2025sim\endcsname{Chaffre et al. (\protect\hyperlink{ref-chaffre2025sim}{2025})}
\expandafter\def\csname ESWAtextchebotar2019closing\endcsname{Chebotar et al. (\protect\hyperlink{ref-chebotar2019closing}{2019})}
\expandafter\def\csname ESWAtextfan2020distributed\endcsname{Fan et al. (\protect\hyperlink{ref-fan2020distributed}{2020})}
\expandafter\def\csname ESWAtexthe2025asap\endcsname{T. He, Gao, et al. (\protect\hyperlink{ref-he2025asap}{2025})}
\expandafter\def\csname ESWAtextkumar2021rma\endcsname{Kumar et al. (\protect\hyperlink{ref-kumar2021rma}{2021})}
\expandafter\def\csname ESWAtextli2018watergan\endcsname{Li et al. (\protect\hyperlink{ref-li2018watergan}{2018})}
\expandafter\def\csname ESWAtextloquercio2020deep\endcsname{Loquercio et al. (\protect\hyperlink{ref-loquercio2020deep}{2020})}
\expandafter\def\csname ESWAtextloquercio2021learning\endcsname{Loquercio et al. (\protect\hyperlink{ref-loquercio2021learning}{2021})}
\expandafter\def\csname ESWAtextma2024dreureka\endcsname{Ma, Liang, Wang, Wang, et al. (\protect\hyperlink{ref-ma2024dreureka}{2024})}
\expandafter\def\csname ESWAtextmuratore2021data\endcsname{Muratore et al. (\protect\hyperlink{ref-muratore2021data}{2021})}
\expandafter\def\csname ESWAtextpinto2017robust\endcsname{Pinto et al. (\protect\hyperlink{ref-pinto2017robust}{2017})}
\expandafter\def\csname ESWAtextqi2022inhand\endcsname{Qi, Kumar, et al. (\protect\hyperlink{ref-qi2022inhand}{2023})}
\expandafter\def\csname ESWAtextqi2023general\endcsname{Qi, Yi, et al. (\protect\hyperlink{ref-qi2023general}{2023})}
\expandafter\def\csname ESWAtextqureshi2025splatsim\endcsname{Qureshi et al. (\protect\hyperlink{ref-qureshi2025splatsim}{2025})}
\expandafter\def\csname ESWAtextradosavovic2024real\endcsname{Radosavovic et al. (\protect\hyperlink{ref-radosavovic2024real}{2024})}
\expandafter\def\csname ESWAtextrajeswaran2017epopt\endcsname{Rajeswaran et al. (\protect\hyperlink{ref-rajeswaran2017epopt}{2017})}
\expandafter\def\csname ESWAtextrusu2017sim\endcsname{Rusu et al. (\protect\hyperlink{ref-rusu2017sim}{2017})}
\expandafter\def\csname ESWAtextsong2024learning\endcsname{Song et al. (\protect\hyperlink{ref-song2024learning}{2025})}
\expandafter\def\csname ESWAtexttorne2024reconciling\endcsname{Torne et al. (\protect\hyperlink{ref-torne2024reconciling}{2024})}
\expandafter\def\csname ESWAtextzhuang2023robot\endcsname{Zhuang et al. (\protect\hyperlink{ref-zhuang2023robot}{2023})}
\expandafter\def\csname ESWAtextschaff2023sim\endcsname{Schaff et al. (\protect\hyperlink{ref-schaff2023sim}{2023})}
\expandafter\def\csname ESWAtextthuruthel2019model\endcsname{Thuruthel, Falotico, et al. (\protect\hyperlink{ref-thuruthel2019model}{2019})}
\expandafter\def\csname ESWAtextthuruthel2019soft\endcsname{Thuruthel, Shih, et al. (\protect\hyperlink{ref-thuruthel2019soft}{2019})}
\expandafter\def\csname ESWAtextyang2023unisim\endcsname{S. Yang et al. (\protect\hyperlink{ref-yang2023unisim}{2024})}
\expandafter\def\csname ESWAtextchen2017socially\endcsname{Y. F. Chen, Everett, et al. (\protect\hyperlink{ref-chen2017socially}{2017})}
\expandafter\def\csname ESWAtextjames2019sim\endcsname{James et al. (\protect\hyperlink{ref-james2019sim}{2019})}
\expandafter\def\csname ESWAtextchi2023diffusion\endcsname{Chi et al. (\protect\hyperlink{ref-chi2023diffusion}{2025})}
\expandafter\def\csname ESWAtextcully2015robots\endcsname{Cully et al. (\protect\hyperlink{ref-cully2015robots}{2015})}
\expandafter\def\csname ESWAtextfinn2017model\endcsname{Finn, Abbeel, et al. (\protect\hyperlink{ref-finn2017model}{2017})}
\expandafter\def\csname ESWAtextma2024eureka\endcsname{Ma, Liang, Wang, Huang, et al. (\protect\hyperlink{ref-ma2024eureka}{2024})}
\expandafter\def\csname ESWAtextnichol2018first\endcsname{Nichol et al. (\protect\hyperlink{ref-nichol2018first}{2018})}
\expandafter\def\csname ESWAtexttouati2023does\endcsname{Touati et al. (\protect\hyperlink{ref-touati2023does}{2023})}
\expandafter\def\csname ESWAtextball2023efficient\endcsname{Ball et al. (\protect\hyperlink{ref-ball2023efficient}{2023})}
\expandafter\def\csname ESWAtextbarreto2017successor\endcsname{Barreto et al. (\protect\hyperlink{ref-barreto2017successor}{2017})}
\expandafter\def\csname ESWAtextbrohan2022rt1\endcsname{Brohan et al. (\protect\hyperlink{ref-brohan2022rt1}{2023})}
\expandafter\def\csname ESWAtextduan2016rl\endcsname{Duan et al. (\protect\hyperlink{ref-duan2016rl}{2016})}
\expandafter\def\csname ESWAtexteysenbach2019diversity\endcsname{Eysenbach et al. (\protect\hyperlink{ref-eysenbach2019diversity}{2019})}
\expandafter\def\csname ESWAtextfu2024mobile\endcsname{Fu, Zhao, and Finn (\protect\hyperlink{ref-fu2024mobile}{2025})}
\expandafter\def\csname ESWAtextgoyal2023rvt\endcsname{Goyal et al. (\protect\hyperlink{ref-goyal2023rvt}{2023})}
\expandafter\def\csname ESWAtexthe2025hover\endcsname{T. He, Xiao, et al. (\protect\hyperlink{ref-he2025hover}{2025})}
\expandafter\def\csname ESWAtexthiguera2024sparsh\endcsname{Higuera et al. (\protect\hyperlink{ref-higuera2024sparsh}{2025})}
\expandafter\def\csname ESWAtextjames2018taskembedded\endcsname{James et al. (\protect\hyperlink{ref-james2018taskembedded}{2018})}
\expandafter\def\csname ESWAtextjohannink2019residual\endcsname{Johannink et al. (\protect\hyperlink{ref-johannink2019residual}{2019})}
\expandafter\def\csname ESWAtextkalashnikov2021mt\endcsname{Kalashnikov et al. (\protect\hyperlink{ref-kalashnikov2021mt}{2021})}
\expandafter\def\csname ESWAtextlaskin2023incontext\endcsname{Laskin et al. (\protect\hyperlink{ref-laskin2023incontext}{2023})}
\expandafter\def\csname ESWAtextluo2024serl\endcsname{Luo et al. (\protect\hyperlink{ref-luo2024serl}{2024})}
\expandafter\def\csname ESWAtextluo2025precise\endcsname{Luo et al. (\protect\hyperlink{ref-luo2025precise}{2025})}
\expandafter\def\csname ESWAtextnagabandi2019learning\endcsname{Nagabandi et al. (\protect\hyperlink{ref-nagabandi2019learning}{2019})}
\expandafter\def\csname ESWAtextnair2020awac\endcsname{A. Nair et al. (\protect\hyperlink{ref-nair2020awac}{2020})}
\expandafter\def\csname ESWAtextnakamoto2023calql\endcsname{Nakamoto et al. (\protect\hyperlink{ref-nakamoto2023calql}{2023})}
\expandafter\def\csname ESWAtextpertsch2020accelerating\endcsname{Pertsch et al. (\protect\hyperlink{ref-pertsch2020accelerating}{2020})}
\expandafter\def\csname ESWAtextrakelly2019efficient\endcsname{Rakelly et al. (\protect\hyperlink{ref-rakelly2019efficient}{2019})}
\expandafter\def\csname ESWAtextren2024dppo\endcsname{Ren et al. (\protect\hyperlink{ref-ren2024dppo}{2025})}
\expandafter\def\csname ESWAtextrusu2016policy\endcsname{Rusu, Colmenarejo, et al. (\protect\hyperlink{ref-rusu2016policy}{2016})}
\expandafter\def\csname ESWAtextrusu2016progressive\endcsname{Rusu, Rabinowitz, et al. (\protect\hyperlink{ref-rusu2016progressive}{2016})}
\expandafter\def\csname ESWAtextshridhar2022peract\endcsname{Shridhar et al. (\protect\hyperlink{ref-shridhar2022peract}{2023})}
\expandafter\def\csname ESWAtextsong2023reaching\endcsname{Song et al. (\protect\hyperlink{ref-song2023reaching}{2023})}
\expandafter\def\csname ESWAtexttan2025ript\endcsname{S. Tan et al. (\protect\hyperlink{ref-tan2025ript}{2025})}
\expandafter\def\csname ESWAtexttirinzoni2025zero\endcsname{Tirinzoni et al. (\protect\hyperlink{ref-tirinzoni2025zero}{2025})}
\expandafter\def\csname ESWAtexttouati2021learning\endcsname{Touati and Ollivier (\protect\hyperlink{ref-touati2021learning}{2021})}
\expandafter\def\csname ESWAtextze2024dp3\endcsname{Ze et al. (\protect\hyperlink{ref-ze2024dp3}{2024})}
\expandafter\def\csname ESWAtextzhou2025dinowm\endcsname{G. Zhou et al. (\protect\hyperlink{ref-zhou2025dinowm}{2025})}
\expandafter\def\csname ESWAtextamin2025pistar\endcsname{Physical Intelligence, Amin, et al. (\protect\hyperlink{ref-amin2025pistar}{2025})}
\expandafter\def\csname ESWAtextyuan2025policy\endcsname{Yuan et al. (\protect\hyperlink{ref-yuan2025policy}{2025})}
\expandafter\def\csname ESWAtextai2025towards\endcsname{Ai et al. (\protect\hyperlink{ref-ai2025towards}{2025})}
\expandafter\def\csname ESWAtextbousmalis2023robocat\endcsname{Bousmalis et al. (\protect\hyperlink{ref-bousmalis2023robocat}{2023})}
\expandafter\def\csname ESWAtextbu2025agibot\endcsname{AgiBot World Contributors et al. (\protect\hyperlink{ref-bu2025agibot}{2025})}
\expandafter\def\csname ESWAtextchen2017decentralized\endcsname{Y. F. Chen, Liu, et al. (\protect\hyperlink{ref-chen2017decentralized}{2017})}
\expandafter\def\csname ESWAtextchen2018hardware\endcsname{T. Chen et al. (\protect\hyperlink{ref-chen2018hardware}{2018})}
\expandafter\def\csname ESWAtextchen2024mirage\endcsname{L. Y. Chen et al. (\protect\hyperlink{ref-chen2024mirage}{2024})}
\expandafter\def\csname ESWAtextchen2024rovi\endcsname{L. Y. Chen et al. (\protect\hyperlink{ref-chen2024rovi}{2025})}
\expandafter\def\csname ESWAtextdasari2019robonet\endcsname{Dasari et al. (\protect\hyperlink{ref-dasari2019robonet}{2019})}
\expandafter\def\csname ESWAtextdoshi2024scaling\endcsname{Doshi et al. (\protect\hyperlink{ref-doshi2024scaling}{2025})}
\expandafter\def\csname ESWAtextghosh2024octo\endcsname{Octo Model Team et al. (\protect\hyperlink{ref-ghosh2024octo}{2024})}
\expandafter\def\csname ESWAtextgupta2022metamorph\endcsname{Gupta et al. (\protect\hyperlink{ref-gupta2022metamorph}{2022})}
\expandafter\def\csname ESWAtexthejna2024remix\endcsname{Hejna et al. (\protect\hyperlink{ref-hejna2024remix}{2025})}
\expandafter\def\csname ESWAtexthuang2020one\endcsname{Huang et al. (\protect\hyperlink{ref-huang2020one}{2020})}
\expandafter\def\csname ESWAtextlong2018towards\endcsname{Long et al. (\protect\hyperlink{ref-long2018towards}{2018})}
\expandafter\def\csname ESWAtextoneill2023open\endcsname{Open X-Embodiment Collaboration et al. (\protect\hyperlink{ref-oneill2023open}{2024})}
\expandafter\def\csname ESWAtextshah2023gnm\endcsname{Shah, Sridhar, Bhorkar, et al. (\protect\hyperlink{ref-shah2023gnm}{2023})}
\expandafter\def\csname ESWAtextshah2023vint\endcsname{Shah, Sridhar, Dashora, et al. (\protect\hyperlink{ref-shah2023vint}{2023})}
\expandafter\def\csname ESWAtextsridhar2024nomad\endcsname{Sridhar et al. (\protect\hyperlink{ref-sridhar2024nomad}{2024})}
\expandafter\def\csname ESWAtextwang2018nervenet\endcsname{T. Wang et al. (\protect\hyperlink{ref-wang2018nervenet}{2018})}
\expandafter\def\csname ESWAtextwang2024scaling\endcsname{L. Wang et al. (\protect\hyperlink{ref-wang2024scaling}{2024})}
\expandafter\def\csname ESWAtextyang2024pushing\endcsname{J. Yang et al. (\protect\hyperlink{ref-yang2024pushing}{2024})}
\expandafter\def\csname ESWAtextye2024latent\endcsname{Ye et al. (\protect\hyperlink{ref-ye2024latent}{2025})}
\expandafter\def\csname ESWAtextbu2025univla\endcsname{Bu et al. (\protect\hyperlink{ref-bu2025univla}{2025})}
\expandafter\def\csname ESWAtexteschmann2026raptor\endcsname{Eschmann et al. (\protect\hyperlink{ref-eschmann2026raptor}{2026})}
\expandafter\def\csname ESWAtextreed2022generalist\endcsname{Reed et al. (\protect\hyperlink{ref-reed2022generalist}{2022})}
\expandafter\def\csname ESWAtextzheng2025universal\endcsname{Zheng et al. (\protect\hyperlink{ref-zheng2025universal}{2025})}
\expandafter\def\csname ESWAtextjiang2024dexmimicgen\endcsname{Z. Jiang et al. (\protect\hyperlink{ref-jiang2024dexmimicgen}{2025})}
\expandafter\def\csname ESWAtextbahl2023affordances\endcsname{Bahl et al. (\protect\hyperlink{ref-bahl2023affordances}{2023})}
\expandafter\def\csname ESWAtextduan2017oneshot\endcsname{Duan et al. (\protect\hyperlink{ref-duan2017oneshot}{2017})}
\expandafter\def\csname ESWAtextfinn2017oneshot\endcsname{Finn, Yu, et al. (\protect\hyperlink{ref-finn2017oneshot}{2017})}
\expandafter\def\csname ESWAtextjain2024vid2robot\endcsname{Jain et al. (\protect\hyperlink{ref-jain2024vid2robot}{2024})}
\expandafter\def\csname ESWAtextjang2022bcz\endcsname{E. Jang et al. (\protect\hyperlink{ref-jang2022bcz}{2022})}
\expandafter\def\csname ESWAtextkareer2024egomimic\endcsname{Kareer et al. (\protect\hyperlink{ref-kareer2024egomimic}{2025})}
\expandafter\def\csname ESWAtextlepert2025phantom\endcsname{Lepert et al. (\protect\hyperlink{ref-lepert2025phantom}{2025})}
\expandafter\def\csname ESWAtextliu2024maniwav\endcsname{Z. Liu et al. (\protect\hyperlink{ref-liu2024maniwav}{2025})}
\expandafter\def\csname ESWAtextmandlekar2023mimicgen\endcsname{Mandlekar et al. (\protect\hyperlink{ref-mandlekar2023mimicgen}{2023})}
\expandafter\def\csname ESWAtextpeng2020learning\endcsname{Peng et al. (\protect\hyperlink{ref-peng2020learning}{2020})}
\expandafter\def\csname ESWAtextqiu2025humanoid\endcsname{Qiu et al. (\protect\hyperlink{ref-qiu2025humanoid}{2025})}
\expandafter\def\csname ESWAtextshaw2022videodex\endcsname{Shaw et al. (\protect\hyperlink{ref-shaw2022videodex}{2023})}
\expandafter\def\csname ESWAtexttorabi2018behavioral\endcsname{Torabi et al. (\protect\hyperlink{ref-torabi2018behavioral}{2018})}
\expandafter\def\csname ESWAtextwang2023mimicplay\endcsname{C. Wang et al. (\protect\hyperlink{ref-wang2023mimicplay}{2023})}
\expandafter\def\csname ESWAtextyu2018oneshot\endcsname{T. Yu et al. (\protect\hyperlink{ref-yu2018oneshot}{2018})}
\expandafter\def\csname ESWAtextzakka2021xirl\endcsname{Zakka et al. (\protect\hyperlink{ref-zakka2021xirl}{2022})}
\expandafter\def\csname ESWAtextzhao2023learning\endcsname{Zhao et al. (\protect\hyperlink{ref-zhao2023learning}{2023})}
\expandafter\def\csname ESWAtextbahl2022human\endcsname{Bahl et al. (\protect\hyperlink{ref-bahl2022human}{2022})}
\expandafter\def\csname ESWAtextcheng2024expressive\endcsname{Cheng et al. (\protect\hyperlink{ref-cheng2024expressive}{2024})}
\expandafter\def\csname ESWAtextchi2024universal\endcsname{Chi et al. (\protect\hyperlink{ref-chi2024universal}{2024})}
\expandafter\def\csname ESWAtextfu2024humanplus\endcsname{Fu, Zhao, Wu, et al. (\protect\hyperlink{ref-fu2024humanplus}{2025})}
\expandafter\def\csname ESWAtexthe2024learning\endcsname{T. He et al. (\protect\hyperlink{ref-he2024learning}{2024})}
\expandafter\def\csname ESWAtexthe2024omnih2o\endcsname{T. He, Luo, et al. (\protect\hyperlink{ref-he2024omnih2o}{2025})}
\expandafter\def\csname ESWAtextjang2025dreamgen\endcsname{J. Jang et al. (\protect\hyperlink{ref-jang2025dreamgen}{2025})}
\expandafter\def\csname ESWAtextli2024okami\endcsname{Li et al. (\protect\hyperlink{ref-li2024okami}{2025a})}
\expandafter\def\csname ESWAtextwang2024dexcap\endcsname{C. Wang et al. (\protect\hyperlink{ref-wang2024dexcap}{2024})}
\expandafter\def\csname ESWAtextyu2024mimictouch\endcsname{K. Yu et al. (\protect\hyperlink{ref-yu2024mimictouch}{2025})}
\expandafter\def\csname ESWAtextburns2024what\endcsname{Burns et al. (\protect\hyperlink{ref-burns2024what}{2025})}
\expandafter\def\csname ESWAtextdasari2023unbiased\endcsname{Dasari et al. (\protect\hyperlink{ref-dasari2023unbiased}{2023})}
\expandafter\def\csname ESWAtexthansen2023pretraining\endcsname{Hansen et al. (\protect\hyperlink{ref-hansen2023pretraining}{2023})}
\expandafter\def\csname ESWAtextmajumdar2023where\endcsname{Majumdar et al. (\protect\hyperlink{ref-majumdar2023where}{2023})}
\expandafter\def\csname ESWAtextahn2022saycan\endcsname{Ichter et al. (\protect\hyperlink{ref-ahn2022saycan}{2023})}
\expandafter\def\csname ESWAtextassran2025vjepa\endcsname{Assran et al. (\protect\hyperlink{ref-assran2025vjepa}{2025})}
\expandafter\def\csname ESWAtextbjorck2025groot\endcsname{NVIDIA, Bjorck, et al. (\protect\hyperlink{ref-bjorck2025groot}{2025})}
\expandafter\def\csname ESWAtextblack2024pi0\endcsname{Black et al. (\protect\hyperlink{ref-black2024pi0}{2025})}
\expandafter\def\csname ESWAtextblack2025pi05\endcsname{Physical Intelligence, Black, et al. (\protect\hyperlink{ref-black2025pi05}{2025})}
\expandafter\def\csname ESWAtextbrohan2023rt2\endcsname{Zitkovich et al. (\protect\hyperlink{ref-brohan2023rt2}{2023})}
\expandafter\def\csname ESWAtextdriess2023palme\endcsname{Driess et al. (\protect\hyperlink{ref-driess2023palme}{2023})}
\expandafter\def\csname ESWAtextdriess2025knowledge\endcsname{Driess et al. (\protect\hyperlink{ref-driess2025knowledge}{2025})}
\expandafter\def\csname ESWAtextdu2023learning\endcsname{Du et al. (\protect\hyperlink{ref-du2023learning}{2023})}
\expandafter\def\csname ESWAtextjiang2025robots\endcsname{G. Jiang et al. (\protect\hyperlink{ref-jiang2025robots}{2025})}
\expandafter\def\csname ESWAtextkaramcheti2023languagedriven\endcsname{Karamcheti et al. (\protect\hyperlink{ref-karamcheti2023languagedriven}{2023})}
\expandafter\def\csname ESWAtextkim2024openvla\endcsname{Kim, Pertsch, et al. (\protect\hyperlink{ref-kim2024openvla}{2025})}
\expandafter\def\csname ESWAtextkim2025oft\endcsname{Kim, Finn, et al. (\protect\hyperlink{ref-kim2025oft}{2025})}
\expandafter\def\csname ESWAtextli2026robovlm\endcsname{Li et al. (\protect\hyperlink{ref-li2026robovlm}{2026})}
\expandafter\def\csname ESWAtextliang2023code\endcsname{Liang et al. (\protect\hyperlink{ref-liang2023code}{2023})}
\expandafter\def\csname ESWAtextma2023vip\endcsname{Ma et al. (\protect\hyperlink{ref-ma2023vip}{2023})}
\expandafter\def\csname ESWAtextnair2022r3m\endcsname{S. Nair et al. (\protect\hyperlink{ref-nair2022r3m}{2023})}
\expandafter\def\csname ESWAtextparisi2022unsurprising\endcsname{Parisi et al. (\protect\hyperlink{ref-parisi2022unsurprising}{2022})}
\expandafter\def\csname ESWAtextpertsch2025fast\endcsname{Pertsch et al. (\protect\hyperlink{ref-pertsch2025fast}{2025})}
\expandafter\def\csname ESWAtextradosavovic2022realworld\endcsname{Radosavovic et al. (\protect\hyperlink{ref-radosavovic2022realworld}{2023})}
\expandafter\def\csname ESWAtextschneider2024surprising\endcsname{Schneider et al. (\protect\hyperlink{ref-schneider2024surprising}{2024})}
\expandafter\def\csname ESWAtextshang2024theia\endcsname{Shang et al. (\protect\hyperlink{ref-shang2024theia}{2025})}
\expandafter\def\csname ESWAtextshridhar2021cliport\endcsname{Shridhar et al. (\protect\hyperlink{ref-shridhar2021cliport}{2022})}
\expandafter\def\csname ESWAtextshukor2025smolvla\endcsname{Shukor et al. (\protect\hyperlink{ref-shukor2025smolvla}{2025})}
\expandafter\def\csname ESWAtextwu2024unleashing\endcsname{H. Wu et al. (\protect\hyperlink{ref-wu2024unleashing}{2024})}
\expandafter\def\csname ESWAtextzawalski2024ecot\endcsname{Zawalski et al. (\protect\hyperlink{ref-zawalski2024ecot}{2025})}
\expandafter\def\csname ESWAtextai2026pi07\endcsname{Physical Intelligence et al. (\protect\hyperlink{ref-ai2026pi07}{2026})}
\expandafter\def\csname ESWAtextcheang2024gr2\endcsname{Cheang et al. (\protect\hyperlink{ref-cheang2024gr2}{2024})}
\expandafter\def\csname ESWAtextgemini2025robotics\endcsname{Gemini Robotics Team, Abeyruwan, et al. (\protect\hyperlink{ref-gemini2025robotics}{2025})}
\expandafter\def\csname ESWAtextgemini2025robotics15\endcsname{Gemini Robotics Team, Abdolmaleki, et al. (\protect\hyperlink{ref-gemini2025robotics15}{2025})}
\expandafter\def\csname ESWAtexthu2025video\endcsname{Hu et al. (\protect\hyperlink{ref-hu2025video}{2025})}
\expandafter\def\csname ESWAtexthuang2022inner\endcsname{Huang, Xia, et al. (\protect\hyperlink{ref-huang2022inner}{2023})}
\expandafter\def\csname ESWAtexthuang2022language\endcsname{Huang et al. (\protect\hyperlink{ref-huang2022language}{2022})}
\expandafter\def\csname ESWAtexthuang2023voxposer\endcsname{Huang, Wang, et al. (\protect\hyperlink{ref-huang2023voxposer}{2023})}
\expandafter\def\csname ESWAtextliu2024rdt\endcsname{S. Liu et al. (\protect\hyperlink{ref-liu2024rdt}{2025})}
\expandafter\def\csname ESWAtextqu2025spatialvla\endcsname{Qu et al. (\protect\hyperlink{ref-qu2025spatialvla}{2025})}
\expandafter\def\csname ESWAtextshah2022lmnav\endcsname{Shah, Osiński, et al. (\protect\hyperlink{ref-shah2022lmnav}{2023})}
\expandafter\def\csname ESWAtextshen2023distilled\endcsname{Shen et al. (\protect\hyperlink{ref-shen2023distilled}{2023})}
\expandafter\def\csname ESWAtextzhen20243dvla\endcsname{Zhen et al. (\protect\hyperlink{ref-zhen20243dvla}{2024})}
\expandafter\def\csname ESWAtextzhou2025chatvla\endcsname{Zhou, Zhu, et al. (\protect\hyperlink{ref-zhou2025chatvla}{2025})}
\expandafter\def\csname ESWAtextfigure2025helix\endcsname{Figure AI (\protect\hyperlink{ref-figure2025helix}{2025})}

\hypersetup{hidelinks,bookmarksdepth=2,pdfauthor={Yuanzhi He, Jiayu Xu},pdftitle={What Transfers in Robot Learning?}}
\newenvironment{CSLReferences}[2]{\begingroup\small\setlength{\parindent}{0pt}\setlength{\leftskip}{0pt}\setlength{\parskip}{4pt}\everypar{\hangindent=1.27cm\hangafter=1}}{\par\endgroup}
\ExplSyntaxOn
\RenewDocumentEnvironment {Abstract} {o}
{
  \group_begin:
  \IfNoValueTF {#1} {} {\tex_gdef:D \abstractname {#1}}
  \parindent0pt
  \box_if_empty:NTF \g_stm_key_box
    {\leftskip=.35\textwidth}
    {
      \dim_gset:Nn \l_tmpa_dim {\box_ht:N \g_stm_key_box}
      \dim_gadd:Nn \l_tmpa_dim {\box_dp:N \g_stm_key_box}
      \leftskip=.35\textwidth
      \hspace*{-.35\textwidth}
      \noindent\hbox_to_wd:nn {0pt} {\box\g_stm_key_box\hss}
      \skip_vertical:n {-\l_tmpa_dim}
    }
  \noindent\abstractname\par
  \skip_vertical:n {-4pt}
  \noindent\rule{.65\textwidth}{.2pt}\par\footnotesize
  \ignorespaces\everypar{\parindent=1.5em}
}{\par\group_end:}
\cs_set:Npn \__first_footerline: {\small\itshape Preprint}
\ExplSyntaxOff
\renewcommand{\printorcid}{}
\begin{document}
\shorttitle{What Transfers in Robot Learning?}
\shortauthors{Y. He and J. Xu}
\title[mode=title]{What Transfers in Robot Learning? Taxonomy and Evidence from Domain Randomization to VLA Models}

\author[1]{Yuanzhi He}
\ead{HeY65@cardiff.ac.uk}
\author[1]{Jiayu Xu}
\cormark[1]
\ead{XuJ79@cardiff.ac.uk}
\affiliation[1]{organization={School of Computational and Mathematical Sciences, Cardiff University},
  addressline={Abacws, Senghennydd Road}, city={Cardiff}, postcode={CF24 4AG}, country={United Kingdom}}
\cortext[1]{Corresponding author.}

\begin{abstract}
Transfer learning allows robots to reuse experience when target-specific data are expensive. This survey organizes a curated corpus of 232 works through November 2026 along two axes: five source-to-target gaps (simulation to reality, task to task, embodiment to embodiment, human to robot, and web to robot) and five forms of transferred knowledge (data, representations, parameters and policies, models and simulators, and task conditioning). Each gap is reviewed through its mechanisms, representative evidence, and failure modes. A classical domain-adaptation bound provides a diagnostic vocabulary, with explicit limits for closed-loop control. We grade one focal transfer claim from each of 185 papers on an E0--E4 comparison-control scale and provide the complete coding. Within the coded subset, 145 claims (78.4\%) isolate the transferred artifact, while 15 (8.1\%) additionally report dispersion; no independent replication meeting the stated definition was identified in the corpus. These findings concern reported comparisons rather than field-wide evidence quality. \transferthesis{} Cross-gap synthesis links practical source selection to target-relevant coverage, compatible observation-action interfaces, matched adaptation budgets, and retention testing. Open problems include transferability prediction, harmful-mixture detection, physical validation of world models, and evaluation across laboratories.
\end{abstract}
\maketitle
\hypersetup{pdfauthor={Yuanzhi He, Jiayu Xu},pdfsubject={Robot transfer learning survey}}
\section{Introduction}
\label{sec:intro}

Robot learning depends on experience that is expensive to collect. RT-1 required 17 months and 13 robots for roughly 130{,}000 demonstration episodes \citep{brohan2022rt1}; Open X-Embodiment (OXE) and AgiBot World pool approximately one million trajectories each \citep{oneill2023open, bu2025agibot}. Reusing simulation, previous tasks, other robots, human demonstrations, or web pretraining is consequently central to learning under limited target-robot budgets. These sources differ in what they provide: a simulator supplies executable dynamics, a demonstration supplies behavior, and a vision-language backbone supplies features and semantic associations. Their usefulness cannot be inferred from source scale alone. For an intelligent robotic system, source selection is therefore a design decision that links data acquisition costs, adaptation requirements, and deployment reliability. A useful comparison must identify both the knowledge being reused and the target conditions under which reuse helps.

We examine which knowledge crosses each source-to-target gap and how the available evidence establishes its benefit. Classical transfer in reinforcement learning emphasized mappings, value functions, and sample-efficiency metrics \citep{taylor2009transfer, lazaric2012transfer}. Robot reinforcement learning treated the cost of physical experience as a central constraint \citep{kober2013reinforcement}. Deep robot learning developed gap-specific mechanisms, including domain randomization, meta-learning, and visual adaptation \citep{tobin2017domain, finn2017model, bousmalis2018using}. Recent generalist policies combine them through pretraining, target adaptation, and reinforcement-learning post-training \citep{oneill2023open, kim2024openvla, amin2025pistar}. \transferthesis

The benefits are conditional. Cross-robot pooling can improve specialist policies, yet mixture reweighting changes downstream success substantially \citep{oneill2023open, hejna2024remix}. Heterogeneous datasets can encourage source-identifying shortcuts \citep{xing2025shortcut}; tuned target-only baselines can match pretrained visual representations \citep{hansen2023pretraining, dasari2023unbiased}; action fine-tuning can erode the semantic abilities motivating a vision-language backbone \citep{zhou2025chatvla, driess2025knowledge}. These observations connect current scaling practice to negative transfer, transferability, and retention.

\begin{figure}[pos=tbp]
\centering
\includegraphics[width=\linewidth]{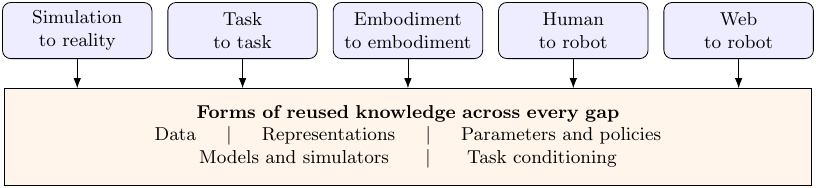}
\caption{Two axes for describing transfer. The gap identifies the source-target difference; the form identifies the reused artifact or interface. Each gap is reviewed in the main text. A system may combine several pathways}
\label{fig:taxonomy}
\Description{Five source-target gaps connect to a shared set of five knowledge forms.}
\end{figure}

Existing reviews provide complementary accounts. Sim-to-real surveys concentrate on the reality gap \citep{muratore2022robot}; transfer-learning reviews organize robot knowledge reuse and its challenges \citep{jaquier2025transfer}; foundation-model reviews cover architectures and robotic functions, including perception and interaction beyond control \citep{firoozi2025foundation}. Our contribution is to connect these accounts through two explicit axes (Fig.~\ref{fig:taxonomy}): five source-to-target gaps and five forms of transferred knowledge. Each gap receives a dedicated section, while shared failure mechanisms are synthesized across them. A second contribution is an E0--E4 rubric applied to one focal transfer claim from each of 185 papers, separating experimental control from uncertainty reporting and independent replication. A third is practical guidance that links source selection to target budget, interface compatibility, and an explicit comparison baseline.

\paragraph{Scope and collection.} The corpus contains 232 works from 1995 to November 2026. The main search closed in July 2026; a targeted November update added five works on task transfer, perception, and evaluation. The 185-paper evidence coding retains its July cutoff; the added works are discussed but were not added to the coded subset. It was assembled through structured keyword searches in arXiv, Semantic Scholar, and Google Scholar, searches of major robotics and machine-learning venues, and citation snowballing. Inclusion prioritized mechanisms relevant to robot control or perception-for-control, evidence testing those mechanisms, and evaluation instruments. Search stopped at mechanism saturation within each keyword family. This is a curated critical survey, not an exhaustive systematic review: database hit counts were not retained, and corpus proportions do not estimate field-wide frequencies. The main coverage concerns manipulation and locomotion; Section~\ref{sec:beyondmanip} examines selected aerial, underwater, soft, and multi-robot cases. Autonomous driving and language-only planning without a physical-action interface are outside scope. Supplementary Material S1 documents collection and coding; Supplementary Data S2 provides machine-readable records and reproduction code.

Section~\ref{sec:taxonomy} defines the framework. Sections~\ref{sec:sim2real}--\ref{sec:foundation} review the five gaps, Section~\ref{sec:eval} evaluates their evidence, and Sections~\ref{sec:patterns}--\ref{sec:challenges} develop the synthesis, guidance, and research agenda.

\section{Background and Taxonomy}
\label{sec:taxonomy}

\subsection{Transfer Problems and a Diagnostic Lens}
\label{sec:defs}

A robot-learning environment is $E=(\mathcal{O},\mathcal{S},\mathcal{A},P,\Omega,R)$: observations, states, actions, transition dynamics, observation model, and task specification. A transfer problem is $(\{(E_i,\mathcal{Z}_i)\}_{i=1}^{N},E_T,B)$, with source data or trained artifacts $\mathcal{Z}_i$, target environment $E_T$, and target-side budget $B$ in demonstrations, interactions, compute, or human time. Transfer uses source knowledge to improve target learning \citep{pan2010survey, taylor2009transfer}; negative transfer means doing worse than a target-only baseline under the same target budget \citep{wang2019characterizing, zhang2023survey}. Zero-shot deployment sets the adaptation budget to zero, although source collection and training still incur costs.

For binary prediction in a shared input-output space, a domain-adaptation bound separates three sources of target error \citep{bendavid2010theory}:
\begin{equation}
\begin{aligned}
\varepsilon_T(h) &\leq \varepsilon_S(h)+\tfrac{1}{2}d_{\mathcal{H}\Delta\mathcal{H}}(\mathcal{D}_S,\mathcal{D}_T)+\lambda^*,\\
\lambda^* &=\min_{h'\in\mathcal{H}}[\varepsilon_S(h')+\varepsilon_T(h')].
\end{aligned}
\label{eq:bound}
\end{equation}
Here the divergence measures source-target differences in the disagreement rates of hypothesis pairs, and $\lambda^*$ measures whether one hypothesis can perform well in both domains. The equation supplies a diagnostic vocabulary: visual shift suggests distribution mismatch; incompatible tasks suggest joint-hypothesis error; different bodies require an aligned observation-action interface before this comparison is meaningful. Changing data, representation, or conditioning can affect several terms simultaneously (Table~\ref{tab:forms}).

This is an organizing analogy for robot policies, not a control guarantee or computable transferability score. Closed-loop errors change future observations and compound over time \citep{ross2011reduction}; partial observability, continuous actions, and changing dynamics exceed the bound's assumptions. The discrepancy distance accommodates other losses \citep{mansour2009domain}, while the simulation lemma relates dynamics error to value error \citep{kearns2002near}. Invariant risk minimization seeks predictors whose optimal classifier is stable across environments \citep{arjovsky2019irm}. These alternatives reinforce the need to distinguish perceptual alignment from dynamical validity.

\subsection{Five Gaps and Five Forms}
\label{sec:gaps}

The \emph{simulation-to-reality gap} changes rendering or dynamics; the \emph{task gap} changes reward, goal, or demonstration specification; the \emph{embodiment gap} changes body, sensors, or action interface; the \emph{human-to-robot gap} maps human observations or motion into executable robot behavior; and the \emph{web-to-robot gap} grounds internet-scale representations and knowledge in control. Human transfer is reviewed separately because raw human video lacks robot action labels and contact measurements, although instrumented collection can supply motion or force information. Human and robot sources increasingly share unified interfaces. Scene, object, and layout changes are evaluation axes across these gaps. Offline-to-online learning is a budget and training-protocol change; it need not involve a new environment.

Five forms answer the independent question of what moves. \emph{Data} are replayed, reweighted, or co-trained. \emph{Representations} include visual features and latent action vocabularies. \emph{Parameters and policies} move through fine-tuning, distillation, or modular reuse. \emph{Models and simulators} generate predictions or synthetic experience. \emph{Task conditioning} reuses a fixed model through goals, language, demonstration prompts, morphology tokens, or interaction history. The first four connect established transfer categories \citep{pan2010survey, taylor2009transfer}; the fifth includes longstanding goal-conditioned reuse and its extension through large sequence models \citep{shah2023gnm, laskin2023incontext, brohan2023rt2}. Conditioning is an interface for reuse rather than evidence that a new learning mechanism lacks historical precedents.

The axes describe pathways, so a system can occupy several cells. For the census, each paper is assigned once: identify changed components of $E$; select the difference the contributed mechanism is designed to address; break remaining ties using the focal claim being graded. Apply the same principle to the carried artifact. Thus a web-pretrained VLA remains a web-transfer claim when its principal contribution grounds that backbone, even if evaluation also changes tasks and robots. Changing team size is coded as embodiment transfer when a policy accommodates variable agent configurations, and as sim-to-real when the focal contribution is hardware deployment. These assignments support transparent aggregation without implying that the taxonomy exhausts all possible gaps.

\begin{table}[pos=tbp]
\centering
\caption{Transferred forms and diagnostic questions. The connection to Eq.~\eqref{eq:bound} is heuristic for closed-loop control}
\label{tab:forms}
\small
\begin{tabularx}{\linewidth}{@{}>{\raggedright\arraybackslash}p{2.5cm}YY@{}}
\toprule
\textbf{Form} & \textbf{Potential benefit} & \textbf{First failure to test} \\
\midrule
Data & Cover target-relevant variation & Mixture mismatch; harmful sources \\
Representations & Reduce effective perceptual mismatch & Source shortcuts; lost action detail \\
Parameters and policies & Reuse a useful hypothesis or behavior & Incompatible tasks; forgetting \\
Models and simulators & Align dynamics or synthesize experience & Incorrect action consequences \\
Task conditioning & Index behavior on task or context & Unseen instructions or interfaces \\
\bottomrule
\end{tabularx}
\end{table}

\subsection{Metrics and Historical Continuity}
\label{sec:metrics}
\label{sec:eras}

Jumpstart, asymptotic gain, and time-to-threshold remain useful transfer metrics \citep{taylor2009transfer}. Zero-shot success is jumpstart at zero adaptation budget; few-shot curves measure performance against target experience. Axis-controlled generalization varies factors such as lighting, objects, or camera pose \citep{pumacay2024colosseum, gao2025taxonomy}. Retention measures prior-task or semantic capabilities after adaptation \citep{liu2023libero, zhou2025chatvla}. A headline success rate does not identify the source of a gain; each numerical comparison below is interpreted within its own study and graded separately in Section~\ref{sec:evidence}.

\begin{figure}[pos=tbp]
\centering
\includegraphics[width=\linewidth]{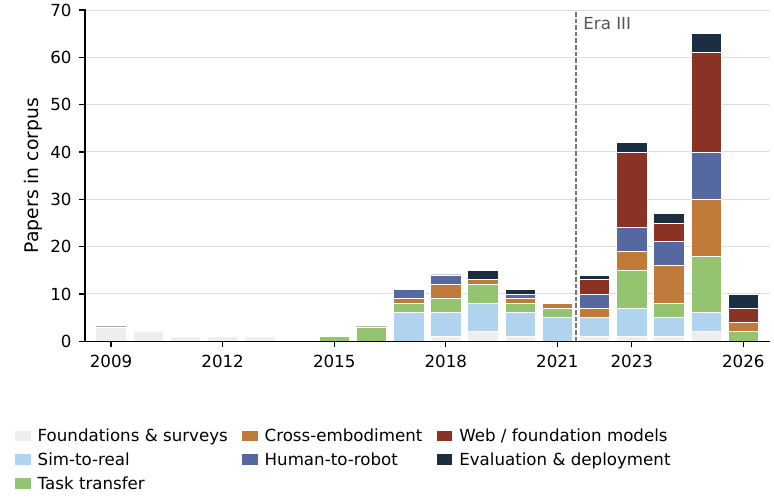}
\caption{Reference years and primary topical categories in the 232-work corpus, using the first online year where it precedes the proceedings imprint year. Coded papers use their primary gap; uncoded works are grouped by their role in the survey. The distribution reflects curated coverage, not field-wide publication rates. Three pre-2009 works precede the plotted range. The main search closed in July 2026, followed by a targeted November update}
\label{fig:corpus}
\Description{Stacked annual counts in the curated corpus show more embodiment and web-transfer works in recent years.}
\end{figure}

Three approximate eras help locate mechanisms (Fig.~\ref{fig:corpus}); the corpus also includes earlier precursors \citep{thrun1995lifelong, sutton1999between, kearns2002near}. Before 2017, work emphasized inter-task mappings, reusable values, and formal metrics \citep{taylor2009transfer, lazaric2012transfer}. From 2017--2021, deep learning supported randomization, simulator grounding, meta-learning, and visual alignment. From 2022, heterogeneous pretraining and target fine-tuning joined these mechanisms into generalist policies, followed by reinforcement-learning post-training and video world models \citep{oneill2023open, kim2024openvla, amin2025pistar, assran2025vjepa}. Earlier mechanisms remain in use across these eras as their combinations and scale change.

\FloatBarrier

\section{Simulation-to-Real Transfer}
\label{sec:sim2real}

Simulation is robotics' lowest-cost source domain: unlimited experience, perfect observability, safe failures. Its cost is a compound shift: dynamics mismatch real actuators, contacts, and latencies; images mismatch cameras and scenes. Three mechanisms address it: randomization, grounding, and deployment-time adaptation, weighed in Section~\ref{sec:s2rsynth} and all made less costly by scale and automation (GPU-parallel physics, language-model-written rewards and ranges). Perception and tactile-deformable transfer are the modality-specific cases.

\subsection{Randomization and Robust Training}
\label{sec:dr}
\label{sec:robust}

Domain randomization (DR) trains on perturbed simulators so the model tolerates the variation. \citet{tobin2017domain} introduced the visual version: a detector trained only on synthetic images with randomized textures, lighting, and camera poses localized real objects to about 1.5\,cm and grasped with no real image. \citet{sadeghi2017cad2rl} flew a real quadrotor on a policy trained in randomized CAD; \citet{peng2018sim} extended it to dynamics, sampling mass, friction, damping, and latency to push objects on hardware without fine-tuning. \citet{tan2018sim} showed the limit on legged hardware: dynamics randomization alone was insufficient, and quadruped transfer required first improving the actuator model and randomizing only residual uncertainty (Section~\ref{sec:grounding}). Distributed RL with visual and dynamics randomization reached vision-based in-hand reorientation on a 24-degree-of-freedom (DoF) Shadow Hand \citep{andrychowicz2020learning}, and automatic domain randomization (ADR), growing the ranges as the policy improves, extended this to a physical Rubik's Cube \citep{akkaya2019solving}.

Real data closes the loop: SimOpt adapts the distribution until simulated rollouts match real trajectories, needing no real-world reward \citep{chebotar2019closing}; BayRn tunes it by Bayesian optimization of sparse real return \citep{muratore2021data}; and \citet{muratore2022robot} review the progression from static to adaptive and adversarial randomization. Ranges stay largely empirical (Sections~\ref{sec:llmsim} and~\ref{sec:challenges}).

\paragraph{Worst-case aggregation.} EPOpt optimizes the conditional value at risk over a model ensemble, adapting it as real trajectories arrive \citep{rajeswaran2017epopt}; robust adversarial RL replaces the ensemble with a learned adversary applying destabilizing forces \citep{pinto2017robust}. Both are DR optimizing a lower tail of the widened $\mathcal{D}_S$ in Eq.~\eqref{eq:bound} instead of its mean. The conservatism helps when the target is unknown and failure expensive, and wastes performance otherwise.

\subsection{Grounding the Simulator}
\label{sec:grounding}

The complementary strategy fixes the simulator. System identification fits physical parameters, but the hardest errors are structural: \citet{hwangbo2019learning} replaced the analytic actuator model in a rigid-body simulator with a learned network, and ANYmal policies trained there transferred zero-shot, including fall recovery. Residual formulations add a corrective policy to a conventional controller, targeting the friction and contact effects the nominal model misses \citep{johannink2019residual}. ASAP applies the residual idea to the simulator, learning a delta action model from real rollouts and fine-tuning agile whole-body policies for a Unitree G1 humanoid \citep{he2025asap}. Differentiable simulation is orthogonal: a surrogate supplies gradients for a high-fidelity forward simulator, and quadruped locomotion transfers zero-shot after minutes of optimization \citep{song2024learning}.

Grounding also runs from the real scene to simulation. RialTo scans real environments into articulated digital twins, imports real demonstrations, robustifies with large-scale RL in the twin, and distills back to hardware; robustness rises by over 67 percentage points against an imitation baseline across eight manipulation tasks \citep{torne2024reconciling}. This real-to-sim-to-real pattern reappears in Sections~\ref{sec:percept} and~\ref{sec:eval}.

\subsection{Deployment-Time Adaptation}
\label{sec:adapt}

A third family infers the dynamics at deployment, reframing the gap as task inference (Section~\ref{sec:meta}). The privileged teacher--student scheme of \citet{lee2020learning} distills a teacher trained on ground-truth terrain into a student reading proprioceptive history, which traversed varied terrain on ANYmal without real-world training. Rapid motor adaptation (RMA) makes the inference explicit: an adaptation module regresses a latent of environment extrinsics from state history, adapting within a fraction of a second on a Unitree A1 \citep{kumar2021rma}. With transformers the mechanism became in-context learning: a causal transformer trained over observation-action histories in randomized simulation walked the Digit humanoid outdoors zero-shot, handling perturbations in context with no weight updates \citep{radosavovic2024real}.

\subsection{Scale and Automation}
\label{sec:gpusim}
\label{sec:llmsim}

Isaac Gym changed the economics around 2021, moving physics and learning onto one GPU with speedups of two to three orders of magnitude over CPU pipelines \citep{makoviychuk2021isaac}. \citet{rudin2021learning} trained ANYmal locomotion with thousands of parallel robots in under four minutes on flat terrain and about twenty minutes on rough terrain, transferring to hardware. The frontier moved to harder embodiments: vision-based quadruped parkour, dexterous in-hand manipulation, and in-hand object rotation all reached hardware \citep{zhuang2023robot, handa2023dextreme, qi2022inhand}, and champion-level drone racing, combining simulated RL with real-flight residuals, beat human world champions \citep{kaufmann2023champion}. In the 2024--2026 humanoid line, H2O retargets human motion into whole-body tracking to teleoperate a Unitree H1 from RGB in real time, OmniH2O adds VR, language, and autonomous policies, and HOVER distills the control modes into one controller \citep{he2024learning, he2024omnih2o, he2025hover} (Section~\ref{sec:humanoid}).

\paragraph{Automating the expert.} Recent systems delegate reward functions, randomization ranges, and curricula to large language models (LLMs). Eureka has an LLM write and evolve reward code against source; it beats human-expert rewards on 83\% of 29 simulated tasks, with an average normalized improvement of 52\% \citep{ma2024eureka}. DrEureka extends the loop to sim-to-real design: its rewards and randomization configurations match human-designed pipelines, and it solves a quadruped yoga-ball balancing task with no hand-tuned baseline \citep{ma2024dreureka}. These compress expert iteration into prompt iteration; whether that extends to safety-critical choices stays open (Section~\ref{sec:challenges}).

\subsection{Perception-Side Transfer}
\label{sec:percept}

Visual sim-to-real works mainly on the RGB channel, where the appearance gap is widest; it began as adversarial domain adaptation. GraspGAN translated simulated images toward real while aligning features adversarially, cutting the real data for target grasping performance by up to a factor of fifty \citep{bousmalis2018using}. RCAN inverted it, mapping randomized images into a canonical simulation domain: 70\% zero-shot real grasp success, 91\% after fine-tuning on 5{,}000 real grasps, comparable to QT-Opt trained on 580{,}000 real grasp attempts \citep{james2019sim, kalashnikov2018qtopt}. Neural rendering then reconstructs appearance instead of adapting it: SplatSim replaces the simulator's renderer with Gaussian-splat reconstructions of real scenes, and RGB policies trained there transfer zero-shot at 86.25\% average success against 97.5\% for real-trained policies \citep{qureshi2025splatsim}. With digital-twin pipelines \citep{torne2024reconciling}, reconstruction has largely displaced adversarial image translation, though feature-level adaptation survives inside pretrained encoders (Section~\ref{sec:pvr}).

\subsection{Tactile and Deformable Transfer}
\label{sec:tactile}
\label{sec:deformable}

Contact-rich manipulation depends on touch, from contact physics that simulators model poorly: stiff, high-frequency, and sensitive to surface microgeometry no unified robot description format (URDF) records. In Section~\ref{sec:defs}'s terms, the tactile gap loads onto $\lambda^{*}$ and away from the divergence term, so grounding and translation should dominate and pure randomization should transfer touch less reliably than vision; the second stays untested for want of a controlled study.

This illustrates the data-generation-apparatus pattern of Section~\ref{sec:patterns}. Optical tactile sensors, imaging elastomer deformation, made touch reproducible across labs. Because they emit images, the machinery of Section~\ref{sec:percept} applies: TACTO renders tactile images fast enough to train in simulation \citep{wang2022tacto}, and \citet{church2021tactile} translate real tactile images into the simulated domain, giving a simulation-trained policy in-distribution input years before real-to-sim inversion became standard for RGB.

Sparsh pretrains self-supervised touch representations across sensors and tasks, a tactile counterpart of the pretrained visual representations of Section~\ref{sec:pvr}, inheriting their pretrained-versus-from-scratch question \citep{higuera2024sparsh}. Fusion with vision helps most, supporting fine-grained bimanual manipulation and cross-shape in-hand rotation \citep{qi2023general}, and supervision is widening beyond the robot, to human tactile demonstrations (Section~\ref{sec:human}) and contact sound from in-the-wild handheld grippers \citep{yu2024mimictouch, liu2024maniwav}. Tactile transfer is roughly where visual transfer was around 2018: no analogue of the pooled cross-embodiment corpora of Section~\ref{sec:datasets}, cross-sensor transfer asserted more often than measured, and the statistical thinness of Section~\ref{sec:realeval}.

\paragraph{Deformables.} Deformables show the widest gap. Cloth, rope, and soft objects have high-dimensional, partially observable configuration spaces, distributed contact, and material parameters that are neither visible nor identified; the gap loads onto $\lambda^{*}$ plus a partial observability Eq.~\eqref{eq:bound} does not express. \citet{matas2018sim} train in simulation with visual and dynamics randomization and deploy zero-shot to a Kinova Mico; the easiest cloth task reaches roughly 90\% while diagonal folding and hanging reach about 40\% and 47\%, transferring perception and gross motion but not material behavior. \citet{seita2020fabric} instead transfer demonstrations from a corner-pulling supervisor with ground-truth fabric state, training imitation on randomized renderings; RGBD variants attain a uniform 83--95\% coverage across 180 physical trials on a da Vinci Research Kit, the teacher--student construction of Section~\ref{sec:adapt} with a hand-designed teacher. \citet{ganapathi2021correspondences} transfer a representation: dense visual correspondences learned on simulated fabric specify goal configurations, reaching 80.3\% average success across ten tasks on two physical platforms. \citet{ha2021flingbot} transfer an action, a self-supervised pick-stretch-fling primitive: over 80\% coverage within three actions on novel cloths in simulation, and, after fine-tuning on the real dual-arm platform, more than four times a quasi-static baseline. The gap narrows most by changing the transferred object (policy, demonstrations, correspondence field, action primitive) and least by randomizing harder; with inconsistent reporting and no shared benchmark, these read as four separate experiments and support no ranking.

\subsection{Synthesis}
\label{sec:s2rsynth}

Randomization supplies robustness at the risk of conservatism; grounding corrects identifiable dynamics or appearance; deployment-time adaptation infers residual mismatch from history. Actuator models, residual dynamics, and real-to-sim twins provide concrete grounding results \citep{hwangbo2019learning, he2025asap, torne2024reconciling}, while history-conditioned policies support fast adaptation \citep{kumar2021rma, radosavovic2024real}. These mechanisms compose, but their separate contributions require matched ablations. Visual reconstruction does not remove contact-model error, and results on rigid objects cannot establish equivalent transfer for tactile or deformable tasks.

\section{Transfer Across Tasks}
\label{sec:task}

Task transfer changes the reward, goal, or demonstration specification while retaining a sufficiently compatible body and environment. A useful policy for one task need not be useful for another: a fixed behavior can conflict with a new reward, while a shared representation or action vocabulary can still reduce learning cost. Lifelong learning framed this ambition before deep learning \citep{thrun1995lifelong}; rapid damage recovery using a precomputed gait map illustrates how compact prior knowledge can replace expensive search \citep{cully2015robots}.

\subsection{Reusing Policies, Values, and Skills}
\label{sec:taskdef}
\label{sec:reuse}
\label{sec:skills}

Policy distillation transfers teachers into a student \citep{rusu2016policy}; progressive networks preserve earlier columns and add new ones, including sim-to-real reaching on a Jaco arm \citep{rusu2016progressive, rusu2017sim}. Fine-tuning is simpler but can overwrite earlier capabilities. Value-side methods instead separate reward-dependent quantities from reusable dynamics. Successor features and generalized policy improvement support reuse across rewards under their formal assumptions \citep{barreto2017successor}; forward--backward representations learn a policy-accessible representation for unseen rewards \citep{touati2021learning, touati2023does}. FB-CPR scales this approach with motion regularization for humanoid behaviors \citep{tirinzoni2025zero}. Planner-side transfer can also reuse a sampling distribution rather than network weights \citep{he2026sample}.

Selective parameter reuse makes the transfer mechanism more specific. On simulated Fetch tasks, \citet{he2025fewshot} transfer first-layer weights and biases along forward and reverse task routes, illustrating that final success and convergence speed can respond differently to the source task. On PyBullet manipulation tasks, \citet{he2026decomposition} combine first-layer critic transfer with staged reward redesign; gains vary by route, and obstacle-aware Pick-and-Place has a lower mean success than training from scratch. These studies motivate selecting both the transferred layer and the task route rather than assuming that broader reuse is always beneficial.

Skills transfer an action vocabulary. Options formalize temporally extended actions through initiation sets, policies, and termination conditions \citep{sutton1999between}. DIAYN discovers diverse reward-free behaviors \citep{eysenbach2019diversity}; SPiRL learns a latent skill space and prior from offline experience to guide downstream exploration \citep{pertsch2020accelerating}. Their practical limit is compositional coverage: a new task outside the available primitives requires new behavior, regardless of how efficiently the old skills are selected. Curricula and expanding randomization ranges similarly organize which experience is available when \citep{zhuang2023robot, akkaya2019solving}.

\subsection{Learning to Adapt and Multi-Task Pretraining}
\label{sec:meta}

Model-agnostic meta-learning optimizes an initialization for adaptation with a few gradient steps; first-order approximations reduce its cost \citep{finn2017model, nichol2018first}. RL$^2$ encodes adaptation in recurrent state, while PEARL infers a latent task context and combines it with off-policy learning \citep{duan2016rl, rakelly2019efficient}. Model-based adaptation instead updates dynamics estimates online, including physical recovery from damage or terrain changes \citep{nagabandi2019learning}. These approaches rely on the meta-training distribution covering the kinds of variation encountered later. Meta-World exposed poor generalization beyond narrow task families \citep{yu2019metaworld}.

Sequence models extend context-based adaptation. Algorithm distillation learns from complete learning histories and improves in context on held-out tasks \citep{laskin2023incontext}; history-conditioned locomotion demonstrates related adaptation on hardware \citep{radosavovic2024real}. This connection does not make explicit meta-learning and ordinary sequence training identical: the training distributions and objectives still determine what adaptation is learned.

Multi-task pretraining shares experience without an explicit adaptation algorithm. MT-Opt pools real-robot RL across tasks \citep{kalashnikov2021mt}; BC-Z uses language or video conditioning \citep{jang2022bcz}; RT-1 shares a transformer across manipulation tasks and data sources \citep{brohan2022rt1}. Such systems move the design problem toward task coverage, conditioning, and interference, anticipating cross-embodiment pooling.

\subsection{Offline-to-Online Adaptation and Limits}
\label{sec:offline2online}
\label{sec:tasksynth}

Offline-to-online learning is a protocol that can accompany any gap: prior data initializes or guides a policy, then target interaction improves it. AWAC uses advantage-weighted updates \citep{nair2020awac}; Cal-QL addresses value miscalibration during the transition \citep{nakamoto2023calql}; RLPD shows that off-policy design choices can match specialized procedures without a separate pretraining phase \citep{ball2023efficient}. SERL and HIL-SERL bring demonstration-assisted and correction-assisted learning to precise physical manipulation \citep{luo2024serl, luo2025precise}. Policy Decorator preserves a pretrained policy and learns a residual, while diffusion-policy optimization updates the generative policy itself \citep{yuan2025policy, ren2024dppo}. Section~\ref{sec:posttrain} follows these mechanisms into VLA post-training.

The strongest hardware demonstrations in this group concern targeted adaptation rather than unrestricted task generalization. Successor-feature and forward--backward methods offer structured reward reuse, but much of their evidence is simulated; multi-task fine-tuning avoids correspondence engineering but inherits interference and forgetting. A useful comparison therefore fixes the target task and interaction budget, tests a target-only learner, and measures both adaptation speed and retention. No scalar measure of task similarity in this corpus substitutes for that comparison.

\section{Transfer Across Embodiments}
\label{sec:embodiment}

Robot data is siloed by body: each arm has its own kinematics, gripper, cameras, and interface. Cross-embodiment transfer asks whether data pooled from many robots can train policies for any of them, even ones outside the data; the pretraining template of Section~\ref{sec:foundation} is only as good as its pooled data.

\subsection{Conditioning on the Body}
\label{sec:morphology}

A policy assumes a fixed observation and action space, which robots with different link counts do not share. Two routes emerged: body as data, or network shaped to the body.

Hardware-conditioned policies feed a kinematics-and-dynamics vector into one universal network controlling manipulators of different degrees of freedom, transferring to held-out robots \citep{chen2018hardware}.

Morphology-aware architecture holds the sharpest negative result. NerveNet passes messages along the body's graph, transferring to bodies with more limbs or different topologies \citep{wang2018nervenet}; Shared Modular Policies place one locally coordinated module at every actuator across many morphologies, including unseen ones \citep{huang2020one}. \citet{kurin2021morphology} find the graph a poor inductive bias: on the same incompatible-morphology benchmarks, an amortized transformer attending over limbs matches or beats message passing, tracing its weakness to propagation over many hops.

MetaMorph, a transformer over morphology-token sequences, generalizes zero- and few-shot to unseen procedurally generated morphologies \citep{gupta2022metamorph}. A 2025 scaling law shows locomotion policies trained on roughly one thousand procedurally varied embodiments generalize better as bodies multiply, transferring zero-shot sim-to-real to Unitree Go2 and H1 robots absent from training \citep{ai2025towards}. Almost all evaluation is simulated, the one hardware result within a shared actuator convention.

\subsection{Pooling Data Across Robots}
\label{sec:pooling}

The complementary approach skips morphology encoding, tokenizing everything for one sequence model. Only held-out evaluation tests cross-embodiment generalization.

Headline results are in-pool. Open X-Embodiment pooled standardized episodes from 22 embodiments; RT-1-X beat specialists on the smaller in-domain datasets, and RT-2-X extended the gains while acquiring emergent skills \citep{oneill2023open}. CrossFormer spans single arms, quadcopters, and quadrupeds, matching in-pool specialists without manual observation or action alignment \citep{doshi2024scaling}, with bidirectional positive transfer between manipulation and navigation \citep{yang2024pushing}. HPT's shared trunk, fed by per-embodiment stems, improves unseen-task performance by over 20\% \citep{wang2024scaling}.

Held-out-robot results are fewer, seldom zero-shot: RoboNet's seven-platform video pool improved fine-tuning to held-out robots and viewpoints \citep{dasari2019robonet}; Gato and RoboCat adapt to new tasks and arms from 100--1{,}000 demonstrations \citep{reed2022generalist, bousmalis2023robocat}; Octo fine-tunes to new sensor suites and action spaces within hours \citep{ghosh2024octo}. Target-robot data is still required; pooling lowers its price.

\subsection{Aligning Observations and Actions}
\label{sec:actionspaces}

Pooling works best when data already looks similar; two families manufacture it. On the observation side, Mirage paints the target robot out and the source in, with a forward-dynamics action correction, transferring to an unseen arm zero-shot \citep{chen2024mirage}; RoVi-Aug repaints demonstrations with different robots and viewpoints via diffusion, raising multi-robot success up to 30\% and deploying zero-shot on an unrecorded robot-camera combination \citep{chen2024rovi}. On the action side, latent-action methods learn a shared vocabulary over each body's commands. LAPA vector-quantizes inter-frame visual change into latent actions, pretraining on actionless internet and human video; a small action-labeled fine-tune then beats a state-of-the-art VLA trained on ground-truth actions \citep{ye2024latent}. UniAct learns a universal codebook of atomic behaviors with lightweight per-platform decoders, its 0.5B model reportedly outperforming a 14$\times$ larger OpenVLA \citep{zheng2025universal}; UniVLA learns task-centric latent actions in DINO feature space and exceeds OpenVLA with under one twentieth of the pretraining compute \citep{bu2025univla}. These cross-system comparisons vary in data, training, and action space, so the ratios describe whole systems. Latent actions also bridge the human-to-robot gap (Section~\ref{sec:human}) by inferring unobserved commands from observed motion.

\subsection{Navigation}
\label{sec:navigation}

GNM trains one goal-conditioned visual navigation policy on six platforms spanning wheeled bases, a quadruped, and an aerial vehicle, deploying on robots absent from training \citep{shah2023gnm}; ViNT scales this to a transformer trained on roughly a hundred hours of heterogeneous data with a swappable goal-conditioning head \citep{shah2023vint}; NoMaD folds undirected exploration into the same diffusion policy by masking the goal token \citep{sridhar2024nomad}.

These results are cleaner than any manipulation counterpart because the action space is nearly embodiment-invariant: reduced to $\mathrm{SE}(2)$ waypoints or body velocities, a wheeled base, quadruped, and quadrotor differ only in ways an onboard controller absorbs. That controller supplies the shared space Section~\ref{sec:actionspaces} labors to construct, at engineering cost.

Simulated navigation was narrowly solved earlier: DD-PPO reached near-perfect PointGoal navigation with 2.5 billion frames in Habitat \citep{savva2019habitat}, yet it did not transfer to the real world, where progress came from real heterogeneous data (Section~\ref{sec:sim2real}). LM-Nav composes off-the-shelf language, vision-language, and navigation models to follow free-form instructions without fine-tuning \citep{shah2022lmnav}, task-conditioned transfer predating VLAs.

\subsection{Datasets}
\label{sec:datasets}

Robot corpora follow two collection philosophies. Aggregation (Open X-Embodiment, RoboMIND) inherits many platforms' inconsistencies along with their diversity \citep{oneill2023open, wu2024robomind}; standardization (BridgeData V2, DROID, AgiBot World) yields homogeneity while reducing embodiment diversity \citep{walke2023bridgedata, khazatsky2024droid, bu2025agibot}; RH20T alone records force-torque and audio \citep{fang2023rh20t}. A diffusion policy pretrained on AgiBot World reportedly completes tasks about 30\% more often than one pretrained on Open X-Embodiment \citep{bu2025agibot}. This in-domain evaluation confounds data quality with in-domain advantage, though consistent with composition as the operative variable.

\subsection{When Does Pooling Help?}
\label{sec:whenpooling}

A second wave finds the gains fragile. ReMix learned mixture weights over Open X-Embodiment subsets by distributionally robust optimization, beating uniform weighting by over 40\% and human-curated RT-X weights by over 20\% \citep{hejna2024remix}: weighting alone decides much of the outcome. \citet{xing2025shortcut} identify a failure mechanism: when sub-datasets are narrow and mutually dissimilar, policies learn shortcut features separating them, leaving tasks unlearned and generalization collapsed. \citet{lin2025data} fit power laws where generalization tracks distinct environments and objects while demonstrations per environment saturate quickly; with diversity-first collection, four collectors reached roughly 90\% success in novel environments in an afternoon. \citet{saxena2025what} disentangle the axes in simulation, finding camera pose and spatial arrangement critical and retrieval-based selection over DROID improving downstream policies up to 70\%. \citet{yang2026data} distinguish shift types: perceptual shifts respond to broad unstructured diversity, morphology shifts to paired ``analogy'' demonstrations aligning scenes and trajectories across embodiments; recomposing data accordingly improved real-world cross-embodiment transfer by 22.5\% without new collection. Together they revive the negative-transfer question at dataset scale (Section~\ref{sec:ch-negative}).

\paragraph{Lessons.} Evidence is strongest where source and target share, or are given, a compatible action abstraction, and weaker where it must be learned. Well supported: in-pool gains and navigation transfer to held-out platforms. Thinner: zero-shot manipulation on a held-out arm, shown by a few systems on a few arm pairs \citep{chen2024mirage, chen2024rovi}, and latent-action size and compute advantages. Other gains concentrate in bookkeeping: mixture weights, diversity axes, and observation-action alignment \citep{hejna2024remix, saxena2025what}; shortcut learning is the canonical failure mode \citep{xing2025shortcut}. Section~\ref{sec:patterns} generalizes the diversity finding; Section~\ref{sec:guidance} turns these into budget-conditioned defaults.

\section{Transfer from Humans}
\label{sec:human}

Human experience offers task knowledge with observation, action, and embodiment mismatches. Learning from demonstration is well established \citep{argall2009survey}; the transfer question is which parts of that experience become robot supervision. Robot teleoperation produces native actions, raw video does not, and motion capture or hand-held devices occupy intermediate positions. These sources should not be treated as interchangeable data.

\subsection{Few-Shot Imitation and Human Video}
\label{sec:fewshotil}
\label{sec:humanvideo}

One-shot imitation first learned across robot demonstrations, using attention, gradient adaptation, or task embeddings \citep{duan2017oneshot, finn2017oneshot, james2018taskembedded}. Domain-adaptive meta-learning extended the mechanism to a human video through a learned adaptation loss \citep{yu2018oneshot}. BC-Z scaled language- and video-conditioned imitation across tasks \citep{jang2022bcz}. The transferred object can thus be a conditioning interface, rather than an executable human trajectory.

Actionless video requires additional grounding. Behavioral cloning from observation infers actions through inverse dynamics \citep{torabi2018behavioral}; XIRL learns rewards through temporal alignment \citep{zakka2021xirl}; other systems extract interaction priors, retargeted hand poses, or latent plans \citep{bahl2022human, shaw2022videodex, wang2023mimicplay}. VRB transfers affordances learned from human video \citep{bahl2023affordances}, while Vid2Robot uses video prompts \citep{jain2024vid2robot}. EgoMimic co-trains egocentric human and robot data under kinematically similar collection \citep{kareer2024egomimic}; Phantom combines visual robot replacement with hand-pose-derived actions \citep{lepert2025phantom}. These successes depend on how missing actions are supplied or bypassed. Ordinary video still provides little direct information about forces and contact.

\subsection{Collection Interfaces and Humanoid Motion}
\label{sec:interfaces}
\label{sec:humanoid}

Changing collection hardware can reduce the mismatch before learning. ALOHA records bimanual robot teleoperation; Mobile ALOHA extends collection to whole-body mobile manipulation \citep{zhao2023learning, fu2024mobile}. UMI uses a hand-held gripper, wrist camera, relative trajectories, and latency matching to make demonstrations portable \citep{chi2024universal}; DexCap extends instrumented collection to dexterous hands \citep{wang2024dexcap}. Diffusion policies provide a flexible student for multimodal demonstrations \citep{chi2023diffusion}. These interfaces reduce retargeting burdens while retaining differences in dynamics, reachability, and contact. They trade collection design against algorithmic alignment.

Humanoid morphology makes human motion more directly usable, although retargeting and balance remain necessary. Animal-motion imitation preceded current humanoid pipelines \citep{peng2020learning}. ExBody combines expressive upper-body motion with robust locomotion \citep{cheng2024expressive}; HumanPlus couples simulated shadowing with robot imitation \citep{fu2024humanplus}; OKAMI derives object-aware manipulation from RGB-D human video \citep{li2024okami}. H2O, OmniH2O, and HOVER provide motion-tracking controllers \citep{he2024learning, he2024omnih2o, he2025hover}. PH2D co-trains in a unified human--humanoid state-action space \citep{qiu2025humanoid}, bringing human transfer closer to the cross-embodiment problem.

\subsection{Synthesizing Demonstrations and Remaining Gaps}
\label{sec:synthesis}

MimicGen recombines demonstration segments around objects in new scenes. DexMimicGen extends synthesis to bimanual dexterous manipulation \citep{mandlekar2023mimicgen, jiang2024dexmimicgen}. DreamGen instead generates video rollouts and supplies pseudo-actions through latent or inverse-dynamics models \citep{jang2025dreamgen}. Synthesis multiplies source effort but also propagates geometric, dynamical, and labeling errors. Its value requires a comparison at matched source-demonstration and target-adaptation budgets.

Across these routes, representations, rewards, and task prompts are useful even when actions cannot be recovered reliably. Direct data co-training is more plausible when collection constrains the human to a robot-compatible interface. The corpus contains no budget-matched comparison establishing that collection-interface design generally outperforms algorithmic alignment. Force-sensitive manipulation and unfamiliar kinematics remain limits on conclusions drawn from visual success alone.

\section{Transfer from Foundation Models}
\label{sec:foundation}

Web data supplies visual, linguistic, and predictive structure with few robot action labels. Transferring that structure into control requires a physical-action interface. Methods divide by how deep the pretrained model sits: frozen planner, perception encoder, policy initialization, or world model.

\subsection{Frozen Models as Planners}
\label{sec:orchestration}

The first wave kept the model out of the motor loop. Frozen large language models plan household tasks without robot data \citep{huang2022language}; SayCan grounds the LLM in affordances via a value function \citep{ahn2022saycan}; PaLM-E fed images and states into a 562B-parameter language model, transferring from joint web-scale and embodied training \citep{driess2023palme}. Others replan from textual feedback or write policy code over perception APIs, control primitives, and vision-language model (VLM) 3D value maps \citep{huang2022inner, liang2023code, huang2023voxposer}.

\subsection{Pretrained Visual Representations}
\label{sec:pvr}

Pretrained visual representations (PVRs) form the policy's frozen front end, replacing ground-truth state even out-of-domain \citep{parisi2022unsurprising}. R3M, trained on Ego4D human video, raised imitation success over 20\% above from-scratch across twelve simulated tasks \citep{nair2022r3m}. VC-1, selected on CortexBench's seventeen embodied tasks, won on average, though none dominated \citep{majumdar2023where}. Others scaled masked autoencoding to a 307M-parameter vision transformer (ViT), read zero-shot rewards from offline value learning, added language supervision, and distilled CLIP features into spatial and 3D pathways \citep{radosavovic2022realworld, ma2023vip, karamcheti2023languagedriven, shridhar2021cliport, shen2023distilled}.

Re-evaluations complicate this: a tuned, augmented from-scratch baseline matched or beat frozen PVRs on most benchmarks \citep{hansen2023pretraining}; composition mattered more than scale, egocentric video no better than standard datasets \citep{dasari2023unbiased}; robustness under shift tracked ViT emergent segmentation, not ImageNet accuracy or manipulation-specific pretraining \citep{burns2024what}; and frozen PVRs often did not help model-based RL \citep{schneider2024surprising}. Theia distills several vision foundation models into one encoder; MCR pretrains on robot data with action-aware objectives \citep{shang2024theia, jiang2025robots}. 3D-reasoning policies beat 2D at viewpoint transfer and few-demonstration generalization \citep{shridhar2022peract, goyal2023rvt, ze2024dp3}.

Transferred perception also requires a task-specific validity check. In cluttered manipulation, \citet{he2026confidence} distinguish category presence from visibility of the intended object: detectors can retain high confidence while localizing a same-category distractor under occlusion. Confidence alone therefore need not establish that a pretrained detector supplies the target-grounded information required for control or active perception.

\subsection{Vision-Language-Action Models}
\label{sec:vla}

\begin{figure}[pos=!t]
\centering
\includegraphics[width=\linewidth]{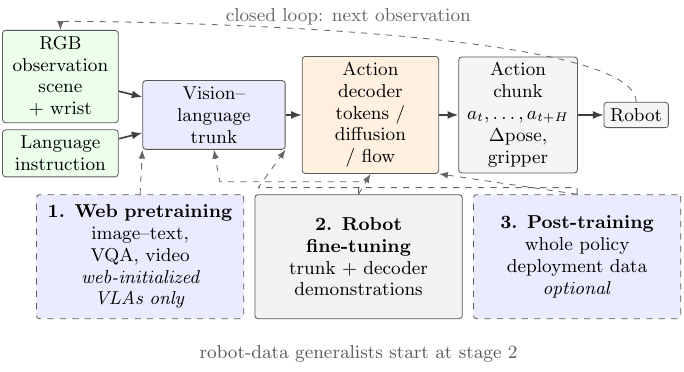}
\caption{Shared anatomy of a vision-language-action model. Training (bottom) mirrors the large-language-model procedure: web pretraining, supervised fine-tuning on teleoperated demonstrations, then optional post-training (Section~\ref{sec:posttrain}). Dashed stage boxes mark steps some models skip: web-initialized VLAs alone run stage 1 (generalists start at stage 2), and post-training (stage 3) is optional. Fine-tuning updates trunk and decoder; post-training updates the whole policy.}
\label{fig:vlapipe}
\Description{Images and instructions enter a shared trunk and action decoder, with web pretraining, robot fine-tuning, and optional post-training.}
\end{figure}

The mainline embeds the model in the policy, in two families: \emph{robot-data generalists} train the trunk on pooled robot demonstrations, using web models at most as frozen encoders; \emph{web-initialized VLAs} fine-tune a web-pretrained vision-language backbone into the policy. A second, architectural split sets monolithic policies against dual-system designs, which stack a slow vision-language module over a fast low-level policy \citep{bjorck2025groot, figure2025helix}, moving Section~\ref{sec:orchestration}'s planner inside. RT-1 set the robot-data foundation \citep{brohan2022rt1}; Octo and RDT-1B carried it to pooled Open X-Embodiment data and a unified bimanual action space \citep{ghosh2024octo, liu2024rdt}. The web-initialized branch opens with RT-2 (Fig.~\ref{fig:vlapipe}), whose manipulation of objects and symbols absent from robot data is the canonical web-to-control evidence \citep{brohan2023rt2}. OpenVLA beat the 55B-parameter RT-2-X with $7\times$ fewer parameters \citep{kim2024openvla}; $\pi_0$ swapped discrete tokens for a flow-matching action expert \citep{black2024pi0}. FAST's tokenizer lets autoregressive VLAs match $\pi_0$ with up to five times less training \citep{pertsch2025fast}; OpenVLA-OFT's parallel decoding, action chunking, and continuous outputs reach 97.1\% average success on LIBERO at roughly 26$\times$ higher throughput \citep{kim2025oft}. SpatialVLA adds 3D position encodings and adaptive action grids \citep{qu2025spatialvla}; a matched-conditions ablation finds backbone, decoder, and training procedure, not scale, separate generalist policies \citep{li2026robovlm}.

The remaining variants are all web-initialized. Embodied chain-of-thought reasoning has the policy generate a plan, subtask, object references, and gripper predictions before acting, improving generalization to unseen instructions and objects at some latency \citep{zawalski2024ecot}. 3D-VLA embeds a generative world model in the VLA, bridging to Section~\ref{sec:worldmodels} \citep{zhen20243dvla}. SmolVLA, compact and trained on community datasets for consumer hardware, tests the approach at laboratory scale \citep{shukor2025smolvla}.

The humanoid case pairs the most-publicized claims with the least-verifiable evidence: GR00T N1 pairs a web-pretrained reasoning module with a diffusion action module \citep{bjorck2025groot}, while its successor and several prominent humanoid systems of 2025--2026 exist only as model cards (Section~\ref{sec:realeval}).

Industry systems released since 2025 are all web-initialized. Gemini Robotics builds VLAs on the Gemini backbone, its 1.5 release adding Motion Transfer across three embodiments and interleaved thinking before acting \citep{gemini2025robotics, gemini2025robotics15}. Figure's Helix splits humanoid control between a 7B reasoner and a fast 80M visuomotor policy \citep{figure2025helix}. $\pi_{0.5}$, the only peer-reviewed system here as of July 2026, showed multi-minute cleaning in unseen homes via heterogeneous co-training \citep{black2025pi05}; $\pi_{0.7}$ claims emergent zero-shot cross-embodiment generalization at further scale, still as a report \citep{ai2026pi07}. Section~\ref{sec:challenges} returns to reproducibility.

\subsection{What Transfers, What Is Lost}
\label{sec:whattransfers}

Transfer also subtracts, and whether web knowledge survives action training has thin evidence. ChatVLA documents ``spurious forgetting'': action fine-tuning overwrites visual-text alignments until the model cannot describe the scene it manipulates \citep{zhou2025chatvla}. Knowledge Insulation gives the clearest controlled result: a naive continuous action expert damages the backbone, while insulating it from action gradients and co-training on vision-language data speeds training and improves generalization \citep{driess2025knowledge}. Both reduce the damage but neither prices it, and no budget-matched study here restores what is lost. Two further losses: an action head fitted to one robot narrows outputs to that workspace and vocabulary, and web data records no forces or contact events. These echo the forgetting progressive networks addressed in 2016 (Section~\ref{sec:reuse}), the target for the retention metrics of Section~\ref{sec:metrics}.

\subsection{Reinforcement-Learning Post-Training}
\label{sec:posttrain}

Imitation-trained VLAs inherit their demonstrations' ceiling, so a third stage learns from deployment experience, importing Section~\ref{sec:offline2online}'s machinery. RIPT-VLA fine-tunes VLAs with reinforcement learning from sparse binary success, with large low-data gains \citep{tan2025ript}. $\pi^{*}_{0.6}$ trains with RECAP, conditioning the policy on advantage estimates from a value function fitted over autonomous experience, corrections, and demonstrations; its report states roughly doubled throughput and halved failure rates on its hardest tasks \citep{amin2025pistar}. Residual and diffusion-specific methods (Section~\ref{sec:offline2online}) serve the same role for smaller policies \citep{yuan2025policy, ren2024dppo}.

\subsection{World Models}
\label{sec:worldmodels}

A final family transfers predictive knowledge instead of behavior. The Dreamer line is its precursor: its world model is fitted to the agent's own experience, none web-pretrained.

A newer branch draws predictive knowledge from web-scale video. UniPi casts decision-making as text-conditioned video generation, an inverse-dynamics model extracting actions \citep{du2023learning}. GR-1 pretrained on video prediction then fine-tuned end to end; in ablation without video pretraining, CALVIN first-task success rises from 88.9\% to 94.9\% (seen) and 82.3\% to 85.4\% (unseen), average completed sequence length from 2.40 to 3.06 unseen \citep{wu2024unleashing}; GR-2 scaled to 38 million pretraining clips \citep{cheang2024gr2}. UniSim argued the learned model can replace the world for training and evaluation; Cosmos made it an open pretrain-then-post-train platform \citep{yang2023unisim, agarwal2025cosmos}. V-JEPA 2 makes the strongest case for action-free video: pretrained on over one million hours of internet video and post-trained with an action-conditioned head on under 62 hours of task-unlabeled robot interaction, it reportedly plans pick-and-place zero-shot on Franka arms in unseen labs \citep{assran2025vjepa}. Lighter variants exploit frozen DINOv2 features \citep{zhou2025dinowm} or a video diffusion model's internals \citep{hu2025video}. Here world models are transferred simulators (Section~\ref{sec:grounding}) and demonstration synthesizers (Section~\ref{sec:synthesis}), increasingly a common foundation under all five gaps.

\paragraph{Lessons.} Transfer is strongest for object recognition, language grounding, and task decomposition \citep{brohan2023rt2, ahn2022saycan} and weakest for dynamics, contact, and low-level control. Robot-data generalists and web-initialized VLAs share a pretrain-adapt structure but draw from different sources, and smaller models sometimes beat larger predecessors, so action representation, data composition, and training procedure matter alongside scale \citep{kim2024openvla, kim2025oft, pertsch2025fast}.

\section{Transfer Beyond Manipulation}
\label{sec:beyondmanip}

Manipulation and legged locomotion dominate the reviewed evidence. Four selected domains test the framework's reach; they do not constitute comprehensive domain surveys.

\paragraph{Aerial vehicles.} Randomization and privileged-teacher distillation support simulation-trained flight \citep{loquercio2020deep, loquercio2021learning}; residual dynamics fitted from real flights support high-performance racing \citep{kaufmann2023champion}. Grounding the simulator uses target data to improve the source, while the resulting policy is the transferred artifact. Controlled comparisons also distinguish task-level RL objectives from tracking-based optimal control \citep{song2023reaching}. RAPTOR distills many teachers into an adaptive student tested across physically distinct quadrotors \citep{eschmann2026raptor}. Body-rate or normalized actuator interfaces make cross-body reuse easier than manipulation with incompatible grippers.

\paragraph{Underwater vehicles.} WaterGAN uses a physical image-formation model to synthesize underwater appearance \citep{li2018watergan}. Control additionally faces drag, added mass, and currents; randomized simulation can transfer adaptive controllers to a physical vehicle \citep{chaffre2025sim}. Learning directly on one vehicle \citep{carlucho2018adaptive} is relevant background but does not alone establish transfer. Within the reviewed examples, corpus-scale cross-vehicle learning remains untested.

\paragraph{Soft robots.} Continuum dynamics, hysteresis, and material drift complicate the assumption of a fixed body. Target-fitted dynamics models and recurrent state estimators address modeling and sensing \citep{thuruthel2019model, thuruthel2019soft}; without a second domain, these are enabling tools rather than evidence of cross-body policy reuse. Simulator-trained morphology and control can transfer to fabricated crawlers \citep{schaff2023sim}. Transfer must specify whether the target is a new body, a fabricated design, or the same body after material change.

\paragraph{Multi-robot systems.} Decentralized policies transfer across numbers of neighbors when local inputs and shared action conventions remain compatible \citep{chen2017decentralized, long2018towards}. Simulation-to-hardware navigation supplies a separate reality-gap claim \citep{chen2017socially, fan2020distributed}. Changing team size also changes each agent's effective dynamics, including the behavior of other agents; the supervised bound does not model this nonstationarity.

These examples support the usefulness of gap and artifact labels while limiting the generalist-policy narrative. Shared low-level interfaces facilitate reuse; unavailable corpora and domain-specific collection constraints limit tests of large-scale pretraining. The absence of a manipulation-scale pipeline in these examples does not establish that such a pipeline would fail.

\section{Benchmarks and Evaluation}
\label{sec:eval}

\subsection{What Benchmarks Measure}
\label{sec:simbench}

Meta-World tests task-family generalization; RLBench supports varied manipulation; CALVIN measures language-conditioned sequences; LIBERO combines knowledge transfer and lifelong learning; ManiSkill3 supports GPU-parallel interaction \citep{yu2019metaworld, james2020rlbench, mees2022calvin, liu2023libero, tao2024maniskill3}. These instruments differ in task distribution, observation interface, and adaptation budget, so their success rates are not interchangeable measures of transfer.

Axis-controlled suites expose failures hidden by aggregate scores. THE COLOSSEUM perturbs visual and physical factors and reports large degradation when shifts combine \citep{pumacay2024colosseum}; STAR-Gen distinguishes visual, semantic, and behavioral generalization \citep{gao2025taxonomy}. Such tests operationalize specific shifts, but their scores do not estimate the divergence term in Eq.~\eqref{eq:bound}. Saturating a fixed suite, such as high fine-tuned LIBERO success \citep{kim2025oft}, leaves questions about held-out robots, new scenes, and target-data cost unresolved.

Evaluation interfaces can change what is measured without producing an execution error. \citet{he2026observation} show that observation pathways in manipulation benchmarks can drop or reorder state fields, confounding modality comparisons unless the state contract is matched. In a CALVIN scene-D case study, \citet{he2026benchmark} isolate a user-side export-horizon defect from the limitations of open-loop teacher replay under the benchmark's documented reset and chaining rules. These findings support checking observation content, action semantics, export horizon, and reset behavior before attributing a score difference to transfer; the CALVIN checklist remains a single-pipeline proposal.

\subsection{Physical Evaluation and Transfer Controls}
\label{sec:realeval}
\label{sec:measuring}

SIMPLER calibrates simulation to reproduce real-policy rankings \citep{li2024simpler}; AutoEval automates resets and scoring \citep{zhou2025autoeval}; RoboArena distributes physical comparisons across institutions \citep{atreya2025roboarena}. These address different costs and do not remove uncertainty in small-sample trials. Success bounds under deployment shift are themselves a research question \citep{vincent2024how}. A few successful rollouts may demonstrate feasibility while leaving moderate differences between policies unresolved.

To identify transfer, comparisons should vary the transferred artifact while matching target budget, architecture where applicable, and evaluation protocol. A target-only baseline is especially useful for fine-tuning; for zero-shot mechanisms, a matched ablation can supply the relevant contrast. Source training costs should be stated separately from target adaptation costs. Table~\ref{tab:headtohead} connects representative controlled studies to the decisions they inform. Their findings are within-study contrasts rather than a cross-paper ranking.

\begin{table}[pos=tbp]
\centering
\caption{Representative controlled comparisons and the limits of their implications.}
\label{tab:headtohead}
\small
\begin{tabularx}{\linewidth}{@{}>{\raggedright\arraybackslash}p{3.0cm}YY@{}}
\toprule
\textbf{Comparison} & \textbf{Finding} & \textbf{Limit} \\
\midrule
Visual pretraining vs. tuned target-only learning \citep{hansen2023pretraining, dasari2023unbiased} & Pretraining need not outperform a tuned baseline & Depends on task and training procedure \\
Mixture weights at fixed source pool \citep{hejna2024remix} & Reweighting changes downstream success substantially & No universal best mixture follows \\
Morphology graphs vs. attention \citep{kurin2021morphology} & Graph structure need not be the best inductive bias & Evidence mainly from simulation \\
Diversity vs. repeated demonstrations \citep{lin2025data, saxena2025what} & Target-relevant axes of diversity matter & Axes and evaluation distributions differ \\
Backbone adaptation with vs. without insulation \citep{driess2025knowledge} & Gradient routing and co-training affect retention & Restoration cost is not settled \\
\bottomrule
\end{tabularx}
\end{table}

\subsection{Evidence Census: Definition and Coding}
\label{sec:evidence}

We apply the E0--E4 rubric (Table~\ref{tab:rubric}) to one focal transfer claim per coded paper. It measures internal comparison control and, at E4, uncertainty reporting. It does not measure effect size, external validity, realism, or overall scientific value. A physical E3 study can therefore be more informative for deployment than a simulated E4 study. A negative result can receive the same grade as a positive result.

\begin{table}[pos=!tbp]
\centering
\caption{Experimental-control rubric for one focal transfer claim per paper. E4 adds uncertainty reporting to E3. E0--E3 describe comparison types, rather than cumulative requirements. Replication is assessed separately.}
\label{tab:rubric}
\small
\setlength{\tabcolsep}{5pt}
\begin{tabularx}{\linewidth}{@{}>{\raggedright\arraybackslash}p{2.8cm}Y@{}}
\toprule
\textbf{Level} & \textbf{What the paper must report for the transfer claim} \\
\midrule
E0 \enspace Demonstration & The system performs the target task. No comparison condition of any kind is reported. \\
\addlinespace[2pt]
E1 \enspace External comparison & Performance is compared against results reported elsewhere, or against a system with a different hardware, protocol, or training budget. The comparison exists but confounds the transferred artifact with everything else that differs. \\
\addlinespace[2pt]
E2 \enspace Internal, uncontrolled & The authors run an alternative condition themselves, but the conditions differ in more than the transferred artifact (model scale, data budget, or architecture), so the comparison does not isolate transfer. \\
\addlinespace[2pt]
E3 \enspace Controlled comparison & The compared conditions differ only in the transferred artifact, and consume the same target-side budget under the same evaluation protocol. For a zero-shot claim this means matched ablations of the transfer mechanism evaluated under one protocol; a target-only baseline is not required, since at zero target budget none exists. \\
\addlinespace[2pt]
E4 \enspace Controlled + dispersion & As E3, and the paper reports the number of trials or seeds together with a dispersion measure (a confidence interval, a standard deviation, or per-seed results) for the transfer claim itself. \\
\bottomrule
\end{tabularx}
\end{table}

Screening covered 218 of the 232 corpus works: 185 are represented in the coded subset and 33 are not. The five works added in November are outside the coded subset. A further 14 corpus works did not enter this screening. The numerical results below concern only the 185 retained rows. Gap and form follow the assignment rule of Section~\ref{sec:gaps}; the coding records the comparison supporting each grade. The primary coding was performed by one rater. We grade the strongest evidence for the specific claim cited here, rather than the strongest experiment anywhere in a paper, and assign the lower level when reporting is ambiguous. The full row-level table and machine-readable transcription accompany the article; Supplementary Material S1, Section S2, gives worked examples and boundary rules.

Independent replication is assessed separately: a group sharing no author with the originating paper must re-run its specific transfer contrast under matched or closely comparable conditions. A new task, platform, or pipeline, a benchmark score, or a paper sharing an original author is corroboration or extension under this definition. We did not identify such a replication for the 185 coded claims within the July 2026 coding corpus; this does not establish its absence elsewhere or later.

Table~\ref{tab:evidence} reports the counts by gap, transferred form, and era.

\begin{table}[pos=!tbp]
\centering
\caption{Counts by primary gap, form, and era in the fixed coded subset ($N=185$). E3+ and E4 are percentages of the row total. These describe curated coverage and reported control, not field-wide evidence quality.}
\label{tab:evidence}
\small
\setlength{\tabcolsep}{5pt}
\begin{tabularx}{\linewidth}{@{}Yrrrrrrrr@{}}
\toprule
 & \textbf{E0} & \textbf{E1} & \textbf{E2} & \textbf{E3} & \textbf{E4} & \textbf{N} & \textbf{E3+} & \textbf{E4+} \\
\midrule
\multicolumn{9}{@{}l}{\emph{By gap}}\\
\quad Simulation $\rightarrow$ reality & 0 & 3 & 6 & 31 & 3 & 43 & 79.1\% & 7.0\% \\
\quad Task $\rightarrow$ task & 0 & 0 & 2 & 30 & 6 & 38 & 94.7\% & 15.8\% \\
\quad Embodiment $\rightarrow$ embodiment & 0 & 0 & 4 & 26 & 1 & 31 & 87.1\% & 3.2\% \\
\quad Human $\rightarrow$ robot & 0 & 0 & 10 & 17 & 1 & 28 & 64.3\% & 3.6\% \\
\quad Web $\rightarrow$ robot & 1 & 0 & 14 & 26 & 4 & 45 & 66.7\% & 8.9\% \\
\addlinespace[3pt]
\multicolumn{9}{@{}l}{\emph{By transferred form}}\\
\quad Data & 0 & 1 & 7 & 31 & 3 & 42 & 81.0\% & 7.1\% \\
\quad Representations & 0 & 1 & 5 & 29 & 6 & 41 & 85.4\% & 14.6\% \\
\quad Parameters and policies & 1 & 1 & 10 & 36 & 3 & 51 & 76.5\% & 5.9\% \\
\quad Models and simulators & 0 & 0 & 9 & 15 & 2 & 26 & 65.4\% & 7.7\% \\
\quad Task conditioning & 0 & 0 & 5 & 19 & 1 & 25 & 80.0\% & 4.0\% \\
\addlinespace[3pt]
\multicolumn{9}{@{}l}{\emph{By era}}\\
\quad I \enspace ($\leq$2016) & 0 & 0 & 0 & 3 & 1 & 4 & --- & --- \\
\quad II \enspace (2017--2021) & 0 & 3 & 4 & 45 & 4 & 56 & 87.5\% & 7.1\% \\
\quad III (2022--2026) & 1 & 0 & 32 & 82 & 10 & 125 & 73.6\% & 8.0\% \\
\midrule
\textbf{All} & \textbf{1} & \textbf{3} & \textbf{36} & \textbf{130} & \textbf{15} & \textbf{185} & \textbf{78.4\%} & \textbf{8.1\%} \\
\bottomrule
\end{tabularx}
\end{table}

\subsection{Findings and Their Scope}

Of 185 coded claims, 145 (78.4\%) reach E3 or E4: the reported comparison isolates the transferred artifact under the rubric. Fifteen (8.1\%) additionally report trial or seed counts with dispersion for that comparison. The principal observed gap is therefore between comparison control and uncertainty reporting. An unreported dispersion measure cannot distinguish experiments that omitted repeats from experiments that ran repeats but omitted their results.

Task-to-task claims reach E3+ in 36 of 38 cases (94.7\%), compared with 18 of 28 (64.3\%) for human-to-robot claims. By form, representations reach E3+ in 35 of 41 cases (85.4\%) and models or simulators in 17 of 26 (65.4\%). These are small, selectively assembled groups. Different tasks, model scales, hardware costs, and publication practices could explain their differences; the coding does not isolate those causes.

Under the published grades, E2 accounts for 4 of 56 claims in 2017--2021 (7.1\%) and 32 of 125 in 2022--2026 (25.6\%). E4 shares are similar, 4 of 56 (7.1\%) and 10 of 125 (8.0\%). This is a descriptive association in the fixed coded subset. The later era contains many more web-transfer claims, and ambiguous budget reporting is conservatively assigned E2. Without an exhaustive sampling frame and independently documented ambiguity flags, these counts cannot establish a temporal decline in field-wide practice or a robust causal trend. Attributing gains in scale-changing system comparisons to transfer requires matched ablations.

\paragraph{Reporting priorities.} A reusable transfer result states the source artifact and adaptation procedure; separates source costs from target demonstrations, interactions, compute, and human interventions; includes a target-only or matched-ablation comparison; tests defined generalization axes; reports trial or seed counts with dispersion; and measures retention and negative transfer where applicable. E4 makes uncertainty visible but is not a substitute for checking leakage, task relevance, or physical validity. Automated and distributed evaluation can make these controls affordable across platforms rather than only within individual laboratories.

\section{Cross-Cutting Patterns}
\label{sec:patterns}

\subsection{Target-Relevant Diversity Rather Than Volume Alone}
\label{sec:pattern-diversity}

Across reality, embodiment, and web transfer, useful source coverage follows the target's variation. Randomization builds tolerance to the factors varied in simulation \citep{tobin2017domain, akkaya2019solving}; cross-robot studies distinguish object and environment diversity from repeated demonstrations \citep{lin2025data, saxena2025what}. Perceptual and morphology shifts can require different data structures, with paired cross-body demonstrations particularly useful for the latter \citep{yang2026data}. ReMix and shortcut-learning results explain why merely concatenating heterogeneous sources can hurt \citep{hejna2024remix, xing2025shortcut}.

This pattern does not imply that volume is irrelevant. It predicts that, at matched collection or training budgets, additional target-relevant coverage should be more useful than repeated examples of already covered variation. Its evidence is strongest in manipulation and sim-to-real training; it remains a hypothesis for underrepresented domains. A controlled result in which irrelevant diversity consistently outperforms targeted coverage would challenge the proposed explanation.

\subsection{Interfaces and Collection Infrastructure Shape Reuse}
\label{sec:pattern-apparatus}

Navigation reuses waypoint or velocity interfaces across platforms \citep{shah2023gnm, shah2023vint}; manipulation often has to learn or engineer comparable observation-action alignment \citep{chen2024mirage, ye2024latent}. UMI and ALOHA make collection interfaces part of the transfer mechanism \citep{chi2024universal, zhao2023learning}. GPU simulators similarly change which experiments are affordable \citep{makoviychuk2021isaac}. These examples suggest that source availability and interface compatibility help explain which transfer methods become practical.

This is a historical and mechanistic reading, not a matched comparison of infrastructure against algorithms. Collection changes data quality, quantity, and embodiment mismatch together. The hypothesis would be strengthened by controlled comparisons at equal human effort and weakened by consistent algorithmic gains after those collection and interface differences are removed.

\subsection{Controlled Re-Evaluation Can Narrow Capability Claims}
\label{sec:pattern-reeval}

A working system establishes that a combination is feasible; it need not establish which reused component caused the gain. Later visual-pretraining comparisons found benefits conditional on tuning and task \citep{majumdar2023where, hansen2023pretraining, dasari2023unbiased}; pooled-data studies found strong dependence on mixture weights and source shortcuts \citep{hejna2024remix, xing2025shortcut}. These are two documented episodes, rather than a universal law that later evidence must weaken earlier claims.

The census makes the distinction operational: E2 system comparisons confound the artifact with another change, whereas E3 isolates it and E4 adds dispersion. The era counts do not establish when re-evaluation usually arrives or whether attenuation is inevitable. They motivate earlier matched-budget studies of VLA post-training and world-model pretraining, including conditions where the original benefit is confirmed.

\section{Choosing a Transfer Strategy}
\label{sec:guidance}

The first choice is which source addresses the target's binding constraint: dynamics, behavior, embodiment alignment, or semantics. Table~\ref{tab:guidance} links common situations to a candidate mechanism and the baseline needed to test it. These are evidence-informed starting points. They are not validated decision rules, and the census's limited uncertainty reporting and replication constrain confidence in general prescriptions.

\begin{table}[pos=tbp]
\centering
\caption{Budget-conditioned starting points. Each choice requires the stated assumption and a comparison under the same target budget. Citations identify mechanisms, not a ranking of systems.}
\label{tab:guidance}
\small
\begin{tabularx}{\linewidth}{@{}>{\raggedright\arraybackslash}p{2.8cm}YY@{}}
\toprule
\textbf{Situation} & \textbf{Candidate and assumption} & \textbf{Baseline and main risk} \\
\midrule
Simulator; little real data & Ground measurable dynamics, randomize residuals, adapt from history \citep{hwangbo2019learning, kumar2021rma} & Randomization without grounding; unmodeled contact \\
Few target demonstrations & Fine-tune a generalist with a compatible sensor/action interface \citep{ghosh2024octo, kim2025oft} & Tuned target-only policy; forgetting or interface mismatch \\
Reward or corrections available & Online improvement or a residual over the pretrained policy \citep{luo2025precise, yuan2025policy} & Continued imitation at equal interaction cost; exploration failures \\
New body, no demonstrations & Visual alignment if dynamics and gripper are compatible; target-fitted simulation if observations are available \citep{chen2024mirage, torne2024reconciling} & Unchanged source policy or unfitted simulator; limited shift coverage \\
New body, some demonstrations & Pooled pretraining with target decoder or latent actions \citep{ye2024latent, bu2025univla} & Native-action fine-tuning; misweighted sources \\
Human task data & Retarget compatible motion, or transfer affordances and collect robot-compatible actions \citep{qiu2025humanoid, bahl2023affordances, chi2024universal} & Robot demonstrations at equal human effort; missing forces \\
Semantic generalization & Web-initialized VLA with retention controls, or language planning over existing skills \citep{driess2025knowledge, ahn2022saycan} & Uninsulated fine-tuning or skill-only policy; semantic loss or skill ceiling \\
\bottomrule
\end{tabularx}
\end{table}

\subsection{Selecting and Testing a Candidate}
\label{sec:thresholds}

Choose transfer using a small target pilot rather than an assumed demonstration threshold. Train the transferred policy and a tuned target-only policy at several affordable budgets, using the same control interface and evaluation. Read sample efficiency from the curves; an asymptotic gain and a cheaper start are different benefits. Report source-training costs separately so that reusable infrastructure and target deployment costs remain visible.

Simulator construction is most attractive when dominant dynamics are measurable and an asset will serve several tasks. Visual reconstruction can reduce scene-modeling effort \citep{torne2024reconciling, qureshi2025splatsim}; it does not establish contact fidelity. For pooled data, inspect per-source and per-axis validation before increasing volume. A source classifier can reveal identifying cues, but its accuracy is a shortcut warning rather than a validated predictor of transfer failure \citep{xing2025shortcut}.

\subsection{Diagnosing Failed Transfer}
\label{sec:recovery}

For conservative sim-to-real behavior, test whether randomization is unnecessarily broad; for abrupt failure, compare real and simulated trajectories for missing dynamics. For failed fine-tuning, inspect action units, frequency, chunking, observation compatibility, and baseline tuning before attributing the result to pretraining itself. For successful manipulation with poor semantic behavior, test retention on fixed vision-language probes and evaluate insulation or co-training \citep{driess2025knowledge}. For harmful pooling, reweight or remove sources under a fixed validation budget. In every case, size the evaluation for the difference that would change the deployment decision and report uncertainty; unstable measurements cannot select a strategy reliably.

\section{Open Challenges}
\label{sec:challenges}

\subsection{Predicting Transfer and Harmful Mixtures}
\label{sec:ch-theory}
\label{sec:ch-negative}

Supervised transferability scores suggest a model-selection route \citep{ding2024which}, but the reviewed robot-control methods do not provide a validated general predictor of closed-loop transfer. A useful predictor must distinguish appearance, dynamics, task, and body mismatch under a stated target budget, then be tested against actual adaptation outcomes. Mixture-level prediction is harder: reweighting, source shortcuts, and paired cross-body structure interact \citep{hejna2024remix, xing2025shortcut, yang2026data}. Recording failed transfers alongside successes is necessary to learn when additional data is harmful.

\subsection{Retention and Embodiment Compatibility}
\label{sec:ch-continual}
\label{sec:ch-embodiment}

Continual learning formalizes stability versus plasticity \citep{lesort2020continual}; task benchmarks and VLA fine-tuning expose both behavioral forgetting and semantic erosion \citep{liu2023libero, zhou2025chatvla, driess2025knowledge}. Post-training adds another opportunity for loss. Fixed retention probes and adaptation-budget curves should accompany each stage before architectural mitigations are compared.

Compatibility also needs controlled measurement. Benchmarks should vary morphology while holding perception fixed, then vary perception on the same bodies. This would separate action alignment from visual robustness. Force and tactile information require particular attention: they appear in some datasets and task-specific policies \citep{fang2023rh20t, qi2023general}, yet are not supplied by ordinary human or web video. Gains from morphology diversity in locomotion \citep{ai2025towards} do not establish equivalent transfer across dexterous hands or soft bodies.

\subsection{Control-Valid World Models and Safe Adaptation}
\label{sec:ch-worldmodel}
\label{sec:ch-safety}

Visual plausibility does not establish correct action consequences. World models used for planning, policy training, or evaluation \citep{yang2023unisim, assran2025vjepa} need physical-consistency tests and downstream physical validation: contact events, object permanence, and action-conditioned predictions are especially consequential. Otherwise training and evaluation may share the same model error.

Adapted policies also require deployment constraints that survive fine-tuning. Reachability shields and staged sim-to-lab transfer offer partial mechanisms \citep{hsu2023simtolabtoreal}; they do not certify arbitrary generalist-policy updates. Useful tests should evaluate the entire adaptation pipeline, including exploration, monitoring, and acceptance criteria for an updated policy.

\subsection{Evaluation That Accumulates Across Laboratories}
\label{sec:ch-eval}

Technical reports and model cards can document capabilities while leaving matched-budget reproduction difficult \citep{gemini2025robotics15, figure2025helix, ai2026pi07}. Shared physical facilities, automated resets, distributed comparisons, and calibrated simulation proxies address complementary costs \citep{atreya2025roboarena, zhou2025autoeval, li2024simpler}. The next step is to connect them with public protocols specifying generalization axes, source and target budgets, trial counts, uncertainty, and retention. Re-running a specific transfer contrast independently would test reproducibility more directly than another aggregate leaderboard score.

\section{Conclusion}
\label{sec:conclusion}

Robot learning transfers source-side structure that remains useful under the target's task, body, dynamics, and observation conditions. Five gaps and five transferred forms connect classical methods to generalist policies, post-training, and world models.

Across the reviewed evidence, target-relevant source coverage and compatible observation-action interfaces repeatedly matter. These conditions guide source choice but do not replace a target-budget comparison. Of 185 coded claims, 78.4\% isolate the transferred artifact under the rubric, while 8.1\% additionally report dispersion. The corpus therefore supports stronger conclusions about mechanisms and reporting gaps than about universal method rankings or changes in field-wide quality. Progress requires predicting harmful transfer, measuring retention, validating learned dynamics through physical consequences, and making controlled comparisons reproducible across laboratories.

\section*{Supplementary material}
Supplementary Material S1 documents the collection protocol, coding rules, and complete per-paper evidence table. Supplementary Data S2 supplies the 185-row coding, 232-work catalogue and bibliography, aggregate counts, and a script reproducing those counts.

\section*{CRediT authorship contribution statement}
\textbf{Yuanzhi He:} Conceptualization, Methodology, Investigation, Data curation, Formal analysis, Visualization, Writing -- original draft. \textbf{Jiayu Xu:} Writing -- review \& editing, Supervision.

\section*{Funding}
This research did not receive any specific grant from funding agencies in the public, commercial, or not-for-profit sectors.

\section*{Declaration of competing interest}
The authors declare that they have no known competing financial interests or personal relationships that could have appeared to influence the work reported in this paper.

\section*{Data availability}
The corpus catalogue, bibliography, evidence coding, and reproduction script are supplied in Supplementary Data S2. Supplementary Material S1 documents collection and coding and reproduces the complete row-level table. No new robot experiments were performed. The catalogue identifies works outside the coded set without assigning undocumented exclusion reasons.

\section*{References}
\hypertarget{refs}{}
\begin{CSLReferences}{1}{0}
\leavevmode\vadjust pre{\hypertarget{ref-bu2025agibot}{}}%
AgiBot World Contributors, Bu, Q., Cai, J., Chen, L., Cui, X., Ding, Y., Feng, S., Gao, S., He, X., Hu, X., Huang, X., Jiang, S., Jiang, Y., Jing, C., Li, H., Li, J., Liu, C., Liu, Y., Lu, Y., \ldots{} Zhu, J. (2025). {AgiBot World Colosseo}: A large-scale manipulation platform for scalable and intelligent embodied systems. \emph{IEEE/RSJ International Conference on Intelligent Robots and Systems (IROS)}. \url{https://arxiv.org/abs/2503.06669v5}

\leavevmode\vadjust pre{\hypertarget{ref-ai2025towards}{}}%
Ai, B., Dai, L., Bohlinger, N., Li, D., Mu, T., Wu, Z., Fay, K., Christensen, H. I., Peters, J., \& Su, H. (2025). Towards embodiment scaling laws in robot locomotion. \emph{Conference on Robot Learning (CoRL)}. \url{https://arxiv.org/abs/2505.05753v2}

\leavevmode\vadjust pre{\hypertarget{ref-andrychowicz2020learning}{}}%
Andrychowicz, M., Baker, B., Chociej, M., Józefowicz, R., McGrew, B., Pachocki, J., Petron, A., Plappert, M., Powell, G., Ray, A., Schneider, J., Sidor, S., Tobin, J., Welinder, P., Weng, L., \& Zaremba, W. (2020). Learning dexterous in-hand manipulation. \emph{The International Journal of Robotics Research}, \emph{39}(1), 3--20. \url{https://doi.org/10.1177/0278364919887447}

\leavevmode\vadjust pre{\hypertarget{ref-argall2009survey}{}}%
Argall, B. D., Chernova, S., Veloso, M., \& Browning, B. (2009). A survey of robot learning from demonstration. \emph{Robotics and Autonomous Systems}, \emph{57}(5), 469--483. \url{https://doi.org/10.1016/j.robot.2008.10.024}

\leavevmode\vadjust pre{\hypertarget{ref-arjovsky2019irm}{}}%
Arjovsky, M., Bottou, L., Gulrajani, I., \& Lopez-Paz, D. (2019). Invariant risk minimization. \emph{arXiv Preprint arXiv:1907.02893}. \url{https://arxiv.org/abs/1907.02893}

\leavevmode\vadjust pre{\hypertarget{ref-assran2025vjepa}{}}%
Assran, M., Bardes, A., Fan, D., Garrido, Q., Howes, R., Mojtaba, Komeili, Muckley, M., Rizvi, A., Roberts, C., Sinha, K., Zholus, A., Arnaud, S., Gejji, A., Martin, A., Hogan, F. R., Dugas, D., Bojanowski, P., Khalidov, V., \ldots{} Ballas, N. (2025). {V-JEPA 2}: Self-supervised video models enable understanding, prediction and planning. \emph{arXiv Preprint arXiv:2506.09985}. \url{https://arxiv.org/abs/2506.09985}

\leavevmode\vadjust pre{\hypertarget{ref-atreya2025roboarena}{}}%
Atreya, P., Pertsch, K., Lee, T., Kim, M. J., Jain, A., Kuramshin, A., Eppner, C., Neary, C., Hu, E., Ramos, F., Tremblay, J., Arora, K., Ellis, K., Macesanu, L., Villasevil, M. T., Leonard, M., Cho, M., Aslan, O., Dass, S., \ldots{} Levine, S. (2025). {RoboArena}: Distributed real-world evaluation of generalist robot policies. \emph{Conference on Robot Learning (CoRL)}. \url{https://arxiv.org/abs/2506.18123v2}

\leavevmode\vadjust pre{\hypertarget{ref-bahl2022human}{}}%
Bahl, S., Gupta, A., \& Pathak, D. (2022). Human-to-robot imitation in the wild. \emph{Robotics: Science and Systems (RSS)}. \url{https://arxiv.org/abs/2207.09450v1}

\leavevmode\vadjust pre{\hypertarget{ref-bahl2023affordances}{}}%
Bahl, S., Mendonca, R., Chen, L., Jain, U., \& Pathak, D. (2023). Affordances from human videos as a versatile representation for robotics. \emph{IEEE/CVF Conference on Computer Vision and Pattern Recognition (CVPR)}, 13778--13790. \url{https://arxiv.org/abs/2304.08488v1}

\leavevmode\vadjust pre{\hypertarget{ref-ball2023efficient}{}}%
Ball, P. J., Smith, L., Kostrikov, I., \& Levine, S. (2023). Efficient online reinforcement learning with offline data. \emph{International Conference on Machine Learning (ICML)}, 1577--1594. \url{https://proceedings.mlr.press/v202/ball23a.html}

\leavevmode\vadjust pre{\hypertarget{ref-barreto2017successor}{}}%
Barreto, A., Dabney, W., Munos, R., Hunt, J. J., Schaul, T., Hasselt, H. van, \& Silver, D. (2017). Successor features for transfer in reinforcement learning. \emph{Advances in Neural Information Processing Systems (NeurIPS)}. \url{https://arxiv.org/abs/1606.05312v2}

\leavevmode\vadjust pre{\hypertarget{ref-bendavid2010theory}{}}%
Ben-David, S., Blitzer, J., Crammer, K., Kulesza, A., Pereira, F., \& Vaughan, J. W. (2010). A theory of learning from different domains. \emph{Machine Learning}, \emph{79}(1--2), 151--175. \url{https://doi.org/10.1007/s10994-009-5152-4}

\leavevmode\vadjust pre{\hypertarget{ref-black2024pi0}{}}%
Black, K., Brown, N., Driess, D., Esmail, A., Equi, M., Finn, C., Fusai, N., Groom, L., Hausman, K., Ichter, B., Jakubczak, S., Jones, T., Ke, L., Levine, S., Li-Bell, A., Mothukuri, M., Nair, S., Pertsch, K., Shi, L. X., \ldots{} Zhilinsky, U. (2025). \(\pi_0\): A vision-language-action flow model for general robot control. \emph{Robotics: Science and Systems (RSS)}. \url{https://arxiv.org/abs/2410.24164v4}

\leavevmode\vadjust pre{\hypertarget{ref-bousmalis2018using}{}}%
Bousmalis, K., Irpan, A., Wohlhart, P., Bai, Y., Kelcey, M., Kalakrishnan, M., Downs, L., Ibarz, J., Pastor, P., Konolige, K., Levine, S., \& Vanhoucke, V. (2018). Using simulation and domain adaptation to improve efficiency of deep robotic grasping. \emph{IEEE International Conference on Robotics and Automation (ICRA)}. \url{https://doi.org/10.1109/icra.2018.8460875}

\leavevmode\vadjust pre{\hypertarget{ref-bousmalis2023robocat}{}}%
Bousmalis, K., Vezzani, G., Rao, D., Devin, C., Lee, A. X., Bauza, M., Davchev, T., Zhou, Y., Gupta, A., Raju, A., Laurens, A., Fantacci, C., Dalibard, V., Zambelli, M., Martins, M., Pevceviciute, R., Blokzijl, M., Denil, M., Batchelor, N., \ldots{} Heess, N. (2023). {RoboCat}: A self-improving generalist agent for robotic manipulation. \emph{Transactions on Machine Learning Research}. \url{https://arxiv.org/abs/2306.11706v2}

\leavevmode\vadjust pre{\hypertarget{ref-brohan2022rt1}{}}%
Brohan, A., Brown, N., Carbajal, J., Chebotar, Y., Dabis, J., Finn, C., Gopalakrishnan, K., Hausman, K., Herzog, A., Hsu, J., Ibarz, J., Ichter, B., Irpan, A., Jackson, T., Jesmonth, S., Joshi, N., Julian, R., Kalashnikov, D., Kuang, Y., \ldots{} Zitkovich, B. (2023). {RT-1}: Robotics transformer for real-world control at scale. \emph{Robotics: Science and Systems (RSS)}. \url{https://doi.org/10.15607/rss.2023.xix.025}

\leavevmode\vadjust pre{\hypertarget{ref-bu2025univla}{}}%
Bu, Q., Yang, Y., Cai, J., Gao, S., Ren, G., Yao, M., Luo, P., \& Li, H. (2025). Learning to act anywhere with task-centric latent actions. \emph{Robotics: Science and Systems (RSS)}. \url{https://arxiv.org/abs/2505.06111v3}

\leavevmode\vadjust pre{\hypertarget{ref-burns2024what}{}}%
Burns, K., Witzel, Z., Hamid, J. I., Yu, T., Finn, C., \& Hausman, K. (2025). What makes pre-trained visual representations successful for robust manipulation? \emph{Conference on Robot Learning (CoRL)}, 4525--4545. \url{https://proceedings.mlr.press/v270/burns25a.html}

\leavevmode\vadjust pre{\hypertarget{ref-carlucho2018adaptive}{}}%
Carlucho, I., Paula, M. D., Wang, S., Petillot, Y., \& Acosta, G. G. (2018). Adaptive low-level control of autonomous underwater vehicles using deep reinforcement learning. \emph{Robotics and Autonomous Systems}, \emph{107}, 71--86. \url{https://doi.org/10.1016/j.robot.2018.05.016}

\leavevmode\vadjust pre{\hypertarget{ref-chaffre2025sim}{}}%
Chaffre, T., Wheare, J., Lammas, A., Santos, P., Le Chenadec, G., Sammut, K., \& Clement, B. (2025). Sim-to-real transfer of adaptive control parameters for {AUV} stabilisation under current disturbance. \emph{The International Journal of Robotics Research}, \emph{44}(3), 407--430. \url{https://doi.org/10.1177/02783649241272115}

\leavevmode\vadjust pre{\hypertarget{ref-cheang2024gr2}{}}%
Cheang, C.-L., Chen, G., Jing, Y., Kong, T., Li, H., Li, Y., Liu, Y., Wu, H., Xu, J., Yang, Y., Zhang, H., \& Zhu, M. (2024). {GR-2}: A generative video-language-action model with web-scale knowledge for robot manipulation. \emph{arXiv Preprint arXiv:2410.06158}. \url{https://arxiv.org/abs/2410.06158}

\leavevmode\vadjust pre{\hypertarget{ref-chebotar2019closing}{}}%
Chebotar, Y., Handa, A., Makoviychuk, V., Macklin, M., Issac, J., Ratliff, N., \& Fox, D. (2019). Closing the sim-to-real loop: Adapting simulation randomization with real world experience. \emph{IEEE International Conference on Robotics and Automation (ICRA)}. \url{https://doi.org/10.1109/icra.2019.8793789}

\leavevmode\vadjust pre{\hypertarget{ref-chen2024mirage}{}}%
Chen, L. Y., Hari, K., Dharmarajan, K., Xu, C., Vuong, Q., \& Goldberg, K. (2024). Mirage: Cross-embodiment zero-shot policy transfer with cross-painting. \emph{Robotics: Science and Systems (RSS)}. \url{https://doi.org/10.15607/rss.2024.xx.069}

\leavevmode\vadjust pre{\hypertarget{ref-chen2024rovi}{}}%
Chen, L. Y., Xu, C., Dharmarajan, K., Cheng, R., Keutzer, K., Tomizuka, M., Vuong, Q., \& Goldberg, K. (2025). {RoVi-Aug}: Robot and viewpoint augmentation for cross-embodiment robot learning. \emph{Conference on Robot Learning (CoRL)}, 209--233. \url{https://proceedings.mlr.press/v270/chen25a.html}

\leavevmode\vadjust pre{\hypertarget{ref-chen2018hardware}{}}%
Chen, T., Murali, A., \& Gupta, A. (2018). Hardware conditioned policies for multi-robot transfer learning. \emph{Advances in Neural Information Processing Systems (NeurIPS)}. \url{https://arxiv.org/abs/1811.09864v2}

\leavevmode\vadjust pre{\hypertarget{ref-chen2017socially}{}}%
Chen, Y. F., Everett, M., Liu, M., \& How, J. P. (2017). Socially aware motion planning with deep reinforcement learning. \emph{IEEE/RSJ International Conference on Intelligent Robots and Systems (IROS)}, 1343--1350. \url{https://doi.org/10.1109/iros.2017.8202312}

\leavevmode\vadjust pre{\hypertarget{ref-chen2017decentralized}{}}%
Chen, Y. F., Liu, M., Everett, M., \& How, J. P. (2017). Decentralized non-communicating multiagent collision avoidance with deep reinforcement learning. \emph{IEEE International Conference on Robotics and Automation (ICRA)}, 285--292. \url{https://doi.org/10.1109/icra.2017.7989037}

\leavevmode\vadjust pre{\hypertarget{ref-cheng2024expressive}{}}%
Cheng, X., Ji, Y., Chen, J., Yang, R., Yang, G., \& Wang, X. (2024). Expressive whole-body control for humanoid robots. \emph{Robotics: Science and Systems (RSS)}. \url{https://doi.org/10.15607/rss.2024.xx.107}

\leavevmode\vadjust pre{\hypertarget{ref-chi2023diffusion}{}}%
Chi, C., Xu, Z., Feng, S., Cousineau, E., Du, Y., Burchfiel, B., Tedrake, R., \& Song, S. (2025). Diffusion policy: Visuomotor policy learning via action diffusion. \emph{The International Journal of Robotics Research}, \emph{44}(10--11), 1684--1704. \url{https://doi.org/10.1177/02783649241273668}

\leavevmode\vadjust pre{\hypertarget{ref-chi2024universal}{}}%
Chi, C., Xu, Z., Pan, C., Cousineau, E., Burchfiel, B., Feng, S., Tedrake, R., \& Song, S. (2024). Universal manipulation interface: In-the-wild robot teaching without in-the-wild robots. \emph{Robotics: Science and Systems (RSS)}. \url{https://arxiv.org/abs/2402.10329v3}

\leavevmode\vadjust pre{\hypertarget{ref-church2021tactile}{}}%
Church, A., Lloyd, J., Hadsell, R., \& Lepora, N. F. (2022). Tactile sim-to-real policy transfer via real-to-sim image translation. \emph{Conference on Robot Learning (CoRL)}, 1645--1654. \url{https://proceedings.mlr.press/v164/church22a.html}

\leavevmode\vadjust pre{\hypertarget{ref-cully2015robots}{}}%
Cully, A., Clune, J., Tarapore, D., \& Mouret, J.-B. (2015). Robots that can adapt like animals. \emph{Nature}, \emph{521}(7553), 503--507. \url{https://doi.org/10.1038/nature14422}

\leavevmode\vadjust pre{\hypertarget{ref-dasari2019robonet}{}}%
Dasari, S., Ebert, F., Tian, S., Nair, S., Bucher, B., Schmeckpeper, K., Singh, S., Levine, S., \& Finn, C. (2019). {RoboNet}: Large-scale multi-robot learning. \emph{Conference on Robot Learning (CoRL)}. \url{https://arxiv.org/abs/1910.11215v2}

\leavevmode\vadjust pre{\hypertarget{ref-dasari2023unbiased}{}}%
Dasari, S., Srirama, M. K., Jain, U., \& Gupta, A. (2023). An unbiased look at datasets for visuo-motor pre-training. \emph{Conference on Robot Learning (CoRL)}, 1183--1198. \url{https://proceedings.mlr.press/v229/dasari23a.html}

\leavevmode\vadjust pre{\hypertarget{ref-ding2024which}{}}%
Ding, Y., Jiang, B., Yu, A., Zheng, A., \& Liang, J. (2024). Which model to transfer? A survey on transferability estimation. \emph{arXiv Preprint arXiv:2402.15231}. \url{https://arxiv.org/abs/2402.15231}

\leavevmode\vadjust pre{\hypertarget{ref-doshi2024scaling}{}}%
Doshi, R., Walke, H. R., Mees, O., Dasari, S., \& Levine, S. (2025). Scaling cross-embodied learning: One policy for manipulation, navigation, locomotion and aviation. \emph{Conference on Robot Learning (CoRL)}, 496--512. \url{https://proceedings.mlr.press/v270/doshi25a.html}

\leavevmode\vadjust pre{\hypertarget{ref-driess2025knowledge}{}}%
Driess, D., Springenberg, J. T., Ichter, B., Yu, L., Li-Bell, A., Pertsch, K., Ren, A. Z., Walke, H., Vuong, Q., Shi, L. X., \& Levine, S. (2025). Knowledge insulating vision-language-action models: Train fast, run fast, generalize better. \emph{Advances in Neural Information Processing Systems (NeurIPS)}. \url{https://doi.org/10.52202/085713-3439}

\leavevmode\vadjust pre{\hypertarget{ref-driess2023palme}{}}%
Driess, D., Xia, F., Sajjadi, M. S. M., Lynch, C., Chowdhery, A., Ichter, B., Wahid, A., Tompson, J., Vuong, Q., Yu, T., Huang, W., Chebotar, Y., Sermanet, P., Duckworth, D., Levine, S., Vanhoucke, V., Hausman, K., Toussaint, M., Greff, K., \ldots{} Florence, P. (2023). {PaLM-E}: An embodied multimodal language model. \emph{International Conference on Machine Learning (ICML)}, 8469--8488. \url{https://proceedings.mlr.press/v202/driess23a.html}

\leavevmode\vadjust pre{\hypertarget{ref-du2023learning}{}}%
Du, Y., Yang, S., Dai, B., Dai, H., Nachum, O., Tenenbaum, J. B., Schuurmans, D., \& Abbeel, P. (2023). Learning universal policies via text-guided video generation. \emph{Advances in Neural Information Processing Systems (NeurIPS)}. \url{https://arxiv.org/abs/2302.00111v3}

\leavevmode\vadjust pre{\hypertarget{ref-duan2017oneshot}{}}%
Duan, Y., Andrychowicz, M., Stadie, B., Ho, J., Schneider, J., Sutskever, I., Abbeel, P., \& Zaremba, W. (2017). One-shot imitation learning. \emph{Advances in Neural Information Processing Systems (NeurIPS)}. \url{https://arxiv.org/abs/1703.07326v3}

\leavevmode\vadjust pre{\hypertarget{ref-duan2016rl}{}}%
Duan, Y., Schulman, J., Chen, X., Bartlett, P. L., Sutskever, I., \& Abbeel, P. (2016). {RL}\(^2\): Fast reinforcement learning via slow reinforcement learning. \emph{arXiv Preprint arXiv:1611.02779}. \url{https://arxiv.org/abs/1611.02779}

\leavevmode\vadjust pre{\hypertarget{ref-eschmann2026raptor}{}}%
Eschmann, J., Albani, D., \& Loianno, G. (2026). {RAPTOR}: A foundation policy for quadrotor control. \emph{Science Robotics}, \emph{11}(114), eaec1481. \url{https://doi.org/10.1126/scirobotics.aec1481}

\leavevmode\vadjust pre{\hypertarget{ref-eysenbach2019diversity}{}}%
Eysenbach, B., Gupta, A., Ibarz, J., \& Levine, S. (2019). Diversity is all you need: Learning skills without a reward function. \emph{International Conference on Learning Representations (ICLR)}. \url{https://arxiv.org/abs/1802.06070v6}

\leavevmode\vadjust pre{\hypertarget{ref-fan2020distributed}{}}%
Fan, T., Long, P., Liu, W., \& Pan, J. (2020). Distributed multi-robot collision avoidance via deep reinforcement learning for navigation in complex scenarios. \emph{The International Journal of Robotics Research}, \emph{39}(7), 856--892. \url{https://doi.org/10.1177/0278364920916531}

\leavevmode\vadjust pre{\hypertarget{ref-fang2023rh20t}{}}%
Fang, H.-S., Fang, H., Tang, Z., Liu, J., Wang, C., Wang, J., Zhu, H., \& Lu, C. (2024). {RH20T}: A comprehensive robotic dataset for learning diverse skills in one-shot. \emph{IEEE International Conference on Robotics and Automation (ICRA)}. \url{https://doi.org/10.1109/icra57147.2024.10611615}

\leavevmode\vadjust pre{\hypertarget{ref-figure2025helix}{}}%
Figure AI. (2025). \emph{Helix: A vision-language-action model for generalist humanoid control}. Technical report, \url{https://www.figure.ai/news/helix}, retrieved 20 July 2026. \url{https://www.figure.ai/news/helix}

\leavevmode\vadjust pre{\hypertarget{ref-finn2017model}{}}%
Finn, C., Abbeel, P., \& Levine, S. (2017). Model-agnostic meta-learning for fast adaptation of deep networks. \emph{International Conference on Machine Learning (ICML)}, 1126--1135. \url{https://arxiv.org/abs/1703.03400v3}

\leavevmode\vadjust pre{\hypertarget{ref-finn2017oneshot}{}}%
Finn, C., Yu, T., Zhang, T., Abbeel, P., \& Levine, S. (2017). One-shot visual imitation learning via meta-learning. \emph{Conference on Robot Learning (CoRL)}. \url{https://arxiv.org/abs/1709.04905v1}

\leavevmode\vadjust pre{\hypertarget{ref-firoozi2025foundation}{}}%
Firoozi, R., Tucker, J., Tian, S., Majumdar, A., Sun, J., Liu, W., Zhu, Y., Song, S., Kapoor, A., Hausman, K., Ichter, B., Driess, D., Wu, J., Lu, C., \& Schwager, M. (2025). Foundation models in robotics: Applications, challenges, and the future. \emph{The International Journal of Robotics Research}, \emph{44}(5), 701--739. \url{https://doi.org/10.1177/02783649241281508}

\leavevmode\vadjust pre{\hypertarget{ref-fu2024humanplus}{}}%
Fu, Z., Zhao, Q., Wu, Q., Wetzstein, G., \& Finn, C. (2025). {HumanPlus}: Humanoid shadowing and imitation from humans. \emph{Conference on Robot Learning (CoRL)}, 2828--2844. \url{https://proceedings.mlr.press/v270/fu25a.html}

\leavevmode\vadjust pre{\hypertarget{ref-fu2024mobile}{}}%
Fu, Z., Zhao, T. Z., \& Finn, C. (2025). Mobile {ALOHA}: Learning bimanual mobile manipulation with low-cost whole-body teleoperation. \emph{Conference on Robot Learning (CoRL)}, 4066--4083. \url{https://proceedings.mlr.press/v270/fu25b.html}

\leavevmode\vadjust pre{\hypertarget{ref-ganapathi2021correspondences}{}}%
Ganapathi, A., Sundaresan, P., Thananjeyan, B., Balakrishna, A., Seita, D., Grannen, J., Hwang, M., Hoque, R., Gonzalez, J. E., Jamali, N., Yamane, K., Iba, S., \& Goldberg, K. (2021). Learning dense visual correspondences in simulation to smooth and fold real fabrics. \emph{IEEE International Conference on Robotics and Automation (ICRA)}. \url{https://doi.org/10.1109/icra48506.2021.9561980}

\leavevmode\vadjust pre{\hypertarget{ref-gao2025taxonomy}{}}%
Gao, J., Belkhale, S., Dasari, S., Balakrishna, A., Shah, D., \& Sadigh, D. (2026). A taxonomy for evaluating generalist robot manipulation policies. \emph{IEEE Robotics and Automation Letters}, \emph{11}(3), 3182--3189. \url{https://doi.org/10.1109/lra.2026.3656785}

\leavevmode\vadjust pre{\hypertarget{ref-gemini2025robotics15}{}}%
Gemini Robotics Team, Abdolmaleki, A., Abeyruwan, S., Ainslie, J., Alayrac, J.-B., Arenas, M. G., Balakrishna, A., Batchelor, N., Bewley, A., Bingham, J., Bloesch, M., Bousmalis, K., Brakel, P., Brohan, A., Buschmann, T., Byravan, A., Cabi, S., Caluwaerts, K., Casarini, F., \ldots{} Zhou, Y. (2025). Gemini robotics 1.5: Pushing the frontier of generalist robots with advanced embodied reasoning, thinking, and motion transfer. \emph{arXiv Preprint arXiv:2510.03342}. \url{https://arxiv.org/abs/2510.03342}

\leavevmode\vadjust pre{\hypertarget{ref-gemini2025robotics}{}}%
Gemini Robotics Team, Abeyruwan, S., Ainslie, J., Alayrac, J.-B., Arenas, M. G., Armstrong, T., Balakrishna, A., Baruch, R., Bauza, M., Blokzijl, M., Bohez, S., Bousmalis, K., Brohan, A., Buschmann, T., Byravan, A., Cabi, S., Caluwaerts, K., Casarini, F., Chang, O., \ldots{} Zhou, Y. (2025). Gemini robotics: Bringing {AI} into the physical world. \emph{arXiv Preprint arXiv:2503.20020}. \url{https://arxiv.org/abs/2503.20020}

\leavevmode\vadjust pre{\hypertarget{ref-goyal2023rvt}{}}%
Goyal, A., Xu, J., Guo, Y., Blukis, V., Chao, Y.-W., \& Fox, D. (2023). {RVT}: Robotic view transformer for {3D} object manipulation. \emph{Conference on Robot Learning (CoRL)}, 694--710. \url{https://proceedings.mlr.press/v229/goyal23a.html}

\leavevmode\vadjust pre{\hypertarget{ref-gupta2022metamorph}{}}%
Gupta, A., Fan, L., Ganguli, S., \& Fei-Fei, L. (2022). {MetaMorph}: Learning universal controllers with transformers. \emph{International Conference on Learning Representations (ICLR)}. \url{https://arxiv.org/abs/2203.11931v1}

\leavevmode\vadjust pre{\hypertarget{ref-ha2021flingbot}{}}%
Ha, H., \& Song, S. (2022). {FlingBot}: The unreasonable effectiveness of dynamic manipulation for cloth unfolding. \emph{Conference on Robot Learning (CoRL)}, 24--33. \url{https://proceedings.mlr.press/v164/ha22a.html}

\leavevmode\vadjust pre{\hypertarget{ref-handa2023dextreme}{}}%
Handa, A., Allshire, A., Makoviychuk, V., Petrenko, A., Singh, R., Liu, J., Makoviichuk, D., Wyk, K. V., Zhurkevich, A., Sundaralingam, B., Narang, Y., Lafleche, J.-F., Fox, D., \& State, G. (2023). {DeXtreme}: Transfer of agile in-hand manipulation from simulation to reality. \emph{IEEE International Conference on Robotics and Automation (ICRA)}. \url{https://arxiv.org/abs/2210.13702v2}

\leavevmode\vadjust pre{\hypertarget{ref-hansen2023pretraining}{}}%
Hansen, N., Yuan, Z., Ze, Y., Mu, T., Rajeswaran, A., Su, H., Xu, H., \& Wang, X. (2023). On pre-training for visuo-motor control: Revisiting a learning-from-scratch baseline. \emph{International Conference on Machine Learning (ICML)}, 12511--12526. \url{https://proceedings.mlr.press/v202/hansen23c.html}

\leavevmode\vadjust pre{\hypertarget{ref-he2025asap}{}}%
He, T., Gao, J., Xiao, W., Zhang, Y., Wang, Z., Wang, J., Luo, Z., He, G., Sobanbab, N., Pan, C., Yi, Z., Qu, G., Kitani, K., Hodgins, J., Fan, L. "Jim", Zhu, Y., Liu, C., \& Shi, G. (2025). {ASAP}: Aligning simulation and real-world physics for learning agile humanoid whole-body skills. \emph{Robotics: Science and Systems (RSS)}. \url{https://arxiv.org/abs/2502.01143v3}

\leavevmode\vadjust pre{\hypertarget{ref-he2024omnih2o}{}}%
He, T., Luo, Z., He, X., Xiao, W., Zhang, C., Zhang, W., Kitani, K. M., Liu, C., \& Shi, G. (2025). {OmniH2O}: Universal and dexterous human-to-humanoid whole-body teleoperation and learning. \emph{Conference on Robot Learning (CoRL)}, 1516--1540. \url{https://proceedings.mlr.press/v270/he25b.html}

\leavevmode\vadjust pre{\hypertarget{ref-he2024learning}{}}%
He, T., Luo, Z., Xiao, W., Zhang, C., Kitani, K., Liu, C., \& Shi, G. (2024). Learning human-to-humanoid real-time whole-body teleoperation. \emph{IEEE/RSJ International Conference on Intelligent Robots and Systems (IROS)}. \url{https://arxiv.org/abs/2403.04436v1}

\leavevmode\vadjust pre{\hypertarget{ref-he2025hover}{}}%
He, T., Xiao, W., Lin, T., Luo, Z., Xu, Z., Jiang, Z., Kautz, J., Liu, C., Shi, G., Wang, X., Fan, L., \& Zhu, Y. (2025). {HOVER}: Versatile neural whole-body controller for humanoid robots. \emph{IEEE International Conference on Robotics and Automation (ICRA)}. \url{https://doi.org/10.1109/icra55743.2025.11128549}

\leavevmode\vadjust pre{\hypertarget{ref-he2026confidence}{}}%
He, Y. (2026). Detector confidence signals presence rather than occlusion in cluttered manipulation. \emph{arXiv Preprint arXiv:2607.13361}. \url{https://doi.org/10.48550/arXiv.2607.13361}

\leavevmode\vadjust pre{\hypertarget{ref-he2026sample}{}}%
He, Y., Romero-Cano, V., Patiño, J. J., Hernández, J. D., Sawtell, W., \& Colombo, G. (2027). Sample-efficient low-level motion planning for robotic manipulation tasks via zero-shot transfer learning. In L. Pasa, A. Lintas, I. V. Tetko, A. Micheli, N. Navarin, A. E. P. Villa, D. Tortorella, \& M. Polato (Eds.), \emph{Artificial neural networks and machine learning -- ICANN 2026} (Vol. 17093, pp. 179--190). Springer. \url{https://doi.org/10.1007/978-3-032-38404-1_15}

\leavevmode\vadjust pre{\hypertarget{ref-he2026decomposition}{}}%
He, Y., Romero-Cano, V., Xu, J., Hernández, J. D., \& Colombo, G. (2026). Enhancing deep reinforcement learning for robotic motion planning via task decomposition. \emph{arXiv Preprint arXiv:2610.35494}. \url{https://doi.org/10.48550/arXiv.2610.35494}

\leavevmode\vadjust pre{\hypertarget{ref-he2025fewshot}{}}%
He, Y., Wallbridge, C. D., Hernández, J. D., \& Colombo, G. B. (2025). Few-shot transfer learning for deep reinforcement learning on robotic manipulation tasks. \emph{Towards Autonomous Robotic Systems, Lecture Notes in Computer Science}, \emph{15052}, 85--92. \url{https://doi.org/10.1007/978-3-031-72062-8_8}

\leavevmode\vadjust pre{\hypertarget{ref-he2026benchmark}{}}%
He, Y., Wang, Q., \& Xu, J. (2026a). Benchmark-path fidelity in parallel robot-policy evaluation: A {CALVIN} case study. \emph{The Journal of Supercomputing}, \emph{82}, 810. \url{https://doi.org/10.1007/s11227-026-08965-w}

\leavevmode\vadjust pre{\hypertarget{ref-he2026observation}{}}%
He, Y., Wang, Q., \& Xu, J. (2026b). Observation-contract auditing for valid modality comparison in robot manipulation benchmarks. \emph{Scientific Reports}. \url{https://doi.org/10.1038/s41598-026-76821-1}

\leavevmode\vadjust pre{\hypertarget{ref-hejna2024remix}{}}%
Hejna, J., Bhateja, C. A., Jiang, Y., Pertsch, K., \& Sadigh, D. (2025). {Re-Mix}: Optimizing data mixtures for large scale imitation learning. \emph{Conference on Robot Learning (CoRL)}, 145--164. \url{https://proceedings.mlr.press/v270/hejna25a.html}

\leavevmode\vadjust pre{\hypertarget{ref-higuera2024sparsh}{}}%
Higuera, C., Sharma, A., Bodduluri, C. K., Fan, T., Lancaster, P., Kalakrishnan, M., Kaess, M., Boots, B., Lambeta, M., Wu, T., \& Mukadam, M. (2025). Sparsh: Self-supervised touch representations for vision-based tactile sensing. \emph{Conference on Robot Learning (CoRL)}, 885--915. \url{https://proceedings.mlr.press/v270/higuera25a.html}

\leavevmode\vadjust pre{\hypertarget{ref-hsu2023simtolabtoreal}{}}%
Hsu, K.-C., Ren, A. Z., Nguyen, D. P., Majumdar, A., \& Fisac, J. F. (2023). Sim-to-lab-to-real: Safe reinforcement learning with shielding and generalization guarantees. \emph{Artificial Intelligence}, \emph{314}, 103811. \url{https://doi.org/10.1016/j.artint.2022.103811}

\leavevmode\vadjust pre{\hypertarget{ref-hu2025video}{}}%
Hu, Y., Guo, Y., Wang, P., Chen, X., Wang, Y.-J., Zhang, J., Sreenath, K., Lu, C., \& Chen, J. (2025). Video prediction policy: A generalist robot policy with predictive visual representations. \emph{International Conference on Machine Learning (ICML)}. \url{https://arxiv.org/abs/2412.14803v2}

\leavevmode\vadjust pre{\hypertarget{ref-huang2022language}{}}%
Huang, W., Abbeel, P., Pathak, D., \& Mordatch, I. (2022). Language models as zero-shot planners: Extracting actionable knowledge for embodied agents. \emph{International Conference on Machine Learning (ICML)}. \url{https://arxiv.org/abs/2201.07207v2}

\leavevmode\vadjust pre{\hypertarget{ref-huang2020one}{}}%
Huang, W., Mordatch, I., \& Pathak, D. (2020). One policy to control them all: Shared modular policies for agent-agnostic control. \emph{International Conference on Machine Learning (ICML)}. \url{https://arxiv.org/abs/2007.04976v1}

\leavevmode\vadjust pre{\hypertarget{ref-huang2023voxposer}{}}%
Huang, W., Wang, C., Zhang, R., Li, Y., Wu, J., \& Fei-Fei, L. (2023). {VoxPoser}: Composable {3D} value maps for robotic manipulation with language models. \emph{Conference on Robot Learning (CoRL)}, 540--562. \url{https://proceedings.mlr.press/v229/huang23b.html}

\leavevmode\vadjust pre{\hypertarget{ref-huang2022inner}{}}%
Huang, W., Xia, F., Xiao, T., Chan, H., Liang, J., Florence, P., Zeng, A., Tompson, J., Mordatch, I., Chebotar, Y., Sermanet, P., Jackson, T., Brown, N., Luu, L., Levine, S., Hausman, K., \& Ichter, B. (2023). Inner monologue: Embodied reasoning through planning with language models. \emph{Conference on Robot Learning (CoRL)}, 1769--1782. \url{https://proceedings.mlr.press/v205/huang23c.html}

\leavevmode\vadjust pre{\hypertarget{ref-hwangbo2019learning}{}}%
Hwangbo, J., Lee, J., Dosovitskiy, A., Bellicoso, D., Tsounis, V., Koltun, V., \& Hutter, M. (2019). Learning agile and dynamic motor skills for legged robots. \emph{Science Robotics}, \emph{4}(26), eaau5872. \url{https://doi.org/10.1126/scirobotics.aau5872}

\leavevmode\vadjust pre{\hypertarget{ref-ahn2022saycan}{}}%
Ichter, B., Brohan, A., Chebotar, Y., Finn, C., Hausman, K., Herzog, A., Ho, D., Ibarz, J., Irpan, A., Jang, E., Julian, R., Kalashnikov, D., Levine, S., Lu, Y., Parada, C., Rao, K., Sermanet, P., Toshev, A. T., Vanhoucke, V., \ldots{} Fu, C. K. (2023). Do as {I} can, not as {I} say: Grounding language in robotic affordances. \emph{Conference on Robot Learning (CoRL)}, 287--318. \url{https://proceedings.mlr.press/v205/ichter23a.html}

\leavevmode\vadjust pre{\hypertarget{ref-jain2024vid2robot}{}}%
Jain, V., Attarian, M., Joshi, N. J., Wahid, A., Driess, D., Vuong, Q., Sanketi, P. R., Sermanet, P., Welker, S., Chan, C., Gilitschenski, I., Bisk, Y., \& Dwibedi, D. (2024). {Vid2Robot}: End-to-end video-conditioned policy learning with cross-attention transformers. \emph{Robotics: Science and Systems (RSS)}. \url{https://arxiv.org/abs/2403.12943v2}

\leavevmode\vadjust pre{\hypertarget{ref-james2018taskembedded}{}}%
James, S., Bloesch, M., \& Davison, A. J. (2018). Task-embedded control networks for few-shot imitation learning. \emph{Conference on Robot Learning (CoRL)}. \url{https://arxiv.org/abs/1810.03237v1}

\leavevmode\vadjust pre{\hypertarget{ref-james2020rlbench}{}}%
James, S., Ma, Z., Arrojo, D. R., \& Davison, A. J. (2020). {RLBench}: The robot learning benchmark \& learning environment. \emph{IEEE Robotics and Automation Letters}, \emph{5}(2), 3019--3026. \url{https://doi.org/10.1109/lra.2020.2974707}

\leavevmode\vadjust pre{\hypertarget{ref-james2019sim}{}}%
James, S., Wohlhart, P., Kalakrishnan, M., Kalashnikov, D., Irpan, A., Ibarz, J., Levine, S., Hadsell, R., \& Bousmalis, K. (2019). Sim-to-real via sim-to-sim: Data-efficient robotic grasping via randomized-to-canonical adaptation networks. \emph{IEEE/CVF Conference on Computer Vision and Pattern Recognition (CVPR)}. \url{https://doi.org/10.1109/cvpr.2019.01291}

\leavevmode\vadjust pre{\hypertarget{ref-jang2022bcz}{}}%
Jang, E., Irpan, A., Khansari, M., Kappler, D., Ebert, F., Lynch, C., Levine, S., \& Finn, C. (2022). {BC-Z}: Zero-shot task generalization with robotic imitation learning. \emph{Conference on Robot Learning (CoRL)}, 991--1002. \url{https://proceedings.mlr.press/v164/jang22a.html}

\leavevmode\vadjust pre{\hypertarget{ref-jang2025dreamgen}{}}%
Jang, J., Ye, S., Lin, Z., Xiang, J., Bjorck, J., Fang, Y., Hu, F., Huang, S., Kundalia, K., Lin, Y.-C., Magne, L., Mandlekar, A., Narayan, A., Tan, Y. L., Wang, G., Wang, J., Wang, Q., Xu, Y., Zeng, X., \ldots{} Fan, L. (2025). {DreamGen}: Unlocking generalization in robot learning through video world models. \emph{Conference on Robot Learning (CoRL)}. \url{https://arxiv.org/abs/2505.12705v2}

\leavevmode\vadjust pre{\hypertarget{ref-jaquier2025transfer}{}}%
Jaquier, N., Welle, M. C., Gams, A., Yao, K., Fichera, B., Billard, A., Ude, A., Asfour, T., \& Kragic, D. (2025). Transfer learning in robotics: An upcoming breakthrough? A review of promises and challenges. \emph{The International Journal of Robotics Research}, \emph{44}(3), 465--485. \url{https://doi.org/10.1177/02783649241273565}

\leavevmode\vadjust pre{\hypertarget{ref-jiang2025robots}{}}%
Jiang, G., Sun, Y., Huang, T., Li, H., Liang, Y., \& Xu, H. (2025). Robots pre-train robots: Manipulation-centric robotic representation from large-scale robot datasets. \emph{International Conference on Learning Representations (ICLR)}. \url{https://arxiv.org/abs/2410.22325v2}

\leavevmode\vadjust pre{\hypertarget{ref-jiang2024dexmimicgen}{}}%
Jiang, Z., Xie, Y., Lin, K., Xu, Z., Wan, W., Mandlekar, A., Fan, L., \& Zhu, Y. (2025). {DexMimicGen}: Automated data generation for bimanual dexterous manipulation via imitation learning. \emph{IEEE International Conference on Robotics and Automation (ICRA)}. \url{https://doi.org/10.1109/icra55743.2025.11127809}

\leavevmode\vadjust pre{\hypertarget{ref-johannink2019residual}{}}%
Johannink, T., Bahl, S., Nair, A., Luo, J., Kumar, A., Loskyll, M., Ojea, J. A., Solowjow, E., \& Levine, S. (2019). Residual reinforcement learning for robot control. \emph{IEEE International Conference on Robotics and Automation (ICRA)}. \url{https://doi.org/10.1109/icra.2019.8794127}

\leavevmode\vadjust pre{\hypertarget{ref-kalashnikov2018qtopt}{}}%
Kalashnikov, D., Irpan, A., Pastor, P., Ibarz, J., Herzog, A., Jang, E., Quillen, D., Holly, E., Kalakrishnan, M., Vanhoucke, V., \& Levine, S. (2018). {QT-Opt}: Scalable deep reinforcement learning for vision-based robotic manipulation. \emph{Conference on Robot Learning (CoRL)}. \url{https://arxiv.org/abs/1806.10293v3}

\leavevmode\vadjust pre{\hypertarget{ref-kalashnikov2021mt}{}}%
Kalashnikov, D., Varley, J., Chebotar, Y., Swanson, B., Jonschkowski, R., Finn, C., Levine, S., \& Hausman, K. (2021). {MT-Opt}: Continuous multi-task robotic reinforcement learning at scale. \emph{arXiv Preprint arXiv:2104.08212}. \url{https://arxiv.org/abs/2104.08212}

\leavevmode\vadjust pre{\hypertarget{ref-karamcheti2023languagedriven}{}}%
Karamcheti, S., Nair, S., Chen, A. S., Kollar, T., Finn, C., Sadigh, D., \& Liang, P. (2023). Language-driven representation learning for robotics. \emph{Robotics: Science and Systems (RSS)}. \url{https://arxiv.org/abs/2302.12766v1}

\leavevmode\vadjust pre{\hypertarget{ref-kareer2024egomimic}{}}%
Kareer, S., Patel, D., Punamiya, R., Mathur, P., Cheng, S., Wang, C., Hoffman, J., \& Xu, D. (2025). {EgoMimic}: Scaling imitation learning via egocentric video. \emph{IEEE International Conference on Robotics and Automation (ICRA)}. \url{https://doi.org/10.1109/icra55743.2025.11127989}

\leavevmode\vadjust pre{\hypertarget{ref-kaufmann2023champion}{}}%
Kaufmann, E., Bauersfeld, L., Loquercio, A., Müller, M., Koltun, V., \& Scaramuzza, D. (2023). Champion-level drone racing using deep reinforcement learning. \emph{Nature}, \emph{620}(7976), 982--987. \url{https://doi.org/10.1038/s41586-023-06419-4}

\leavevmode\vadjust pre{\hypertarget{ref-kearns2002near}{}}%
Kearns, M., \& Singh, S. (2002). Near-optimal reinforcement learning in polynomial time. \emph{Machine Learning}, \emph{49}(2--3), 209--232. \url{https://doi.org/10.1023/a:1017984413808}

\leavevmode\vadjust pre{\hypertarget{ref-khazatsky2024droid}{}}%
Khazatsky, A., Pertsch, K., Nair, S., Balakrishna, A., Dasari, S., Karamcheti, S., Nasiriany, S., Srirama, M. K., Chen, L. Y., Ellis, K., Fagan, P. D., Hejna, J., Itkina, M., Lepert, M., Ma, Y. J., Miller, P. T., Wu, J., Belkhale, S., Dass, S., \ldots{} Finn, C. (2024). {DROID}: A large-scale in-the-wild robot manipulation dataset. \emph{Robotics: Science and Systems (RSS)}. \url{https://arxiv.org/abs/2403.12945v2}

\leavevmode\vadjust pre{\hypertarget{ref-kim2025oft}{}}%
Kim, M. J., Finn, C., \& Liang, P. (2025). Fine-tuning vision-language-action models: Optimizing speed and success. \emph{Robotics: Science and Systems (RSS)}. \url{https://doi.org/10.15607/rss.2025.xxi.017}

\leavevmode\vadjust pre{\hypertarget{ref-kim2024openvla}{}}%
Kim, M. J., Pertsch, K., Karamcheti, S., Xiao, T., Balakrishna, A., Nair, S., Rafailov, R., Foster, E. P., Sanketi, P. R., Vuong, Q., Kollar, T., Burchfiel, B., Tedrake, R., Sadigh, D., Levine, S., Liang, P., \& Finn, C. (2025). {OpenVLA}: An open-source vision-language-action model. \emph{Conference on Robot Learning (CoRL)}, 2679--2713. \url{https://proceedings.mlr.press/v270/kim25c.html}

\leavevmode\vadjust pre{\hypertarget{ref-kober2013reinforcement}{}}%
Kober, J., Bagnell, J. A., \& Peters, J. (2013). Reinforcement learning in robotics: A survey. \emph{The International Journal of Robotics Research}, \emph{32}(11), 1238--1274. \url{https://doi.org/10.1177/0278364913495721}

\leavevmode\vadjust pre{\hypertarget{ref-kumar2021rma}{}}%
Kumar, A., Fu, Z., Pathak, D., \& Malik, J. (2021). {RMA}: Rapid motor adaptation for legged robots. \emph{Robotics: Science and Systems (RSS)}. \url{https://arxiv.org/abs/2107.04034v1}

\leavevmode\vadjust pre{\hypertarget{ref-kurin2021morphology}{}}%
Kurin, V., Igl, M., Rocktäschel, T., Boehmer, W., \& Whiteson, S. (2021). My body is a cage: The role of morphology in graph-based incompatible control. \emph{International Conference on Learning Representations (ICLR)}. \url{https://arxiv.org/abs/2010.01856v2}

\leavevmode\vadjust pre{\hypertarget{ref-laskin2023incontext}{}}%
Laskin, M., Wang, L., Oh, J., Parisotto, E., Spencer, S., Steigerwald, R., Strouse, D., Hansen, S., Filos, A., Brooks, E., Gazeau, M., Sahni, H., Singh, S., \& Mnih, V. (2023). In-context reinforcement learning with algorithm distillation. \emph{International Conference on Learning Representations (ICLR)}. \url{https://arxiv.org/abs/2210.14215v1}

\leavevmode\vadjust pre{\hypertarget{ref-lazaric2012transfer}{}}%
Lazaric, A. (2012). Transfer in reinforcement learning: A framework and a survey. In \emph{Reinforcement learning: State of the art} (pp. 143--173). Springer. \url{https://doi.org/10.1007/978-3-642-27645-3_5}

\leavevmode\vadjust pre{\hypertarget{ref-lee2020learning}{}}%
Lee, J., Hwangbo, J., Wellhausen, L., Koltun, V., \& Hutter, M. (2020). Learning quadrupedal locomotion over challenging terrain. \emph{Science Robotics}, \emph{5}(47), eabc5986. \url{https://doi.org/10.1126/scirobotics.abc5986}

\leavevmode\vadjust pre{\hypertarget{ref-lepert2025phantom}{}}%
Lepert, M., Fang, J., \& Bohg, J. (2025). Phantom: Training robots without robots using only human videos. \emph{Conference on Robot Learning (CoRL)}. \url{https://arxiv.org/abs/2503.00779v2}

\leavevmode\vadjust pre{\hypertarget{ref-lesort2020continual}{}}%
Lesort, T., Lomonaco, V., Stoian, A., Maltoni, D., Filliat, D., \& Díaz-Rodríguez, N. (2020). Continual learning for robotics: Definition, framework, learning strategies, opportunities and challenges. \emph{Information Fusion}, \emph{58}, 52--68. \url{https://doi.org/10.1016/j.inffus.2019.12.004}

\leavevmode\vadjust pre{\hypertarget{ref-li2018watergan}{}}%
Li, J., Skinner, K. A., Eustice, R. M., \& Johnson-Roberson, M. (2018). {WaterGAN}: Unsupervised generative network to enable real-time color correction of monocular underwater images. \emph{IEEE Robotics and Automation Letters}, \emph{3}(1), 387--394. \url{https://doi.org/10.1109/lra.2017.2730363}

\leavevmode\vadjust pre{\hypertarget{ref-li2024okami}{}}%
Li, J., Zhu, Y., Xie, Y., Jiang, Z., Seo, M., Pavlakos, G., \& Zhu, Y. (2025a). {OKAMI}: Teaching humanoid robots manipulation skills through single video imitation. \emph{Conference on Robot Learning (CoRL)}, 299--317. \url{https://proceedings.mlr.press/v270/li25a.html}

\leavevmode\vadjust pre{\hypertarget{ref-li2024simpler}{}}%
Li, X., Hsu, K., Gu, J., Mees, O., Pertsch, K., Walke, H. R., Fu, C., Lunawat, I., Sieh, I., Kirmani, S., Levine, S., Wu, J., Finn, C., Su, H., Vuong, Q., \& Xiao, T. (2025b). Evaluating real-world robot manipulation policies in simulation. \emph{Conference on Robot Learning (CoRL)}, 3705--3728. \url{https://proceedings.mlr.press/v270/li25c.html}

\leavevmode\vadjust pre{\hypertarget{ref-li2026robovlm}{}}%
Li, X., Li, P., Qian, L., Liu, M., Wang, D., Liu, J., Kang, B., Ma, X., Wang, X., Guo, D., Kong, T., Zhang, H., \& Liu, H. (2026). What matters in building vision-language-action models for generalist robots. \emph{Nature Machine Intelligence}, \emph{8}(2), 158--172. \url{https://doi.org/10.1038/s42256-025-01168-7}

\leavevmode\vadjust pre{\hypertarget{ref-liang2023code}{}}%
Liang, J., Huang, W., Xia, F., Xu, P., Hausman, K., Ichter, B., Florence, P., \& Zeng, A. (2023). Code as policies: Language model programs for embodied control. \emph{IEEE International Conference on Robotics and Automation (ICRA)}. \url{https://doi.org/10.1109/icra48891.2023.10160591}

\leavevmode\vadjust pre{\hypertarget{ref-lin2025data}{}}%
Lin, F., Hu, Y., Sheng, P., Wen, C., You, J., \& Gao, Y. (2025). Data scaling laws in imitation learning for robotic manipulation. \emph{International Conference on Learning Representations (ICLR)}. \url{https://arxiv.org/abs/2410.18647v4}

\leavevmode\vadjust pre{\hypertarget{ref-liu2023libero}{}}%
Liu, B., Zhu, Y., Gao, C., Feng, Y., Liu, Q., Zhu, Y., \& Stone, P. (2023). {LIBERO}: Benchmarking knowledge transfer for lifelong robot learning. \emph{NeurIPS Datasets and Benchmarks Track}. \url{https://arxiv.org/abs/2306.03310v2}

\leavevmode\vadjust pre{\hypertarget{ref-liu2024rdt}{}}%
Liu, S., Wu, L., Li, B., Tan, H., Chen, H., Wang, Z., Xu, K., Su, H., \& Zhu, J. (2025). {RDT-1B}: A diffusion foundation model for bimanual manipulation. \emph{International Conference on Learning Representations (ICLR)}. \url{https://arxiv.org/abs/2410.07864v2}

\leavevmode\vadjust pre{\hypertarget{ref-liu2024maniwav}{}}%
Liu, Z., Chi, C., Cousineau, E., Kuppuswamy, N., Burchfiel, B., \& Song, S. (2025). {ManiWAV}: Learning robot manipulation from in-the-wild audio-visual data. \emph{Conference on Robot Learning (CoRL)}, 947--962. \url{https://proceedings.mlr.press/v270/liu25c.html}

\leavevmode\vadjust pre{\hypertarget{ref-long2018towards}{}}%
Long, P., Fan, T., Liao, X., Liu, W., Zhang, H., \& Pan, J. (2018). Towards optimally decentralized multi-robot collision avoidance via deep reinforcement learning. \emph{IEEE International Conference on Robotics and Automation (ICRA)}, 6252--6259. \url{https://doi.org/10.1109/icra.2018.8461113}

\leavevmode\vadjust pre{\hypertarget{ref-loquercio2020deep}{}}%
Loquercio, A., Kaufmann, E., Ranftl, R., Dosovitskiy, A., Koltun, V., \& Scaramuzza, D. (2020). Deep drone racing: From simulation to reality with domain randomization. \emph{IEEE Transactions on Robotics}, \emph{36}(1), 1--14. \url{https://doi.org/10.1109/tro.2019.2942989}

\leavevmode\vadjust pre{\hypertarget{ref-loquercio2021learning}{}}%
Loquercio, A., Kaufmann, E., Ranftl, R., Müller, M., Koltun, V., \& Scaramuzza, D. (2021). Learning high-speed flight in the wild. \emph{Science Robotics}, \emph{6}(59), eabg5810. \url{https://doi.org/10.1126/scirobotics.abg5810}

\leavevmode\vadjust pre{\hypertarget{ref-luo2024serl}{}}%
Luo, J., Hu, Z., Xu, C., Tan, Y. L., Berg, J., Sharma, A., Schaal, S., Finn, C., Gupta, A., \& Levine, S. (2024). {SERL}: A software suite for sample-efficient robotic reinforcement learning. \emph{IEEE International Conference on Robotics and Automation (ICRA)}. \url{https://arxiv.org/abs/2401.16013v4}

\leavevmode\vadjust pre{\hypertarget{ref-luo2025precise}{}}%
Luo, J., Xu, C., Wu, J., \& Levine, S. (2025). Precise and dexterous robotic manipulation via human-in-the-loop reinforcement learning. \emph{Science Robotics}, \emph{10}(105), eads5033. \url{https://doi.org/10.1126/scirobotics.ads5033}

\leavevmode\vadjust pre{\hypertarget{ref-ma2024eureka}{}}%
Ma, Y. J., Liang, W., Wang, G., Huang, D.-A., Bastani, O., Jayaraman, D., Zhu, Y., Fan, L., \& Anandkumar, A. (2024). Eureka: Human-level reward design via coding large language models. \emph{International Conference on Learning Representations (ICLR)}. \url{https://arxiv.org/abs/2310.12931v2}

\leavevmode\vadjust pre{\hypertarget{ref-ma2024dreureka}{}}%
Ma, Y. J., Liang, W., Wang, H.-J., Wang, S., Zhu, Y., Fan, L., Bastani, O., \& Jayaraman, D. (2024). DrEureka: Language model guided sim-to-real transfer. \emph{Robotics: Science and Systems (RSS)}. \url{https://arxiv.org/abs/2406.01967v1}

\leavevmode\vadjust pre{\hypertarget{ref-ma2023vip}{}}%
Ma, Y. J., Sodhani, S., Jayaraman, D., Bastani, O., Kumar, V., \& Zhang, A. (2023). {VIP}: Towards universal visual reward and representation via value-implicit pre-training. \emph{International Conference on Learning Representations (ICLR)}. \url{https://arxiv.org/abs/2210.00030v2}

\leavevmode\vadjust pre{\hypertarget{ref-majumdar2023where}{}}%
Majumdar, A., Yadav, K., Arnaud, S., Ma, J., Chen, C., Silwal, S., Jain, A., Berges, V.-P., Wu, T., Vakil, J., Abbeel, P., Malik, J., Batra, D., Lin, Y., Maksymets, O., Rajeswaran, A., \& Meier, F. (2023). Where are we in the search for an artificial visual cortex for embodied intelligence? \emph{Advances in Neural Information Processing Systems (NeurIPS)}. \url{https://doi.org/10.52202/075280-0031}

\leavevmode\vadjust pre{\hypertarget{ref-makoviychuk2021isaac}{}}%
Makoviychuk, V., Wawrzyniak, L., Guo, Y., Lu, M., Storey, K., Macklin, M., Hoeller, D., Rudin, N., Allshire, A., Handa, A., \& State, G. (2021). Isaac {G}ym: High performance {GPU}-based physics simulation for robot learning. \emph{NeurIPS Datasets and Benchmarks Track}. \url{https://arxiv.org/abs/2108.10470v2}

\leavevmode\vadjust pre{\hypertarget{ref-mandlekar2023mimicgen}{}}%
Mandlekar, A., Nasiriany, S., Wen, B., Akinola, I., Narang, Y., Fan, L., Zhu, Y., \& Fox, D. (2023). {MimicGen}: A data generation system for scalable robot learning using human demonstrations. \emph{Conference on Robot Learning (CoRL)}, 1820--1864. \url{https://proceedings.mlr.press/v229/mandlekar23a.html}

\leavevmode\vadjust pre{\hypertarget{ref-mansour2009domain}{}}%
Mansour, Y., Mohri, M., \& Rostamizadeh, A. (2009). Domain adaptation: Learning bounds and algorithms. \emph{Conference on Learning Theory (COLT)}. \url{https://arxiv.org/abs/0902.3430v3}

\leavevmode\vadjust pre{\hypertarget{ref-matas2018sim}{}}%
Matas, J., James, S., \& Davison, A. J. (2018). Sim-to-real reinforcement learning for deformable object manipulation. \emph{Conference on Robot Learning (CoRL)}. \url{https://arxiv.org/abs/1806.07851v2}

\leavevmode\vadjust pre{\hypertarget{ref-mees2022calvin}{}}%
Mees, O., Hermann, L., Rosete-Beas, E., \& Burgard, W. (2022). {CALVIN}: A benchmark for language-conditioned policy learning for long-horizon robot manipulation tasks. \emph{IEEE Robotics and Automation Letters}, \emph{7}(3), 7327--7334. \url{https://doi.org/10.1109/lra.2022.3180108}

\leavevmode\vadjust pre{\hypertarget{ref-muratore2021data}{}}%
Muratore, F., Eilers, C., Gienger, M., \& Peters, J. (2021). Data-efficient domain randomization with bayesian optimization. \emph{IEEE Robotics and Automation Letters}, \emph{6}(2), 911--918. \url{https://doi.org/10.1109/lra.2021.3052391}

\leavevmode\vadjust pre{\hypertarget{ref-muratore2022robot}{}}%
Muratore, F., Ramos, F., Turk, G., Yu, W., Gienger, M., \& Peters, J. (2022). Robot learning from randomized simulations: A review. \emph{Frontiers in Robotics and AI}, \emph{9}, 799893. \url{https://doi.org/10.3389/frobt.2022.799893}

\leavevmode\vadjust pre{\hypertarget{ref-nagabandi2019learning}{}}%
Nagabandi, A., Clavera, I., Liu, S., Fearing, R. S., Abbeel, P., Levine, S., \& Finn, C. (2019). Learning to adapt in dynamic, real-world environments through meta-reinforcement learning. \emph{International Conference on Learning Representations (ICLR)}. \url{https://arxiv.org/abs/1803.11347v6}

\leavevmode\vadjust pre{\hypertarget{ref-nair2020awac}{}}%
Nair, A., Gupta, A., Dalal, M., \& Levine, S. (2020). {AWAC}: Accelerating online reinforcement learning with offline datasets. \emph{arXiv Preprint arXiv:2006.09359}. \url{https://arxiv.org/abs/2006.09359}

\leavevmode\vadjust pre{\hypertarget{ref-nair2022r3m}{}}%
Nair, S., Rajeswaran, A., Kumar, V., Finn, C., \& Gupta, A. (2023). {R3M}: A universal visual representation for robot manipulation. \emph{Conference on Robot Learning (CoRL)}, 892--909. \url{https://proceedings.mlr.press/v205/nair23a.html}

\leavevmode\vadjust pre{\hypertarget{ref-nakamoto2023calql}{}}%
Nakamoto, M., Zhai, Y., Singh, A., Mark, M. S., Ma, Y., Finn, C., Kumar, A., \& Levine, S. (2023). {Cal-QL}: Calibrated offline {RL} pre-training for efficient online fine-tuning. \emph{Advances in Neural Information Processing Systems (NeurIPS)}. \url{https://arxiv.org/abs/2303.05479v4}

\leavevmode\vadjust pre{\hypertarget{ref-nichol2018first}{}}%
Nichol, A., Achiam, J., \& Schulman, J. (2018). On first-order meta-learning algorithms. \emph{arXiv Preprint arXiv:1803.02999}. \url{https://arxiv.org/abs/1803.02999}

\leavevmode\vadjust pre{\hypertarget{ref-agarwal2025cosmos}{}}%
NVIDIA, Agarwal, N., Ali, A., Bala, M., Balaji, Y., Barker, E., Cai, T., Chattopadhyay, P., Chen, Y., Cui, Y., Ding, Y., Dworakowski, D., Fan, J., Fenzi, M., Ferroni, F., Fidler, S., Fox, D., Ge, S., Ge, Y., \ldots{} Zolkowski, A. (2025). Cosmos world foundation model platform for physical {AI}. \emph{arXiv Preprint arXiv:2501.03575}. \url{https://arxiv.org/abs/2501.03575}

\leavevmode\vadjust pre{\hypertarget{ref-bjorck2025groot}{}}%
NVIDIA, Bjorck, J., Castañeda, F., Cherniadev, N., Da, X., Ding, R., Fan, L. "Jim"., Fang, Y., Fox, D., Hu, F., Huang, S., Jang, J., Jiang, Z., Kautz, J., Kundalia, K., Lao, L., Li, Z., Lin, Z., Lin, K., \ldots{} Zhu, Y. (2025). {GR00T N1}: An open foundation model for generalist humanoid robots. \emph{arXiv Preprint arXiv:2503.14734}. \url{https://arxiv.org/abs/2503.14734}

\leavevmode\vadjust pre{\hypertarget{ref-ghosh2024octo}{}}%
Octo Model Team, Ghosh, D., Walke, H., Pertsch, K., Black, K., Mees, O., Dasari, S., Hejna, J., Kreiman, T., Xu, C., Luo, J., Tan, Y. L., Chen, L. Y., Sanketi, P., Vuong, Q., Xiao, T., Sadigh, D., Finn, C., \& Levine, S. (2024). Octo: An open-source generalist robot policy. \emph{Robotics: Science and Systems (RSS)}. \url{https://arxiv.org/abs/2405.12213v2}

\leavevmode\vadjust pre{\hypertarget{ref-oneill2023open}{}}%
Open X-Embodiment Collaboration, O'Neill, A., Rehman, A., Gupta, A., Maddukuri, A., Gupta, A., Padalkar, A., Lee, A., Pooley, A., Gupta, A., Mandlekar, A., Jain, A., Tung, A., Bewley, A., Herzog, A., Irpan, A., Khazatsky, A., Rai, A., Gupta, A., \ldots{} Lin, Z. (2024). Open {X}-{E}mbodiment: Robotic learning datasets and {RT-X} models. \emph{IEEE International Conference on Robotics and Automation (ICRA)}. \url{https://arxiv.org/abs/2310.08864v9}

\leavevmode\vadjust pre{\hypertarget{ref-akkaya2019solving}{}}%
OpenAI, Akkaya, I., Andrychowicz, M., Chociej, M., Litwin, M., McGrew, B., Petron, A., Paino, A., Plappert, M., Powell, G., Ribas, R., Schneider, J., Tezak, N., Tworek, J., Welinder, P., Weng, L., Yuan, Q., Zaremba, W., \& Zhang, L. (2019). Solving {R}ubik's cube with a robot hand. \emph{arXiv Preprint arXiv:1910.07113}. \url{https://arxiv.org/abs/1910.07113v1}

\leavevmode\vadjust pre{\hypertarget{ref-pan2010survey}{}}%
Pan, S. J., \& Yang, Q. (2010). A survey on transfer learning. \emph{IEEE Transactions on Knowledge and Data Engineering}, \emph{22}(10), 1345--1359. \url{https://doi.org/10.1109/tkde.2009.191}

\leavevmode\vadjust pre{\hypertarget{ref-parisi2022unsurprising}{}}%
Parisi, S., Rajeswaran, A., Purushwalkam, S., \& Gupta, A. (2022). The unsurprising effectiveness of pre-trained vision models for control. \emph{International Conference on Machine Learning (ICML)}. \url{https://arxiv.org/abs/2203.03580v2}

\leavevmode\vadjust pre{\hypertarget{ref-peng2018sim}{}}%
Peng, X. B., Andrychowicz, M., Zaremba, W., \& Abbeel, P. (2018). Sim-to-real transfer of robotic control with dynamics randomization. \emph{IEEE International Conference on Robotics and Automation (ICRA)}. \url{https://doi.org/10.1109/icra.2018.8460528}

\leavevmode\vadjust pre{\hypertarget{ref-peng2020learning}{}}%
Peng, X. B., Coumans, E., Zhang, T., Lee, T.-W., Tan, J., \& Levine, S. (2020). Learning agile robotic locomotion skills by imitating animals. \emph{Robotics: Science and Systems (RSS)}. \url{https://arxiv.org/abs/2004.00784v3}

\leavevmode\vadjust pre{\hypertarget{ref-pertsch2020accelerating}{}}%
Pertsch, K., Lee, Y., \& Lim, J. J. (2020). Accelerating reinforcement learning with learned skill priors. \emph{Conference on Robot Learning (CoRL)}. \url{https://arxiv.org/abs/2010.11944v1}

\leavevmode\vadjust pre{\hypertarget{ref-pertsch2025fast}{}}%
Pertsch, K., Stachowicz, K., Ichter, B., Driess, D., Nair, S., Vuong, Q., Mees, O., Finn, C., \& Levine, S. (2025). {FAST}: Efficient action tokenization for vision-language-action models. \emph{Robotics: Science and Systems (RSS)}. \url{https://doi.org/10.15607/rss.2025.xxi.012}

\leavevmode\vadjust pre{\hypertarget{ref-ai2026pi07}{}}%
Physical Intelligence, Ai, B., Amin, A., Aniceto, R., Balakrishna, A., Balke, G., Black, K., Bokinsky, G., Cao, S., Charbonnier, T., Choudhary, V., Collins, F., Conley, K., Connors, G., Darpinian, J., Dhabalia, K., Dhaka, M., DiCarlo, J., Driess, D., \ldots{} Zhilinsky, U. (2026). \(\pi_{0.7}\): A steerable generalist robotic foundation model with emergent capabilities. \emph{arXiv Preprint arXiv:2604.15483}. \url{https://arxiv.org/abs/2604.15483}

\leavevmode\vadjust pre{\hypertarget{ref-amin2025pistar}{}}%
Physical Intelligence, Amin, A., Aniceto, R., Balakrishna, A., Black, K., Conley, K., Connors, G., Darpinian, J., Dhabalia, K., DiCarlo, J., Driess, D., Equi, M., Esmail, A., Fang, Y., Finn, C., Glossop, C., Godden, T., Goryachev, I., Groom, L., \ldots{} Zhou, Z. (2025). \(\pi^{*}_{0.6}\): A {VLA} that learns from experience. \emph{arXiv Preprint arXiv:2511.14759}. \url{https://arxiv.org/abs/2511.14759}

\leavevmode\vadjust pre{\hypertarget{ref-black2025pi05}{}}%
Physical Intelligence, Black, K., Brown, N., Darpinian, J., Dhabalia, K., Driess, D., Esmail, A., Equi, M., Finn, C., Fusai, N., Galliker, M. Y., Ghosh, D., Groom, L., Hausman, K., Ichter, B., Jakubczak, S., Jones, T., Ke, L., LeBlanc, D., \ldots{} Zhilinsky, U. (2025). \(\pi_{0.5}\): A vision-language-action model with open-world generalization. \emph{Conference on Robot Learning (CoRL)}. \url{https://arxiv.org/abs/2504.16054v1}

\leavevmode\vadjust pre{\hypertarget{ref-pinto2017robust}{}}%
Pinto, L., Davidson, J., Sukthankar, R., \& Gupta, A. (2017). Robust adversarial reinforcement learning. \emph{International Conference on Machine Learning (ICML)}. \url{https://arxiv.org/abs/1703.02702v1}

\leavevmode\vadjust pre{\hypertarget{ref-pumacay2024colosseum}{}}%
Pumacay, W., Singh, I., Duan, J., Krishna, R., Thomason, J., \& Fox, D. (2024). {THE COLOSSEUM}: A benchmark for evaluating generalization for robotic manipulation. \emph{Robotics: Science and Systems (RSS)}. \url{https://doi.org/10.15607/rss.2024.xx.133}

\leavevmode\vadjust pre{\hypertarget{ref-qi2022inhand}{}}%
Qi, H., Kumar, A., Calandra, R., Ma, Y., \& Malik, J. (2023). In-hand object rotation via rapid motor adaptation. \emph{Conference on Robot Learning (CoRL)}, 1722--1732. \url{https://proceedings.mlr.press/v205/qi23a.html}

\leavevmode\vadjust pre{\hypertarget{ref-qi2023general}{}}%
Qi, H., Yi, B., Suresh, S., Lambeta, M., Ma, Y., Calandra, R., \& Malik, J. (2023). General in-hand object rotation with vision and touch. \emph{Conference on Robot Learning (CoRL)}, 2549--2564. \url{https://proceedings.mlr.press/v229/qi23a.html}

\leavevmode\vadjust pre{\hypertarget{ref-qiu2025humanoid}{}}%
Qiu, R.-Z., Yang, S., Cheng, X., Chawla, C., Li, J., He, T., Yan, G., Yoon, D. J., Hoque, R., Paulsen, L., Yang, G., Zhang, J., Yi, S., Shi, G., \& Wang, X. (2025). Humanoid policy \(\sim\) human policy. \emph{Conference on Robot Learning (CoRL)}, 2888--2906. \url{https://arxiv.org/abs/2503.13441v3}

\leavevmode\vadjust pre{\hypertarget{ref-qu2025spatialvla}{}}%
Qu, D., Song, H., Chen, Q., Yao, Y., Ye, X., Ding, Y., Wang, Z., Gu, J., Zhao, B., Wang, D., \& Li, X. (2025). {SpatialVLA}: Exploring spatial representations for visual-language-action models. \emph{Robotics: Science and Systems (RSS)}. \url{https://arxiv.org/abs/2501.15830v5}

\leavevmode\vadjust pre{\hypertarget{ref-qureshi2025splatsim}{}}%
Qureshi, M. N., Garg, S., Yandun, F., Held, D., Kantor, G., \& Silwal, A. (2025). {SplatSim}: Zero-shot {Sim2Real} transfer of {RGB} manipulation policies using {G}aussian splatting. \emph{IEEE International Conference on Robotics and Automation (ICRA)}. \url{https://doi.org/10.1109/icra55743.2025.11128339}

\leavevmode\vadjust pre{\hypertarget{ref-radosavovic2022realworld}{}}%
Radosavovic, I., Xiao, T., James, S., Abbeel, P., Malik, J., \& Darrell, T. (2023). Real-world robot learning with masked visual pre-training. \emph{Conference on Robot Learning (CoRL)}, 416--426. \url{https://proceedings.mlr.press/v205/radosavovic23a.html}

\leavevmode\vadjust pre{\hypertarget{ref-radosavovic2024real}{}}%
Radosavovic, I., Xiao, T., Zhang, B., Darrell, T., Malik, J., \& Sreenath, K. (2024). Real-world humanoid locomotion with reinforcement learning. \emph{Science Robotics}, \emph{9}(89), eadi9579. \url{https://doi.org/10.1126/scirobotics.adi9579}

\leavevmode\vadjust pre{\hypertarget{ref-rajeswaran2017epopt}{}}%
Rajeswaran, A., Ghotra, S., Ravindran, B., \& Levine, S. (2017). {EPOpt}: Learning robust neural network policies using model ensembles. \emph{International Conference on Learning Representations (ICLR)}. \url{https://arxiv.org/abs/1610.01283v4}

\leavevmode\vadjust pre{\hypertarget{ref-rakelly2019efficient}{}}%
Rakelly, K., Zhou, A., Quillen, D., Finn, C., \& Levine, S. (2019). Efficient off-policy meta-reinforcement learning via probabilistic context variables. \emph{International Conference on Machine Learning (ICML)}, 5331--5340. \url{https://arxiv.org/abs/1903.08254v1}

\leavevmode\vadjust pre{\hypertarget{ref-reed2022generalist}{}}%
Reed, S., Zolna, K., Parisotto, E., Colmenarejo, S. G., Novikov, A., Barth-Maron, G., Gimenez, M., Sulsky, Y., Kay, J., Springenberg, J. T., Eccles, T., Bruce, J., Razavi, A., Edwards, A., Heess, N., Chen, Y., Hadsell, R., Vinyals, O., Bordbar, M., \& Freitas, N. de. (2022). A generalist agent. \emph{Transactions on Machine Learning Research}. \url{https://arxiv.org/abs/2205.06175v3}

\leavevmode\vadjust pre{\hypertarget{ref-ren2024dppo}{}}%
Ren, A. Z., Lidard, J., Ankile, L. L., Simeonov, A., Agrawal, P., Majumdar, A., Burchfiel, B., Dai, H., \& Simchowitz, M. (2025). Diffusion policy policy optimization. \emph{International Conference on Learning Representations (ICLR)}. \url{https://arxiv.org/abs/2409.00588v3}

\leavevmode\vadjust pre{\hypertarget{ref-ross2011reduction}{}}%
Ross, S., Gordon, G. J., \& Bagnell, D. (2011). A reduction of imitation learning and structured prediction to no-regret online learning. \emph{International Conference on Artificial Intelligence and Statistics (AISTATS)}, 627--635. \url{https://arxiv.org/abs/1011.0686v3}

\leavevmode\vadjust pre{\hypertarget{ref-rudin2021learning}{}}%
Rudin, N., Hoeller, D., Reist, P., \& Hutter, M. (2022). Learning to walk in minutes using massively parallel deep reinforcement learning. \emph{Conference on Robot Learning (CoRL)}, 91--100. \url{https://proceedings.mlr.press/v164/rudin22a.html}

\leavevmode\vadjust pre{\hypertarget{ref-rusu2016policy}{}}%
Rusu, A. A., Colmenarejo, S. G., Gulcehre, C., Desjardins, G., Kirkpatrick, J., Pascanu, R., Mnih, V., Kavukcuoglu, K., \& Hadsell, R. (2016). Policy distillation. \emph{International Conference on Learning Representations (ICLR)}. \url{https://arxiv.org/abs/1511.06295v2}

\leavevmode\vadjust pre{\hypertarget{ref-rusu2016progressive}{}}%
Rusu, A. A., Rabinowitz, N. C., Desjardins, G., Soyer, H., Kirkpatrick, J., Kavukcuoglu, K., Pascanu, R., \& Hadsell, R. (2016). Progressive neural networks. \emph{arXiv Preprint arXiv:1606.04671}. \url{https://arxiv.org/abs/1606.04671}

\leavevmode\vadjust pre{\hypertarget{ref-rusu2017sim}{}}%
Rusu, A. A., Večerík, M., Rothörl, T., Heess, N., Pascanu, R., \& Hadsell, R. (2017). Sim-to-real robot learning from pixels with progressive nets. \emph{Conference on Robot Learning (CoRL)}. \url{https://arxiv.org/abs/1610.04286v2}

\leavevmode\vadjust pre{\hypertarget{ref-sadeghi2017cad2rl}{}}%
Sadeghi, F., \& Levine, S. (2017). {CAD2RL}: Real single-image flight without a single real image. \emph{Robotics: Science and Systems (RSS)}. \url{https://arxiv.org/abs/1611.04201v4}

\leavevmode\vadjust pre{\hypertarget{ref-savva2019habitat}{}}%
Savva, M., Kadian, A., Maksymets, O., Zhao, Y., Wijmans, E., Jain, B., Straub, J., Liu, J., Koltun, V., Malik, J., Parikh, D., \& Batra, D. (2019). Habitat: A platform for embodied {AI} research. \emph{IEEE/CVF International Conference on Computer Vision (ICCV)}, 9338--9346. \url{https://doi.org/10.1109/iccv.2019.00943}

\leavevmode\vadjust pre{\hypertarget{ref-saxena2025what}{}}%
Saxena, V., Bronars, M., Arachchige, N. R., Wang, K., Shin, W. C., Nasiriany, S., Mandlekar, A., \& Xu, D. (2025). What matters in learning from large-scale datasets for robot manipulation. \emph{International Conference on Learning Representations (ICLR)}. \url{https://arxiv.org/abs/2506.13536v1}

\leavevmode\vadjust pre{\hypertarget{ref-schaff2023sim}{}}%
Schaff, C., Sedal, A., Ni, S., \& Walter, M. R. (2023). Sim-to-real transfer of co-optimized soft robot crawlers. \emph{Autonomous Robots}, \emph{47}(8), 1195--1211. \url{https://doi.org/10.1007/s10514-023-10130-8}

\leavevmode\vadjust pre{\hypertarget{ref-schneider2024surprising}{}}%
Schneider, M., Krug, R., Vaskevicius, N., Palmieri, L., \& Boedecker, J. (2024). The surprising ineffectiveness of pre-trained visual representations for model-based reinforcement learning. \emph{Advances in Neural Information Processing Systems (NeurIPS)}. \url{https://arxiv.org/abs/2411.10175v2}

\leavevmode\vadjust pre{\hypertarget{ref-seita2020fabric}{}}%
Seita, D., Ganapathi, A., Hoque, R., Hwang, M., Cen, E., Tanwani, A. K., Balakrishna, A., Thananjeyan, B., Ichnowski, J., Jamali, N., Yamane, K., Iba, S., Canny, J., \& Goldberg, K. (2020). Deep imitation learning of sequential fabric smoothing from an algorithmic supervisor. \emph{IEEE/RSJ International Conference on Intelligent Robots and Systems (IROS)}. \url{https://doi.org/10.1109/iros45743.2020.9341608}

\leavevmode\vadjust pre{\hypertarget{ref-shah2022lmnav}{}}%
Shah, D., Osiński, B., Ichter, B., \& Levine, S. (2023). {LM-Nav}: Robotic navigation with large pre-trained models of language, vision, and action. \emph{Conference on Robot Learning (CoRL)}, 492--504. \url{https://proceedings.mlr.press/v205/shah23b.html}

\leavevmode\vadjust pre{\hypertarget{ref-shah2023gnm}{}}%
Shah, D., Sridhar, A., Bhorkar, A., Hirose, N., \& Levine, S. (2023). {GNM}: A general navigation model to drive any robot. \emph{IEEE International Conference on Robotics and Automation (ICRA)}, 7226--7233. \url{https://doi.org/10.1109/icra48891.2023.10161227}

\leavevmode\vadjust pre{\hypertarget{ref-shah2023vint}{}}%
Shah, D., Sridhar, A., Dashora, N., Stachowicz, K., Black, K., Hirose, N., \& Levine, S. (2023). {ViNT}: A foundation model for visual navigation. \emph{Conference on Robot Learning (CoRL)}, 711--733. \url{https://proceedings.mlr.press/v229/shah23a.html}

\leavevmode\vadjust pre{\hypertarget{ref-shang2024theia}{}}%
Shang, J., Schmeckpeper, K., May, B. B., Minniti, M. V., Kelestemur, T., Watkins, D., \& Herlant, L. (2025). Theia: Distilling diverse vision foundation models for robot learning. \emph{Conference on Robot Learning (CoRL)}, 724--748. \url{https://proceedings.mlr.press/v270/shang25a.html}

\leavevmode\vadjust pre{\hypertarget{ref-shaw2022videodex}{}}%
Shaw, K., Bahl, S., \& Pathak, D. (2023). {VideoDex}: Learning dexterity from internet videos. \emph{Conference on Robot Learning (CoRL)}, 654--665. \url{https://proceedings.mlr.press/v205/shaw23a.html}

\leavevmode\vadjust pre{\hypertarget{ref-shen2023distilled}{}}%
Shen, W., Yang, G., Yu, A., Wong, J., Kaelbling, L. P., \& Isola, P. (2023). Distilled feature fields enable few-shot language-guided manipulation. \emph{Conference on Robot Learning (CoRL)}, 405--424. \url{https://proceedings.mlr.press/v229/shen23a.html}

\leavevmode\vadjust pre{\hypertarget{ref-shridhar2021cliport}{}}%
Shridhar, M., Manuelli, L., \& Fox, D. (2022). {CLIP}ort: What and where pathways for robotic manipulation. \emph{Conference on Robot Learning (CoRL)}, 894--906. \url{https://proceedings.mlr.press/v164/shridhar22a.html}

\leavevmode\vadjust pre{\hypertarget{ref-shridhar2022peract}{}}%
Shridhar, M., Manuelli, L., \& Fox, D. (2023). Perceiver-actor: A multi-task transformer for robotic manipulation. \emph{Conference on Robot Learning (CoRL)}, 785--799. \url{https://proceedings.mlr.press/v205/shridhar23a.html}

\leavevmode\vadjust pre{\hypertarget{ref-shukor2025smolvla}{}}%
Shukor, M., Aubakirova, D., Capuano, F., Kooijmans, P., Palma, S., Zouitine, A., Aractingi, M., Pascal, C., Russi, M., Marafioti, A., Alibert, S., Cord, M., Wolf, T., \& Cadene, R. (2025). {SmolVLA}: A vision-language-action model for affordable and efficient robotics. \emph{arXiv Preprint arXiv:2506.01844}. \url{https://arxiv.org/abs/2506.01844}

\leavevmode\vadjust pre{\hypertarget{ref-song2024learning}{}}%
Song, Y., Kim, S. bae, \& Scaramuzza, D. (2025). Learning quadruped locomotion using differentiable simulation. \emph{Conference on Robot Learning (CoRL)}, 258--271. \url{https://proceedings.mlr.press/v270/song25a.html}

\leavevmode\vadjust pre{\hypertarget{ref-song2023reaching}{}}%
Song, Y., Romero, A., Müller, M., Koltun, V., \& Scaramuzza, D. (2023). Reaching the limit in autonomous racing: Optimal control versus reinforcement learning. \emph{Science Robotics}, \emph{8}(82), eadg1462. \url{https://doi.org/10.1126/scirobotics.adg1462}

\leavevmode\vadjust pre{\hypertarget{ref-sridhar2024nomad}{}}%
Sridhar, A., Shah, D., Glossop, C., \& Levine, S. (2024). {NoMaD}: Goal masked diffusion policies for navigation and exploration. \emph{IEEE International Conference on Robotics and Automation (ICRA)}, 63--70. \url{https://doi.org/10.1109/icra57147.2024.10610665}

\leavevmode\vadjust pre{\hypertarget{ref-sutton1999between}{}}%
Sutton, R. S., Precup, D., \& Singh, S. (1999). Between {MDP}s and semi-{MDP}s: A framework for temporal abstraction in reinforcement learning. \emph{Artificial Intelligence}, \emph{112}(1-2), 181--211. \url{https://doi.org/10.1016/s0004-3702(99)00052-1}

\leavevmode\vadjust pre{\hypertarget{ref-tan2018sim}{}}%
Tan, J., Zhang, T., Coumans, E., Iscen, A., Bai, Y., Hafner, D., Bohez, S., \& Vanhoucke, V. (2018). Sim-to-real: Learning agile locomotion for quadruped robots. \emph{Robotics: Science and Systems (RSS)}. \url{https://doi.org/10.15607/rss.2018.xiv.010}

\leavevmode\vadjust pre{\hypertarget{ref-tan2025ript}{}}%
Tan, S., Dou, K., Zhao, Y., \& Krähenbühl, P. (2025). Interactive post-training for vision-language-action models. \emph{arXiv Preprint arXiv:2505.17016}. \url{https://arxiv.org/abs/2505.17016}

\leavevmode\vadjust pre{\hypertarget{ref-tao2024maniskill3}{}}%
Tao, S., Xiang, F., Shukla, A., Qin, Y., Hinrichsen, X., Yuan, X., Bao, C., Lin, X., Liu, Y., Chan, T.-K., Gao, Y., Li, X., Mu, T., Xiao, N., Gurha, A., Rajesh, V., Choi, Y., Chen, Y.-R., Huang, Z., \ldots{} Su, H. (2025). Demonstrating {GPU} parallelized robot simulation and rendering for generalizable embodied {AI} with {ManiSkill3}. \emph{Robotics: Science and Systems (RSS)}. \url{https://doi.org/10.15607/rss.2025.xxi.021}

\leavevmode\vadjust pre{\hypertarget{ref-taylor2009transfer}{}}%
Taylor, M. E., \& Stone, P. (2009). Transfer learning for reinforcement learning domains: A survey. \emph{Journal of Machine Learning Research}, \emph{10}, 1633--1685. \url{https://www.jmlr.org/papers/v10/taylor09a.html}

\leavevmode\vadjust pre{\hypertarget{ref-thrun1995lifelong}{}}%
Thrun, S., \& Mitchell, T. M. (1995). Lifelong robot learning. \emph{Robotics and Autonomous Systems}, \emph{15}(1--2), 25--46. \url{https://doi.org/10.1016/0921-8890(95)00004-y}

\leavevmode\vadjust pre{\hypertarget{ref-thuruthel2019model}{}}%
Thuruthel, T. G., Falotico, E., Renda, F., \& Laschi, C. (2019). Model-based reinforcement learning for closed-loop dynamic control of soft robotic manipulators. \emph{IEEE Transactions on Robotics}, \emph{35}(1), 124--134. \url{https://doi.org/10.1109/tro.2018.2878318}

\leavevmode\vadjust pre{\hypertarget{ref-thuruthel2019soft}{}}%
Thuruthel, T. G., Shih, B., Laschi, C., \& Tolley, M. T. (2019). Soft robot perception using embedded soft sensors and recurrent neural networks. \emph{Science Robotics}, \emph{4}(26), eaav1488. \url{https://doi.org/10.1126/scirobotics.aav1488}

\leavevmode\vadjust pre{\hypertarget{ref-tirinzoni2025zero}{}}%
Tirinzoni, A., Touati, A., Farebrother, J., Guzek, M., Kanervisto, A., Xu, Y., Lazaric, A., \& Pirotta, M. (2025). Zero-shot whole-body humanoid control via behavioral foundation models. \emph{International Conference on Learning Representations (ICLR)}. \url{https://arxiv.org/abs/2504.11054v1}

\leavevmode\vadjust pre{\hypertarget{ref-tobin2017domain}{}}%
Tobin, J., Fong, R., Ray, A., Schneider, J., Zaremba, W., \& Abbeel, P. (2017). Domain randomization for transferring deep neural networks from simulation to the real world. \emph{IEEE/RSJ International Conference on Intelligent Robots and Systems (IROS)}. \url{https://doi.org/10.1109/iros.2017.8202133}

\leavevmode\vadjust pre{\hypertarget{ref-torabi2018behavioral}{}}%
Torabi, F., Warnell, G., \& Stone, P. (2018). Behavioral cloning from observation. \emph{International Joint Conference on Artificial Intelligence (IJCAI)}. \url{https://arxiv.org/abs/1805.01954v2}

\leavevmode\vadjust pre{\hypertarget{ref-torne2024reconciling}{}}%
Torne, M., Simeonov, A., Li, Z., Chan, A., Chen, T., Gupta, A., \& Agrawal, P. (2024). Reconciling reality through simulation: A real-to-sim-to-real approach for robust manipulation. \emph{Robotics: Science and Systems (RSS)}. \url{https://arxiv.org/abs/2403.03949v3}

\leavevmode\vadjust pre{\hypertarget{ref-touati2021learning}{}}%
Touati, A., \& Ollivier, Y. (2021). Learning one representation to optimize all rewards. \emph{Advances in Neural Information Processing Systems (NeurIPS)}. \url{https://arxiv.org/abs/2103.07945v3}

\leavevmode\vadjust pre{\hypertarget{ref-touati2023does}{}}%
Touati, A., Rapin, J., \& Ollivier, Y. (2023). Does zero-shot reinforcement learning exist? \emph{International Conference on Learning Representations (ICLR)}. \url{https://arxiv.org/abs/2209.14935v2}

\leavevmode\vadjust pre{\hypertarget{ref-vincent2024how}{}}%
Vincent, J. A., Nishimura, H., Itkina, M., Shah, P., Schwager, M., \& Kollar, T. (2024). How generalizable is my behavior cloning policy? A statistical approach to trustworthy performance evaluation. \emph{IEEE Robotics and Automation Letters}, \emph{9}(10), 8619--8626. \url{https://doi.org/10.1109/lra.2024.3445635}

\leavevmode\vadjust pre{\hypertarget{ref-walke2023bridgedata}{}}%
Walke, H. R., Black, K., Zhao, T. Z., Vuong, Q., Zheng, C., Hansen-Estruch, P., He, A. W., Myers, V., Kim, M. J., Du, M., Lee, A., Fang, K., Finn, C., \& Levine, S. (2023). {BridgeData V2}: A dataset for robot learning at scale. \emph{Conference on Robot Learning (CoRL)}, 1723--1736. \url{https://proceedings.mlr.press/v229/walke23a.html}

\leavevmode\vadjust pre{\hypertarget{ref-wang2023mimicplay}{}}%
Wang, C., Fan, L., Sun, J., Zhang, R., Fei-Fei, L., Xu, D., Zhu, Y., \& Anandkumar, A. (2023). {MimicPlay}: Long-horizon imitation learning by watching human play. \emph{Conference on Robot Learning (CoRL)}, 201--221. \url{https://proceedings.mlr.press/v229/wang23a.html}

\leavevmode\vadjust pre{\hypertarget{ref-wang2024dexcap}{}}%
Wang, C., Shi, H., Wang, W., Zhang, R., Fei-Fei, L., \& Liu, C. K. (2024). {DexCap}: Scalable and portable mocap data collection system for dexterous manipulation. \emph{Robotics: Science and Systems (RSS)}. \url{https://arxiv.org/abs/2403.07788v2}

\leavevmode\vadjust pre{\hypertarget{ref-wang2024scaling}{}}%
Wang, L., Chen, X., Zhao, J., \& He, K. (2024). Scaling proprioceptive-visual learning with heterogeneous pre-trained transformers. \emph{Advances in Neural Information Processing Systems (NeurIPS)}. \url{https://arxiv.org/abs/2409.20537v1}

\leavevmode\vadjust pre{\hypertarget{ref-wang2022tacto}{}}%
Wang, S., Lambeta, M., Chou, P.-W., \& Calandra, R. (2022). {TACTO}: A fast, flexible, and open-source simulator for high-resolution vision-based tactile sensors. \emph{IEEE Robotics and Automation Letters}, \emph{7}(2), 3930--3937. \url{https://doi.org/10.1109/lra.2022.3146945}

\leavevmode\vadjust pre{\hypertarget{ref-wang2018nervenet}{}}%
Wang, T., Liao, R., Ba, J., \& Fidler, S. (2018). {NerveNet}: Learning structured policy with graph neural networks. \emph{International Conference on Learning Representations (ICLR)}. \url{https://iclr.cc/virtual/2018/poster/66}

\leavevmode\vadjust pre{\hypertarget{ref-wang2019characterizing}{}}%
Wang, Z., Dai, Z., Póczos, B., \& Carbonell, J. (2019). Characterizing and avoiding negative transfer. \emph{IEEE/CVF Conference on Computer Vision and Pattern Recognition (CVPR)}, 11285--11294. \url{https://arxiv.org/abs/1811.09751v4}

\leavevmode\vadjust pre{\hypertarget{ref-wu2024unleashing}{}}%
Wu, H., Jing, Y., Cheang, C., Chen, G., Xu, J., Li, X., Liu, M., Li, H., \& Kong, T. (2024). Unleashing large-scale video generative pre-training for visual robot manipulation. \emph{International Conference on Learning Representations (ICLR)}. \url{https://arxiv.org/abs/2312.13139v2}

\leavevmode\vadjust pre{\hypertarget{ref-wu2024robomind}{}}%
Wu, K., Hou, C., Liu, J., Che, Z., Ju, X., Yang, Z., Li, M., Zhao, Y., Xu, Z., Yang, G., Fan, S., Wang, X., Liao, F., Zhao, Z., Li, G., Jin, Z., Wang, L., Mao, J., Liu, N., \ldots{} Tang, J. (2025). {RoboMIND}: Benchmark on multi-embodiment intelligence normative data for robot manipulation. \emph{Robotics: Science and Systems (RSS)}. \url{https://doi.org/10.15607/rss.2025.xxi.152}

\leavevmode\vadjust pre{\hypertarget{ref-xing2025shortcut}{}}%
Xing, Y., Luo, X., Xie, J., Gao, L., Shen, H. T., \& Song, J. (2025). Shortcut learning in generalist robot policies: The role of dataset diversity and fragmentation. \emph{Conference on Robot Learning (CoRL)}. \url{https://arxiv.org/abs/2508.06426v1}

\leavevmode\vadjust pre{\hypertarget{ref-yang2026data}{}}%
Yang, J., Finn, C., \& Sadigh, D. (2026). Data analogies enable efficient cross-embodiment transfer. \emph{arXiv Preprint arXiv:2603.06450}. \url{https://arxiv.org/abs/2603.06450}

\leavevmode\vadjust pre{\hypertarget{ref-yang2024pushing}{}}%
Yang, J., Glossop, C., Bhorkar, A., Shah, D., Vuong, Q., Finn, C., Sadigh, D., \& Levine, S. (2024). Pushing the limits of cross-embodiment learning for manipulation and navigation. \emph{Robotics: Science and Systems (RSS)}. \url{https://arxiv.org/abs/2402.19432v1}

\leavevmode\vadjust pre{\hypertarget{ref-yang2023unisim}{}}%
Yang, S., Du, Y., Ghasemipour, S. K. S., Tompson, J., Kaelbling, L., Schuurmans, D., \& Abbeel, P. (2024). Learning interactive real-world simulators. \emph{International Conference on Learning Representations (ICLR)}. \url{https://arxiv.org/abs/2310.06114v3}

\leavevmode\vadjust pre{\hypertarget{ref-ye2024latent}{}}%
Ye, S., Jang, J., Jeon, B., Joo, S., Yang, J., Peng, B., Mandlekar, A., Tan, R., Chao, Y.-W., Lin, B. Y., Liden, L., Lee, K., Gao, J., Zettlemoyer, L., Fox, D., \& Seo, M. (2025). Latent action pretraining from videos. \emph{International Conference on Learning Representations (ICLR)}. \url{https://arxiv.org/abs/2410.11758v2}

\leavevmode\vadjust pre{\hypertarget{ref-yu2024mimictouch}{}}%
Yu, K., Han, Y., Wang, Q., Saxena, V., Xu, D., \& Zhao, Y. (2025). {MimicTouch}: Leveraging multi-modal human tactile demonstrations for contact-rich manipulation. \emph{Conference on Robot Learning (CoRL)}, 4844--4865. \url{https://proceedings.mlr.press/v270/yu25c.html}

\leavevmode\vadjust pre{\hypertarget{ref-yu2018oneshot}{}}%
Yu, T., Finn, C., Dasari, S., Xie, A., Zhang, T., Abbeel, P., \& Levine, S. (2018). One-shot imitation from observing humans via domain-adaptive meta-learning. \emph{Robotics: Science and Systems (RSS)}. \url{https://doi.org/10.15607/rss.2018.xiv.002}

\leavevmode\vadjust pre{\hypertarget{ref-yu2019metaworld}{}}%
Yu, T., Quillen, D., He, Z., Julian, R., Hausman, K., Finn, C., \& Levine, S. (2019). Meta-world: A benchmark and evaluation for multi-task and meta reinforcement learning. \emph{Conference on Robot Learning (CoRL)}. \url{https://arxiv.org/abs/1910.10897v2}

\leavevmode\vadjust pre{\hypertarget{ref-yuan2025policy}{}}%
Yuan, X., Mu, T., Tao, S., Fang, Y., Zhang, M., \& Su, H. (2025). Policy decorator: Model-agnostic online refinement for large policy model. \emph{International Conference on Learning Representations (ICLR)}. \url{https://arxiv.org/abs/2412.13630v1}

\leavevmode\vadjust pre{\hypertarget{ref-zakka2021xirl}{}}%
Zakka, K., Zeng, A., Florence, P., Tompson, J., Bohg, J., \& Dwibedi, D. (2022). {XIRL}: Cross-embodiment inverse reinforcement learning. \emph{Conference on Robot Learning (CoRL)}, 537--546. \url{https://proceedings.mlr.press/v164/zakka22a.html}

\leavevmode\vadjust pre{\hypertarget{ref-zawalski2024ecot}{}}%
Zawalski, M., Chen, W., Pertsch, K., Mees, O., Finn, C., \& Levine, S. (2025). Robotic control via embodied chain-of-thought reasoning. \emph{Conference on Robot Learning (CoRL)}, 3157--3181. \url{https://proceedings.mlr.press/v270/zawalski25a.html}

\leavevmode\vadjust pre{\hypertarget{ref-ze2024dp3}{}}%
Ze, Y., Zhang, G., Zhang, K., Hu, C., Wang, M., \& Xu, H. (2024). {3D} diffusion policy: Generalizable visuomotor policy learning via simple {3D} representations. \emph{Robotics: Science and Systems (RSS)}. \url{https://doi.org/10.15607/rss.2024.xx.067}

\leavevmode\vadjust pre{\hypertarget{ref-zhang2023survey}{}}%
Zhang, W., Deng, L., Zhang, L., \& Wu, D. (2023). A survey on negative transfer. \emph{IEEE/CAA Journal of Automatica Sinica}, \emph{10}(2), 305--329. \url{https://doi.org/10.1109/jas.2022.106004}

\leavevmode\vadjust pre{\hypertarget{ref-zhao2023learning}{}}%
Zhao, T. Z., Kumar, V., Levine, S., \& Finn, C. (2023). Learning fine-grained bimanual manipulation with low-cost hardware. \emph{Robotics: Science and Systems (RSS)}. \url{https://arxiv.org/abs/2304.13705v1}

\leavevmode\vadjust pre{\hypertarget{ref-zhen20243dvla}{}}%
Zhen, H., Qiu, X., Chen, P., Yang, J., Yan, X., Du, Y., Hong, Y., \& Gan, C. (2024). {3D-VLA}: A 3D vision-language-action generative world model. \emph{International Conference on Machine Learning (ICML)}. \url{https://arxiv.org/abs/2403.09631v1}

\leavevmode\vadjust pre{\hypertarget{ref-zheng2025universal}{}}%
Zheng, J., Li, J., Liu, D., Zheng, Y., Wang, Z., Ou, Z., Liu, Y., Liu, J., Zhang, Y.-Q., \& Zhan, X. (2025). Universal actions for enhanced embodied foundation models. \emph{IEEE/CVF Conference on Computer Vision and Pattern Recognition (CVPR)}. \url{https://arxiv.org/abs/2501.10105v2}

\leavevmode\vadjust pre{\hypertarget{ref-zhou2025dinowm}{}}%
Zhou, G., Pan, H., LeCun, Y., \& Pinto, L. (2025). {DINO-WM}: World models on pre-trained visual features enable zero-shot planning. \emph{International Conference on Machine Learning (ICML)}. \url{https://arxiv.org/abs/2411.04983v2}

\leavevmode\vadjust pre{\hypertarget{ref-zhou2025autoeval}{}}%
Zhou, Z., Atreya, P., Tan, Y. L., Pertsch, K., \& Levine, S. (2025). {AutoEval}: Autonomous evaluation of generalist robot manipulation policies in the real world. \emph{Conference on Robot Learning (CoRL)}. \url{https://arxiv.org/abs/2503.24278v2}

\leavevmode\vadjust pre{\hypertarget{ref-zhou2025chatvla}{}}%
Zhou, Z., Zhu, Y., Zhu, M., Wen, J., Liu, N., Xu, Z., Meng, W., Peng, Y., Shen, C., Feng, F., \& Xu, Y. (2025). {ChatVLA}: Unified multimodal understanding and robot control with vision-language-action model. \emph{Conference on Empirical Methods in Natural Language Processing (EMNLP)}, 5377--5395. \url{https://doi.org/10.18653/v1/2025.emnlp-main.273}

\leavevmode\vadjust pre{\hypertarget{ref-zhuang2023robot}{}}%
Zhuang, Z., Fu, Z., Wang, J., Atkeson, C. G., Schwertfeger, S., Finn, C., \& Zhao, H. (2023). Robot parkour learning. \emph{Conference on Robot Learning (CoRL)}, 73--92. \url{https://proceedings.mlr.press/v229/zhuang23a.html}

\leavevmode\vadjust pre{\hypertarget{ref-brohan2023rt2}{}}%
Zitkovich, B., Yu, T., Xu, S., Xu, P., Xiao, T., Xia, F., Wu, J., Wohlhart, P., Welker, S., Wahid, A., Vuong, Q., Vanhoucke, V., Tran, H., Soricut, R., Singh, A., Singh, J., Sermanet, P., Sanketi, P. R., Salazar, G., \ldots{} Han, K. (2023). {RT-2}: Vision-language-action models transfer web knowledge to robotic control. \emph{Conference on Robot Learning (CoRL)}, 2165--2183. \url{https://proceedings.mlr.press/v229/zitkovich23a.html}

\end{CSLReferences}

\end{document}


\maketitle
\noindent\textbf{Article:} What Transfers in Robot Learning? Taxonomy and Evidence from Domain Randomization to VLA Models

\medskip
\noindent This supplement documents collection and coding and supplies the complete per-paper table. The five transfer categories, synthesis, practical guidance, and open challenges are in the main article. Author--year labels refer to the main article's bibliography; Supplementary Data S2 supplies the complete catalogue and machine-readable records.
\section{Collection Protocol and Scope}
\label{app:method}

\paragraph{Sources and period.} The corpus spans 1995--November 2026. The main search closed in July 2026; a targeted update on 26 November 2026 added five relevant works identified through a targeted publication-record search. Sources were arXiv, Semantic Scholar, Google Scholar, major robotics and machine-learning proceedings (RSS, CoRL, ICRA, IROS, NeurIPS, ICML, ICLR, CVPR), and relevant journals including IJRR, IEEE T-RO, RA-L, Science Robotics, JMLR, and TPAMI. The November update is targeted rather than an exhaustive extension of every search family.

\paragraph{Keyword families.} Searches covered eight families: transfer foundations and surveys; sim-to-real, randomization, and identification; adaptation and pretrained representations; task transfer, meta-learning, skills, and offline-to-online learning; imitation and human transfer; cross-embodiment methods and datasets; foundation and vision-language-action models; and benchmarks and evaluation. A robotics conjunction (\texttt{robot}, \texttt{manipulation}, \texttt{locomotion}, or \texttt{embodied}) was combined with disjunctions of family terms over titles and abstracts. For example, the sim-to-real terms were \texttt{sim-to-real}, \texttt{sim2real}, \texttt{domain randomization}, \texttt{system identification}, \texttt{real-to-sim}, and \texttt{reality gap}; embodiment terms included \texttt{cross-embodiment}, \texttt{multi-embodiment}, \texttt{morphology}, \texttt{action space}, and \texttt{retargeting}. Seminal-work searches were combined with recent-work searches for 2024--2026 and citation snowballing.

\paragraph{Inclusion and stopping.} Included works introduce a mechanism relevant to robot control or perception-for-control, test such a mechanism, or supply an evaluation instrument used by the survey. Peer-reviewed versions were preferred and each work counted once; primary preprints and technical reports were retained for recent systems. Searches covered English-language sources. General computer-vision adaptation and language-only planning without a physical-action interface were excluded. Search within a family stopped when a query round and snowballing round produced no previously unrepresented mechanism; later additions required stronger evidence, an evaluation instrument, or direct relevance to a stated claim. This mechanism-saturation rule favors breadth of mechanisms rather than exhaustive publication counting.

\paragraph{Selection limits.} Database hit counts and a complete search-stage flow were not retained. The updated records contain 232 corpus entries and 185 coded rows. Screening covered 218 works: 185 are represented in the coded subset and 33 are not. The five additions in November are outside the coded subset. A further 14 corpus works lie outside the original screening. The catalogue identifies all 47 works outside the coded set but does not reconstruct missing original screening decisions. The corpus is concentrated in manipulation, locomotion, and recent foundation-model work. Neither marginal nor conditional rates are estimates from a probability sample of the field.

\section{Evidence Coding and Reproduction}
\label{app:coding}

\paragraph{Unit and assignment.} Each row grades one focal empirical transfer claim, not the paper's overall quality. The gap is the source-target difference the mechanism principally addresses; the form is its principally transferred artifact or interface. Where a system spans several gaps, the focal claim breaks the tie. This assigns each of 185 papers once and avoids duplicate counting. Era I ends in 2016, era II spans 2017--2021, and era III spans 2022--July 2026.

\paragraph{Rubric and boundary rules.} The main article defines E0 (demonstration only), E1 (external comparison), E2 (self-run but confounded comparison), E3 (matched controlled comparison), and E4 (E3 plus counts and dispersion for that comparison). Grade the strongest evidence supporting the focal claim; a strong experiment elsewhere in the paper cannot upgrade it. If physical transfer is the claim, a simulation-only ablation does not supply its physical comparison. Mixed or negative outcomes do not lower a grade. Ambiguous target-budget reporting is assigned the lower level. Dispersion attached to a different contrast does not satisfy E4. The primary grades are retained from the complete coding table; the release does not add new paper-level grades.

\paragraph{Worked examples.} Helix (Figure AI, 2025) is E0 because the cited capability has no comparison condition. Rudin et al. (2021) is E1 because the training-time comparison uses externally reported systems with differing hardware or procedures. Zheng et al. (2025) is E2 because compared models differ substantially in scale. Peng et al. (2018) is E3 because the dynamics-randomization ablation fixes the task, policy class, and target protocol but does not report dispersion for the focal contrast. Andrychowicz et al. (2020) and Hansen et al. (2023) are E4 because controlled contrasts are accompanied by stated trials or seeds and dispersion. These labels refer to the specific contrasts recorded below.

\paragraph{Reproduction files.} \texttt{evidence\_coding.csv} transcribes all six columns of Table~\ref{app:coding-table}: citation key, gap, form, era, grade, and basis. It contains 185 unique keys. \texttt{corpus\_catalogue.csv} contains metadata for all 232 works and their membership in the coded set; \texttt{references.bib} retains the complete bibliography. \texttt{reproduce\_counts.py} recomputes the aggregate counts in \texttt{evidence\_summary.json} from the coding CSV. The five additions are recorded in \texttt{corpus\_additions.csv}; \texttt{corpus\_categories.csv} records the primary category of every work. Figure 2 is regenerated from catalogue years and categories, using documented first online dates when these precede the proceedings imprint year. These files are supplied in Supplementary Data S2. The script reproduces the descriptive counts without re-evaluating the source papers. Supplement and main article share one bibliography to avoid duplicating it here.

\paragraph{Interpretation.} Conservative grading can affect groups unequally when reporting ambiguities differ. The era comparison is therefore reported under the retained grades without claims about a field-wide trend or reconstructed sensitivity flags. Independent replication uses the main article's author-disjoint, matched-contrast rule and is not an E level. The absence recorded for this corpus is restricted to its coverage and search date.

\clearpage
\begingroup
\small
\setlength{\tabcolsep}{4pt}
\setlength{\LTcapwidth}{\linewidth}
\setlength{\LTleft}{0pt}
\setlength{\LTright}{0pt}
\begin{longtable}{@{}>{\raggedright\arraybackslash}p{3.25cm} c c c c >{\raggedright\arraybackslash}p{7.7cm}@{}}
\caption{Complete E0--E4 coding of the 185 empirical transfer results in the corpus, ordered by gap and then by descending evidence level. \textsc{Gap}: S2R = simulation$\rightarrow$reality, Task = task$\rightarrow$task, Embod = embodiment$\rightarrow$embodiment, Human = human$\rightarrow$robot, Web = web$\rightarrow$robot. \textsc{Form}: Data, Repr = representations, Param = parameters and policies, Model = models and simulators, Cond = task conditioning. \textsc{Era}: I $\leq$2016, II 2017--2021, III 2022--2026. The \textsc{Basis} column states, for each paper, the comparison on which the grade was assigned; grades are for the specific transfer claim this survey attributes to the paper, not for the paper as a whole.}
\label{app:coding-table}\\
\toprule
\textbf{Reference} & \textbf{Gap} & \textbf{Form} & \textbf{Era} & \textbf{E} & \textbf{Basis for the grade} \\
\midrule
\endfirsthead
\multicolumn{6}{@{}l}{\footnotesize\textit{Table~\thetable\ continued from previous page.}}\\
\toprule
\textbf{Reference} & \textbf{Gap} & \textbf{Form} & \textbf{Era} & \textbf{E} & \textbf{Basis for the grade} \\
\midrule
\endhead
\midrule \multicolumn{6}{r@{}}{\footnotesize\textit{continued on next page}}\\
\endfoot
\bottomrule
\endlastfoot
Andrychowicz et al. (2020) & S2R & Data & II & 4 & ablation table with success distributions over stated physical trials \\
Handa et al. (2023) & S2R & Data & III & 4 & consecutive-success distributions over stated trial counts with ablations \\
Kaufmann et al. (2023) & S2R & Model & III & 4 & component ablations of the full system reported with lap-time dispersion over a week of recorded runs \\
OpenAI et al. (2019) & S2R & Data & II & 3 & ADR against manual randomization; physical successes counted \\
Bousmalis et al. (2018) & S2R & Repr & II & 3 & real data required for a fixed performance level \\
Chaffre et al. (2025) & S2R & Cond & III & 3 & adaptive against non-adaptive model-based control on the same vehicle \\
Chebotar et al. (2019) & S2R & Model & II & 3 & adapted distribution against fixed randomization on the same real tasks \\
Church et al. (2022) & S2R & Repr & II & 3 & real-to-sim translation against direct simulation training on the same tasks \\
Fan et al. (2020) & S2R & Param & II & 3 & against classical and learned baselines with physical transfer \\
Ganapathi et al. (2021) & S2R & Repr & II & 3 & against analytic and prior baselines on the same physical fabric tasks \\
Ha and Song (2022) & S2R & Param & II & 3 & dynamic primitive against quasi-static baseline on the same real platform \\
T. He, Gao, et al. (2025) & S2R & Model & III & 3 & delta-dynamics against randomization and system-identification baselines \\
Hwangbo et al. (2019) & S2R & Model & II & 3 & learned actuator model against analytic and idealized simulators \\
Kumar et al. (2021) & S2R & Cond & II & 3 & against robust and system-identification baselines on the same terrains \\
Lee et al. (2020) & S2R & Param & II & 3 & privileged distillation against the baseline controller on the same terrains \\
Li et al. (2018) & S2R & Repr & II & 3 & corrected against uncorrected imagery on the same scenes \\
Loquercio et al. (2020) & S2R & Data & II & 3 & randomization ablations with real flight trials against baselines \\
Loquercio et al. (2021) & S2R & Param & II & 3 & against expert and non-distilled baselines over stated flight runs \\
Ma, Liang, Wang, Wang, et al. (2024) & S2R & Cond & III & 3 & against human-designed randomization pipelines on the same platforms \\
Muratore et al. (2021) & S2R & Model & II & 3 & against fixed and adaptive randomization baselines on the same platform \\
Peng et al. (2018) & S2R & Data & II & 3 & with/without dynamics randomization and memory on the same task \\
Pinto et al. (2017) & S2R & Data & II & 3 & internal baselines over seeds; simulated perturbation targets only \\
Qi, Kumar, et al. (2023) & S2R & Cond & III & 3 & adaptation-module ablations with a shared real protocol \\
Qi, Yi, et al. (2023) & S2R & Data & III & 3 & training-object diversity ablations with a shared real protocol \\
Qureshi et al. (2025) & S2R & Repr & III & 3 & 86.25\% against 97.5\% real-trained on the same tasks and policy class \\
Radosavovic et al. (2024) & S2R & Cond & III & 3 & context-length ablations against non-adaptive baselines \\
Rajeswaran et al. (2017) & S2R & Data & II & 3 & internal baselines over seeds; simulated targets only \\
Rusu et al. (2017) & S2R & Param & II & 3 & against fine-tuning and from-scratch on the real arm at matched real budgets \\
Seita et al. (2020) & S2R & Data & II & 3 & input-modality conditions compared over 180 stated physical trials \\
Song et al. (2025) & S2R & Model & III & 3 & against model-free baselines at matched sample and wall-clock budgets \\
J. Tan et al. (2018) & S2R & Model & II & 3 & model-improvement and randomization ablations on the same platform \\
Tobin et al. (2017) & S2R & Data & II & 3 & ablations over randomization components with shared detector and protocol \\
Torne et al. (2024) & S2R & Model & III & 3 & +67\% over an imitation baseline from the same demonstrations \\
Zhuang et al. (2023) & S2R & Param & III & 3 & curriculum and distillation ablations on the same platform \\
Matas et al. (2018) & S2R & Data & II & 2 & randomization ablations in simulation; real evaluation has no comparison condition \\
Sadeghi and Levine (2017) & S2R & Data & II & 2 & compared against real-image-trained perception at a different data scale \\
Schaff et al. (2023) & S2R & Model & III & 2 & co-optimized designs transferred; no matched fixed-morphology baseline \\
Thuruthel, Falotico, et al. (2019) & S2R & Model & II & 2 & learned-model control against open-loop control without a matched budget \\
Thuruthel, Shih, et al. (2019) & S2R & Repr & II & 2 & estimation accuracy reported against alternatives on one platform \\
S. Yang et al. (2024) & S2R & Model & III & 2 & policies trained in the learned simulator compared without a matched real baseline \\
Y. F. Chen, Everett, et al. (2017) & S2R & Param & II & 1 & physical deployment reported against prior published behaviour \\
James et al. (2019) & S2R & Repr & II & 1 & headline comparison is against a differently-resourced system in another paper \\
Rudin et al. (2022) & S2R & Data & II & 1 & training times compared against previously published figures \\
\addlinespace[3pt]
Chi et al. (2025) & Task & Param & III & 4 & +46.9\% across four benchmark suites over seeds with dispersion \\
Cully et al. (2015) & Task & Model & I & 4 & against baselines across damage conditions with medians and dispersion \\
Finn, Abbeel, et al. (2017) & Task & Param & II & 4 & against baselines with reported confidence intervals \\
Ma, Liang, Wang, Huang, et al. (2024) & Task & Cond & III & 4 & 29 environments against human rewards with per-seed dispersion \\
Nichol et al. (2018) & Task & Param & II & 4 & against baselines with reported confidence intervals \\
Touati et al. (2023) & Task & Repr & III & 4 & controlled multi-method comparison on a shared benchmark with dispersion \\
Ball et al. (2023) & Task & Param & III & 3 & against offline-to-online baselines at matched online budgets \\
Barreto et al. (2017) & Task & Repr & II & 3 & against non-transfer baselines over seeds \\
Brohan et al. (2023) & Task & Data & III & 3 & against baselines retrained on the same dataset over stated trial counts \\
Duan et al. (2016) & Task & Cond & I & 3 & against non-adaptive baselines on the same task distributions \\
Eysenbach et al. (2019) & Task & Param & II & 3 & against non-pretrained baselines on shared benchmarks \\
Fu, Zhao, and Finn (2025) & Task & Data & III & 3 & co-training against no co-training from the same demonstrations \\
Goyal et al. (2023) & Task & Repr & III & 3 & against PerAct at the same demonstration counts \\
T. He, Xiao, et al. (2025) & Task & Param & III & 3 & unified controller against per-mode specialists on the same modes \\
Higuera et al. (2025) & Task & Repr & III & 3 & against from-scratch and alternative encoders on a shared touch suite \\
James et al. (2018) & Task & Repr & II & 3 & task embeddings against alternative few-shot baselines \\
Johannink et al. (2019) & Task & Param & II & 3 & residual against pure learning and the hand-designed controller alone \\
Kalashnikov et al. (2021) & Task & Data & II & 3 & multi-task against single-task training at the same data scale \\
Laskin et al. (2023) & Task & Cond & III & 3 & against non-in-context baselines over seeds \\
Luo et al. (2024) & Task & Param & III & 3 & against imitation and prior RL baselines on the same real tasks \\
Luo et al. (2025) & Task & Param & III & 3 & RL fine-tuning against imitation from the same demonstrations \\
Nagabandi et al. (2019) & Task & Model & II & 3 & against non-adaptive model-based baselines on the same hardware \\
A. Nair et al. (2020) & Task & Param & II & 3 & offline-to-online against offline-only and online-only baselines \\
Nakamoto et al. (2023) & Task & Param & III & 3 & against offline-to-online baselines on a shared benchmark \\
Pertsch et al. (2020) & Task & Param & II & 3 & skill prior against flat and non-prior baselines at the same budget \\
Rakelly et al. (2019) & Task & Cond & II & 3 & against meta-RL baselines over seeds on shared benchmarks \\
Ren et al. (2025) & Task & Param & III & 3 & against alternative diffusion-policy fine-tuning methods over seeds \\
Rusu, Colmenarejo, et al. (2016) & Task & Param & I & 3 & distilled against individually trained policies on the same tasks \\
Rusu, Rabinowitz, et al. (2016) & Task & Param & I & 3 & against fine-tuning and from-scratch at the same target budget \\
Shridhar et al. (2023) & Task & Repr & III & 3 & against image-based baselines at the same demonstration counts \\
Song et al. (2023) & Task & Param & III & 3 & reinforcement learning against optimal control on the same platform and task \\
S. Tan et al. (2025) & Task & Param & III & 3 & post-training against the same policy before post-training \\
Tirinzoni et al. (2025) & Task & Repr & III & 3 & against zero-shot RL baselines on a shared benchmark \\
Touati and Ollivier (2021) & Task & Repr & II & 3 & against successor-feature and goal-conditioned baselines \\
Ze et al. (2024) & Task & Repr & III & 3 & 3D against 2D diffusion policies from the same demonstrations \\
G. Zhou et al. (2025) & Task & Model & III & 3 & against alternative world models on shared planning tasks \\
Physical Intelligence, Amin, et al. (2025) & Task & Param & III & 2 & post-training gains reported without a stated matched budget \\
Yuan et al. (2025) & Task & Param & III & 2 & baselines reported without a stated matched target budget \\
\addlinespace[3pt]
Kurin et al. (2021) & Embod & Repr & II & 4 & controlled re-evaluation of graph-based control with per-seed dispersion \\
Ai et al. (2025) & Embod & Data & III & 3 & embodiment-count scaling curves; the physical result is a demonstration \\
Bousmalis et al. (2023) & Embod & Data & III & 3 & adaptation from stated demonstration counts against non-pretrained baselines \\
AgiBot World Contributors et al. (2025) & Embod & Data & III & 3 & pretraining corpus varied with the downstream recipe held fixed \\
Y. F. Chen, Liu, et al. (2017) & Embod & Param & II & 3 & against reciprocal-velocity-obstacle baselines in simulation \\
T. Chen et al. (2018) & Embod & Cond & II & 3 & held-out robots against non-conditioned baselines in simulation \\
L. Y. Chen et al. (2024) & Embod & Model & III & 3 & cross-painting against direct deployment on the same unseen arm \\
L. Y. Chen et al. (2025) & Embod & Data & III & 3 & augmented against unaugmented demonstrations at the same count \\
Dasari et al. (2019) & Embod & Data & II & 3 & pretrained against from-scratch fine-tuning on held-out robots \\
Doshi et al. (2025) & Embod & Param & III & 3 & against per-embodiment specialists at the same data scale \\
Octo Model Team et al. (2024) & Embod & Param & III & 3 & fine-tuned against from-scratch policies at the same demonstration counts \\
Gupta et al. (2022) & Embod & Cond & III & 3 & unseen morphologies against non-tokenized baselines \\
Hejna et al. (2025) & Embod & Data & III & 3 & learned weights against uniform and hand weights over the same pool \\
Huang et al. (2020) & Embod & Param & II & 3 & shared modular policies against monolithic baselines on the same agents \\
Lin et al. (2025) & Embod & Data & III & 3 & diversity-first against demonstration-heavy collection at a fixed budget \\
Long et al. (2018) & Embod & Param & II & 3 & against decentralized baselines at matched team sizes in simulation \\
Open X-Embodiment Collaboration et al. (2024) & Embod & Data & III & 3 & pooled against specialist policies on each lab's own in-domain data \\
Saxena et al. (2025) & Embod & Data & III & 3 & diversity axes disentangled in controlled simulation \\
Shah, Sridhar, Bhorkar, et al. (2023) & Embod & Data & III & 3 & pooled against single-robot training with deployment on unseen robots \\
Shah, Sridhar, Dashora, et al. (2023) & Embod & Param & III & 3 & against prior navigation policies on shared routes \\
Sridhar et al. (2024) & Embod & Param & III & 3 & unified against separate goal-directed and exploration policies \\
T. Wang et al. (2018) & Embod & Repr & II & 3 & graph-structured against flat policies on the same morphology transfers \\
L. Wang et al. (2024) & Embod & Param & III & 3 & +20\% on unseen tasks against an identical trunk trained from scratch \\
Xing et al. (2025) & Embod & Data & III & 3 & constructed narrow and dissimilar sub-datasets under one training recipe \\
J. Yang et al. (2024) & Embod & Data & III & 3 & manipulation and navigation co-training against single-domain training \\
J. Yang et al. (2026) & Embod & Data & III & 3 & +22.5\% by recomposing the same data without new collection \\
Ye et al. (2025) & Embod & Repr & III & 3 & latent-action pretraining against action-labelled pretraining at the same fine-tune budget \\
Bu et al. (2025) & Embod & Repr & III & 2 & headline comparison holds neither compute nor model scale fixed \\
Eschmann et al. (2026) & Embod & Param & III & 2 & zero-shot deployment across ten airframes; no matched per-airframe baseline \\
Reed et al. (2022) & Embod & Data & III & 2 & cross-domain comparisons without a matched robot-side budget \\
Zheng et al. (2025) & Embod & Repr & III & 2 & 0.5B model compared against a 14x larger model \\
\addlinespace[3pt]
Z. Jiang et al. (2025) & Human & Data & III & 4 & synthetic against DemoNoise data at 1{,}000 matched demonstrations over 3 seeds with dispersion \\
Bahl et al. (2023) & Human & Repr & III & 3 & affordance priors against no-prior baselines across the same tasks \\
Duan et al. (2017) & Human & Cond & II & 3 & one-shot conditioning against non-conditioned baselines in simulation \\
Finn, Yu, et al. (2017) & Human & Cond & II & 3 & against baselines from the same demonstrations with stated trial counts \\
Jain et al. (2024) & Human & Cond & III & 3 & video conditioning against language conditioning on the same tasks \\
E. Jang et al. (2022) & Human & Cond & II & 3 & held-out tasks against non-conditioned baselines from the same data \\
Kareer et al. (2025) & Human & Data & III & 3 & one added hour of human against one added hour of robot data \\
Lepert et al. (2025) & Human & Data & III & 3 & human-video-derived data against robot data at a stated budget \\
Z. Liu et al. (2025) & Human & Data & III & 3 & audio versus no-audio on the same demonstrations \\
Mandlekar et al. (2023) & Human & Data & III & 3 & synthetic against human demonstrations at matched counts \\
Peng et al. (2020) & Human & Model & II & 3 & motion-imitation ablations against baselines with real deployment \\
Qiu et al. (2025) & Human & Param & III & 3 & co-training against robot-only training at the same robot budget \\
Shaw et al. (2023) & Human & Repr & III & 3 & video priors against no-prior baselines on the same tasks \\
Torabi et al. (2018) & Human & Model & II & 3 & against imitation baselines with access to actions \\
C. Wang et al. (2023) & Human & Repr & III & 3 & against imitation baselines from the same demonstrations \\
T. Yu et al. (2018) & Human & Cond & II & 3 & human-video conditioning against robot-demonstration conditioning \\
Zakka et al. (2022) & Human & Repr & II & 3 & cross-embodiment reward learning against baselines over seeds \\
Zhao et al. (2023) & Human & Data & III & 3 & against four imitation baselines from the same 50 demonstrations \\
Bahl et al. (2022) & Human & Param & III & 2 & baselines reported without a matched target budget \\
Cheng et al. (2024) & Human & Model & III & 2 & ablations reported; no matched-budget comparison \\
Chi et al. (2024) & Human & Data & III & 2 & collection apparatus differs between the compared conditions \\
Fu, Zhao, Wu, et al. (2025) & Human & Model & III & 2 & system demonstration with component ablations \\
T. He et al. (2024) & Human & Model & III & 2 & simulation baselines; the physical result is a demonstration \\
T. He, Luo, et al. (2025) & Human & Cond & III & 2 & interface ablations; the physical result is a demonstration \\
J. Jang et al. (2025) & Human & Data & III & 2 & generated data reported against a baseline of unstated budget \\
Li et al. (2025a) & Human & Model & III & 2 & single-video imitation with ablations; no matched-budget baseline \\
C. Wang et al. (2024) & Human & Data & III & 2 & component ablations without a matched-budget baseline \\
K. Yu et al. (2025) & Human & Data & III & 2 & baselines reported; human and robot conditions differ in collection budget \\
\addlinespace[3pt]
Burns et al. (2025) & Web & Repr & III & 4 & controlled factorial study of representation properties with dispersion \\
Dasari et al. (2023) & Web & Repr & III & 4 & explicitly unbiased protocol across representations with dispersion \\
Hansen et al. (2023) & Web & Repr & III & 4 & tuned from-scratch against frozen representations over seeds with dispersion \\
Majumdar et al. (2023) & Web & Repr & III & 4 & 17 tasks under one protocol with per-task dispersion \\
Ichter et al. (2023) & Web & Cond & III & 3 & against non-grounded planning baselines over stated trial counts \\
Assran et al. (2025) & Web & Model & III & 3 & against baselines on the same zero-shot manipulation tasks \\
NVIDIA, Bjorck, et al. (2025) & Web & Param & III & 3 & against baselines with data-mixture ablations \\
Black et al. (2025) & Web & Param & III & 3 & against OpenVLA and Octo on shared tasks \\
Physical Intelligence, Black, et al. (2025) & Web & Data & III & 3 & data-composition ablations under one open-world protocol \\
Zitkovich et al. (2023) & Web & Data & III & 3 & against RT-1 and baselines under one evaluation protocol \\
Driess et al. (2023) & Web & Cond & III & 3 & against non-embodied and single-domain baselines on shared tasks \\
Driess et al. (2025) & Web & Param & III & 3 & insulation ablated with the backbone and data held fixed \\
Du et al. (2023) & Web & Model & III & 3 & against non-generative planning baselines on shared tasks \\
G. Jiang et al. (2025) & Web & Repr & III & 3 & against R3M, VC-1 and MVP on a shared evaluation \\
Karamcheti et al. (2023) & Web & Repr & III & 3 & against R3M and MVP on a shared evaluation suite \\
Kim, Pertsch, et al. (2025) & Web & Param & III & 3 & against RT-2-X and Octo across 29 tasks under one protocol \\
Kim, Finn, et al. (2025) & Web & Param & III & 3 & recipe components ablated with the backbone held fixed \\
Li et al. (2026) & Web & Param & III & 3 & controlled study of design choices under one training recipe \\
Liang et al. (2023) & Web & Cond & III & 3 & against language-to-plan baselines on shared benchmarks \\
Ma et al. (2023) & Web & Repr & III & 3 & value-implicit pretraining against alternative representations \\
S. Nair et al. (2023) & Web & Repr & III & 3 & against CLIP, ImageNet and from-scratch encoders at the same demonstration counts \\
Parisi et al. (2022) & Web & Repr & III & 3 & self-supervised features against ground-truth state on the same tasks \\
Pertsch et al. (2025) & Web & Repr & III & 3 & tokenizer variants compared with the policy held fixed \\
Radosavovic et al. (2023) & Web & Repr & III & 3 & masked pretraining against from-scratch encoders at matched budgets \\
Schneider et al. (2024) & Web & Repr & III & 3 & representation choices under one recipe; seeds and dispersion reported but the from-scratch baseline runs at lower input resolution \\
Shang et al. (2025) & Web & Repr & III & 3 & distilled against source vision models on a shared downstream suite \\
Shridhar et al. (2022) & Web & Repr & II & 3 & against Transporter baselines at the same demonstration counts \\
Shukor et al. (2025) & Web & Param & III & 3 & against larger models on shared benchmarks at stated budgets \\
H. Wu et al. (2024) & Web & Model & III & 3 & video pretraining against from-scratch on the same benchmark splits \\
Zawalski et al. (2025) & Web & Cond & III & 3 & reasoning conditioning against the same backbone without it \\
Physical Intelligence et al. (2026) & Web & Param & III & 2 & comparisons reported without a stated matched budget \\
Cheang et al. (2024) & Web & Param & III & 2 & ablations reported; pretraining scale not held fixed \\
Gemini Robotics Team, Abeyruwan, et al. (2025) & Web & Param & III & 2 & comparisons reported with limited protocol detail \\
Gemini Robotics Team, Abdolmaleki, et al. (2025) & Web & Param & III & 2 & comparisons reported with limited protocol detail \\
Hu et al. (2025) & Web & Model & III & 2 & ablations reported without a matched-budget baseline \\
Huang, Xia, et al. (2023) & Web & Cond & III & 2 & feedback ablations without a matched-budget baseline \\
Huang et al. (2022) & Web & Cond & III & 2 & planning comparisons without a matched execution budget \\
Huang, Wang, et al. (2023) & Web & Cond & III & 2 & ablations reported; comparison conditions differ in supervision \\
S. Liu et al. (2025) & Web & Param & III & 2 & ablations reported without a matched-budget baseline \\
Qu et al. (2025) & Web & Repr & III & 2 & ablations reported without a matched-budget baseline \\
Shah, Osiński, et al. (2023) & Web & Cond & III & 2 & component ablations without a matched-budget comparison \\
Shen et al. (2023) & Web & Repr & III & 2 & feature-field ablations without a matched-budget baseline \\
Zhen et al. (2024) & Web & Model & III & 2 & ablations reported without a matched-budget baseline \\
Zhou, Zhu, et al. (2025) & Web & Param & III & 2 & retention comparisons without a matched adaptation budget \\
Figure AI (2025) & Web & Param & III & 0 & capability demonstration with no comparison condition \\
\end{longtable}

\endgroup